\documentclass[letterpaper]{article} %
\usepackage{aaai2026}  %
\usepackage{times}  %
\usepackage{helvet}  %
\usepackage{courier}  %
\usepackage[hyphens]{url}  %
\usepackage{graphicx} %
\usepackage{natbib}  %
\usepackage{caption} %
\usepackage{algorithm}
\usepackage{algorithmic}

\usepackage{booktabs}
\usepackage{placeins} 
\usepackage{amssymb}
\usepackage{amsmath} 
\usepackage{multirow}
\newcommand{\imgwidth}{0.12\linewidth}

\usepackage{newfloat}
\usepackage{listings}
\DeclareCaptionStyle{ruled}{labelfont=normalfont,labelsep=colon,strut=off} %
\floatstyle{ruled}
\newfloat{listing}{tb}{lst}{}
\floatname{listing}{Listing}
\title{Learning Compact Latent Space for Representing Neural Signed Distance Functions with High-fidelity Geometry Details}
\author{
    Qiang Bai\textsuperscript{\rm 1},
    Bojian Wu\textsuperscript{\rm 2},
    Xi Yang\textsuperscript{\rm 1}\thanks{Corresponding author.}, 
    Zhizhong Han\textsuperscript{\rm 3}
}
\affiliations{
    \textsuperscript{\rm 1} Jilin University, Changchun, China \\
    \textsuperscript{\rm 2} Zhejiang University, Hangzhou, China \\
    \textsuperscript{\rm 3} Wayne State University, Detroit, USA \\

    baiqiang23@mails.jlu.edu.cn, 
    ustcbjwu@gmail.com,
    yangxi21@jlu.edu.cn,
    h312h@wayne.edu
}

\begin{document}

\maketitle

\begin{abstract}
Neural signed distance functions (SDFs) have been a vital representation to represent 3D shapes or scenes with neural networks. An SDF is an implicit function that can query signed distances at specific coordinates for recovering a 3D surface. Although implicit functions work well on a single shape or scene, they pose obstacles when analyzing multiple SDFs with high-fidelity geometry details, due to the limited information encoded in the latent space for SDFs and the loss of geometry details. To overcome these obstacles, we introduce a method to represent multiple SDFs in a common space, aiming to recover more high-fidelity geometry details with more compact latent representations. Our key idea is to take full advantage of the benefits of generalization-based and overfitting-based learning strategies, which manage to preserve high-fidelity geometry details with compact latent codes. Based on this framework, we also introduce a novel sampling strategy to sample training queries. The sampling can improve the training efficiency and eliminate artifacts caused by the influence of other SDFs. We report numerical and visual evaluations on widely used benchmarks to validate our designs and show advantages over the latest methods in terms of the representative ability and compactness.
\end{abstract}

\begin{links}
    \link{Code}{https://github.com/eoozbq/Compact-SDF}
\end{links}

\section{Introduction}

\begin{figure*}[tp]
  \centering
  \includegraphics[width=0.9\linewidth]{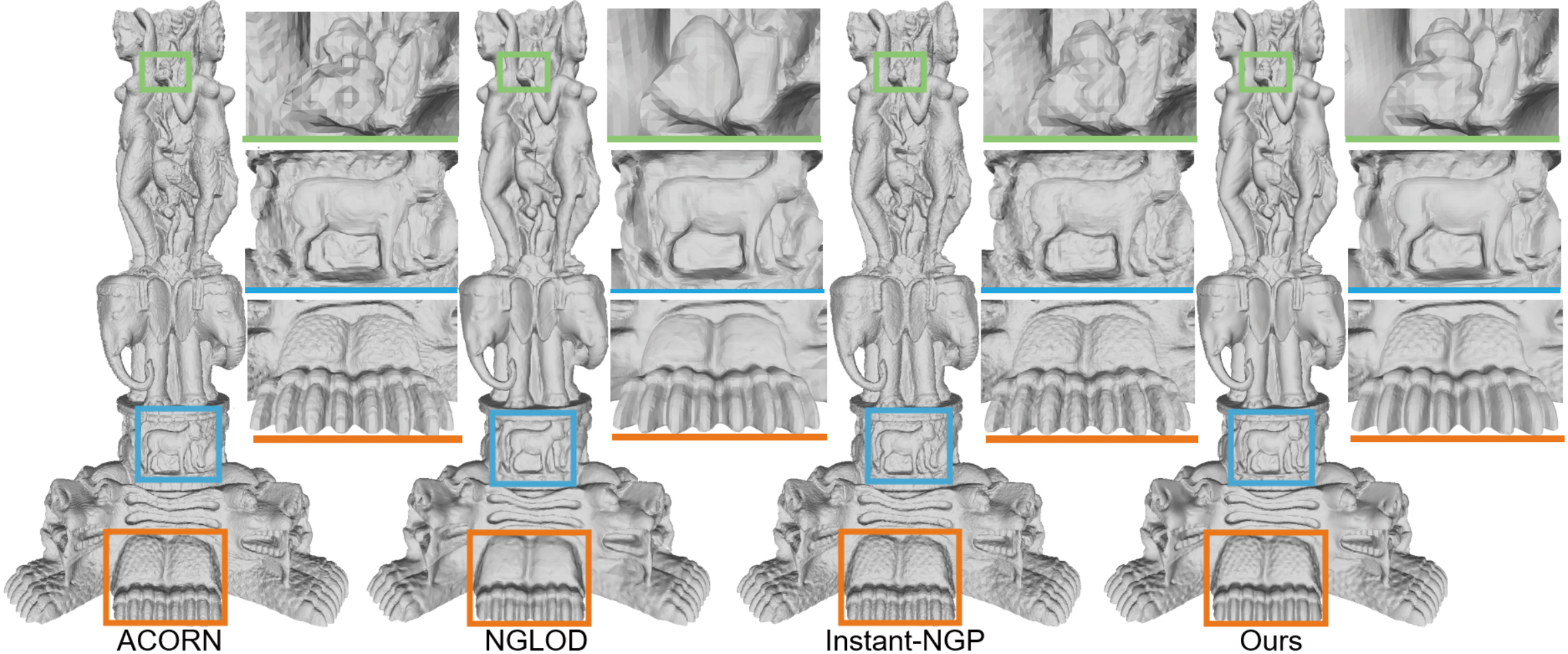}
  \caption{
    Reconstruction of a statuette using Marching Cubes at $512^3$ resolution.
 }
  \label{fig:teaser}
\end{figure*}

Signed distance functions (SDFs) are a vital representation for 3D perception in computer vision and robotics. It predicts a signed distance at an arbitrary location, resulting in an implicit function which can be parameterized well by deep neural networks, dubbed neural SDFs~\cite{Park_2019_CVPR,xie2022neural}. To reconstruct a surface, an SDF predicts signed distances on all vertices of a volume grid, which is then used in the marching cubes algorithm~\cite{10.1145/37402.37422}. Although SDFs are good at representing geometry with arbitrary topology, training neural SDFs requires a large amount of forward and backward procedures, and is limited in recovering high-frequency geometry details, due to the bias of neural networks on low-frequency geometry.

To recover more geometry details, the latest methods employ 3D volume grids to represent signed distances or spatial features that can be decoded into signed distances~\cite{9578205,Chen2021Multiresolution,9578427,mueller2022instant,yariv2023mosaicsdf}. Unlike neural networks, signed distances or spatial features on vertices of a grid can get tuned without a constraint of consistency to the ones on the neighboring vertices, leading to a capability of recovering more high-fidelity geometry details, such as sharper edges and smoother surfaces with less artifacts~\cite{9156855}. These methods usually employ an overfitting strategy to overfit one volume grid on a single shape. Due to the lack of a compact latent space to represent multiple volume grids, these methods can not generalize to other shapes like some priors trained in a data-driven manner. This also makes it challenging to analyze multiple SDFs which are expected to preserve high-fidelity geometry details.

To overcome this challenge, we introduce a method to learn a compact latent space for representing multiple SDFs, where more high-fidelity geometry details can be recovered. Our key idea is to take full advantage of the benefits from generalization-based neural networks and overfitting-based volume grids. Specifically, to learn SDF, we use a volume grid to model the signed distance field near the surface, while using a neural network to model the signed distance field far away from the surface, where the neural network and the volume grid share the same compact latent code representing the shape. This design aims to leverage the generalization ability of neural networks to remove artifacts caused by influence from different shapes, which removes the need of sampling dense queries in areas that are far away from the surface, while using the volume grid to improve the accuracy of the SDF near the surface, where we sample dense queries to significantly improve the geometry details. Based on this design, the learned latent codes are compact and more representative of shapes with high-fidelity geometry details.
We report numerical and visual evaluations to show the superiority of our method over the latest methods on widely used benchmarks. Our main contributions are listed below,
\begin{itemize}
\item We introduce a novel method to represent multiple neural SDFs with high-fidelity geometry details, leading to a compact latent space for representing multiple SDFs. 
\item We introduce a novel sampling strategy to improve the efficiency and performance of learning implicit functions with volume grids.
\item We report the state-of-the-art performance and show advantages over the latest methods on widely used benchmarks in representative ability and compactness.
\end{itemize}

\section{Related Work}\label{sec:related}

Recently, there has been a growing interest in using neural fields~\cite{xie2022neural}, such as occupancy functions~\cite{8953655,Peng2020ECCV} or signed distance functions (SDFs)~\cite{Park_2019_CVPR}, to represent shapes with neural networks, which allows mesh extraction through contouring methods, like marching cubes~\cite{10.1145/37402.37422}.

Early approaches used low-resolution occupancy grids~\cite{choy20163d,Girdhar16b,wang2018global}, while later methods employed octrees~\cite{8237492,wang2017cnn} or sparse grids~\cite{museth2013vdb,setaluri2014spgrid,martel2021acorn} to reduce memory costs, though discretization introduced artifacts. To enable continuous implicit representation, seminal works like DeepSDF~\cite{Park_2019_CVPR}, and others~\cite{8953765,8953655,michalkiewicz2019deep,10.5555/3454287.3455032,hertz2021sape,lipman2021phase,remelli2020meshsdf,davies2021on,ErlerEtAl:Points2Surf:ECCV:2020,9157823,10658156} encoded shapes into latent vectors using the MLPs, allowing distance field queries at any 3D point. Recent techniques improved surface smoothness via auto-decoding~\cite{Park_2019_CVPR}, curriculum learning~\cite{duan2020curriculum}, adversarial training~\cite{kleineberg2020adversarial}, or Lipschitz regularization~\cite{10.1145/3528233.3530713}. However, single global latent codes often lack fine details. Some methods address this by using periodic activations~\cite{sitzmann2020implicit} or Fourier features~\cite{tancik2020fourier} to enhance high-frequency detail recovery.

To better capture local details, recent methods partition 3D space into local grids~\cite{Local_Implicit_Grid_CVPR20,10.1007/978-3-030-58526-6_36,chibane20ifnet,genova2020local,genova2019learning,martel2021acorn,saragadam2022miner,9607613}, assigning each a dedicated latent code or using interpolated feature grids. DeepLS~\cite{10.1007/978-3-030-58526-6_36} stores independent latent codes per grid cell, while NGLOD~\cite{9578205} uses a level-of-detail structure for progressive resolution. MDIF~\cite{Chen2021Multiresolution} employs hierarchical latent grids for multi-resolution decoding, and IMLS~\cite{9578427} defines implicit functions on point sets. Instant-NGP~\cite{mueller2022instant} introduces efficient multi-resolution hashing for compact representation, and MosaicSDF~\cite{yariv2023mosaicsdf} focuses grids near shape boundaries. By encoding only local geometry, these methods achieve finer details than global representations. Unlike existing SDF fitting techniques, our approach learns multiple SDFs in a shared latent space, enabling higher-fidelity details with more compact storage.

Additionally, in recent 3D generation tasks, encoding shapes into a shared latent space is key, as seen in HyperDiffusion~\cite{erkoç2023hyperdiffusion}, 3DILG~\cite{zhang2022dilg}, and 3DShape2VecSet~\cite{zhang20233dshape2vecset}. While these methods focus on latent space mapping, they often overlook geometric details. In contrast, we identify the causes of detail loss and introduce dual branches to jointly encode high- and low-frequency information, enabling more precise geometric reconstruction.

\begin{figure*}[t]
  \centering
  \includegraphics[width=0.85\linewidth]{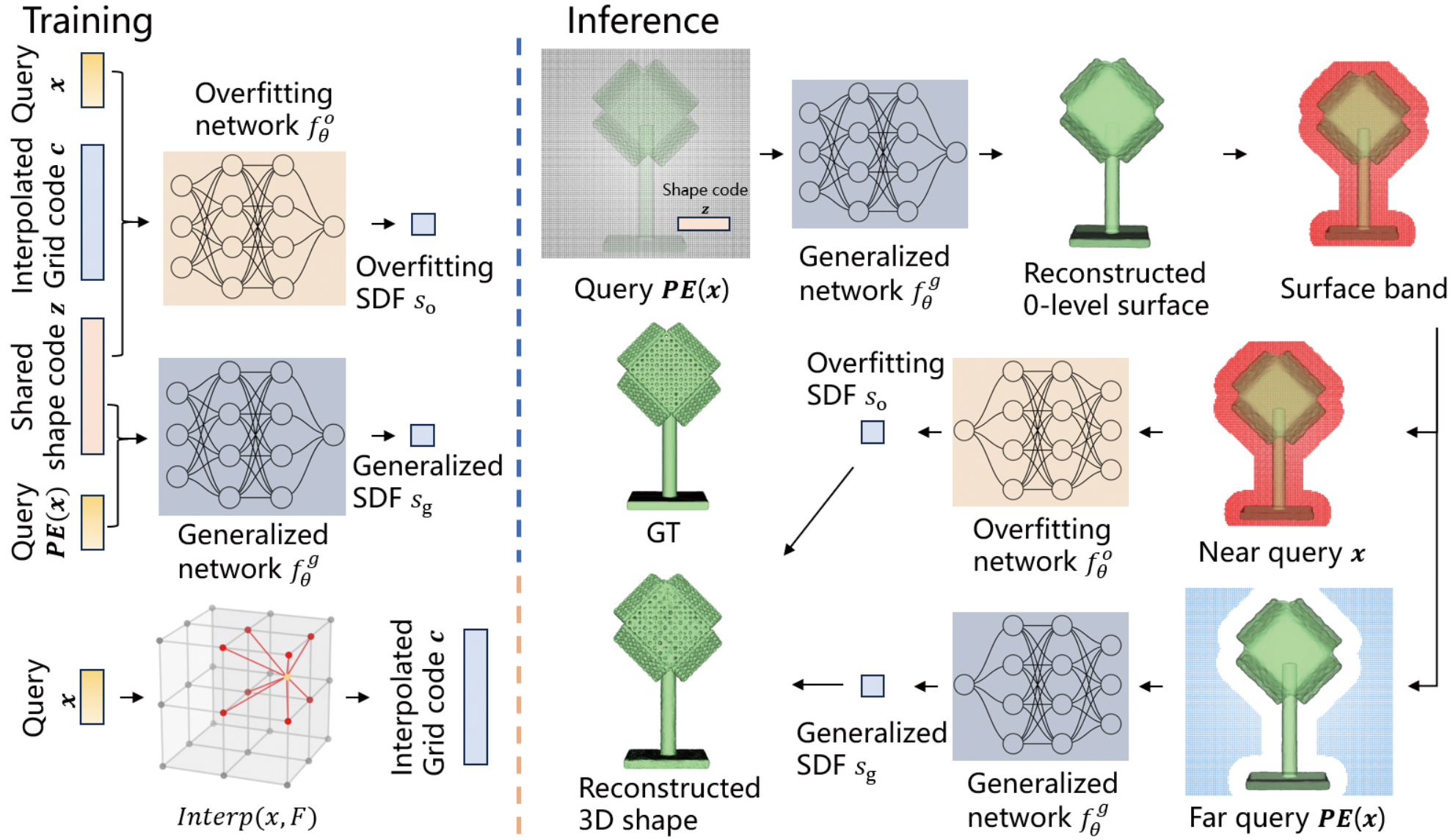}
  \caption{Overview. Our method fully leverages the advantages of both generalization-based and overfitting-based networks, combining their strengths to enhance training efficiency and performance.}
  \label{fig:overview}
\end{figure*}

\section{Method}
\paragraph{Overview.}
We aim to learn multiple SDFs with geometry details in a compact latent space. As illustrated in Fig.~\ref{fig:overview}, our method is formed by two branches, one is a generalization-based network, the other is an overfitting-based network, which shares a learnable latent code that differentiates SDFs. We aim to learn parameters in the two branches, latent codes $\{\mathbf{z}\}$ for representing multiple SDFs, and spatial features on vertices of a volume grid where queries can get spatial features $\mathbf{c}$ by trilinear interpolation. The latent code $\mathbf{z}$ is a sample in the learned latent code space.
\begin{figure}
  \centering
  \includegraphics[width=0.9\linewidth]{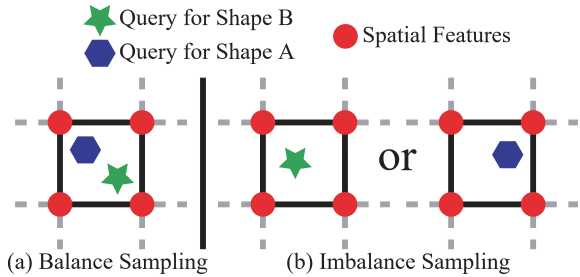}
  \caption{Imbalance query sampling.
  }
  \label{fig:sampling-strategy-result1}
\end{figure}
The first branch maps a coordinate with the latent code into a signed distance. We employ its generalization ability to generalize the prior to predict signed distances in areas that are far away from a surface even we do not sample dense training samples. The second branch works with a spatial feature grid which is shared with different 3D shapes. We first interpolates the feature of a coordinate with the grid, and then maps the coordinate feature and the latent code into a signed distance. We leverage its overfitting of high-frequency 3D signals to preserve fine surface geometry using dense sampling samples.

Our dual branch architecture takes full advantages of the benefits of generalization-based and overfitting-based methods, prompting the strengths of both kinds of methods to improve the training efficiency and performance, aiming to represent multiple SDFs with geometry details using the latent codes. We fuse the signed distance prediction from both branches to reconstruct 3D surfaces using the Marching Cubes algorithm~\cite{10.1145/37402.37422}. 

\paragraph{Motivation.}
Although overfitting-based method works well on single objects with a learnable spatial feature grid, multiple SDFs usually got influence from each other if they share a spatial feature grid with each other. The issue comes from the imbalanced queries sampled for different shapes. For example, one shape has much more sampled queries in one voxel in the spatial feature grid than other shapes. This fact makes features in this voxel lean to be more representative on the shape with more queries, resulting in poor detail recovery on some other shapes.

One solution is to ensure that all shapes have balanced queries sampled in each voxel on the grid, as shown in Fig.~\ref{fig:sampling-strategy-result1} (a). However, that would be a huge computational burden with a grid of cubic complexity, which is barely possible for training. To resolve this issue, we introduce a novel sampling strategy with a balance constraint. We find a trade off between the complexity and performance, i.e., we only ensure that all shapes have queries sampled in each voxel near their surfaces, which is a dense sampling in a bandwidth, while only sampling sparse points outside the bandwidth. Besides the overfitting branch trained by the dense sampling in the bandwidth, our generalization branch will generalize the prior of signed distances in areas outside the bandwidth, which does not produce artifacts even if we merely sample sparse queries outside the bandwidth. Our networks and sampling can resolve the imbalance issues.

\paragraph{Neural Signed Distance Function.}
An SDF is a continuous function that maps a given spatial point $\mathbf{x} \in \mathbb{R}^3$ to a scalar $s \in \mathbb{R}$, indicates the distance from the point to the surface and the sign of $s$ determines whether the point lies inside or outside the surface, defined as $s = SDF(\mathbf{x})$. Then, its zero-level set implicitly defines the underlying 3D shape. Our due-branch network predicts signed distances $s_g$ and $s_o$ at $\mathbf{x}$, which are fused into one distance $s$ during inference.

\paragraph{Generalized SDF.} 
We employ a generalization-based network $s_g = f^g_\theta (PE(\mathbf{x}), \mathbf{z})$ as a branch to estimate a coarse surface, where $PE$ denotes the positional encoding. The network takes a query point $\mathbf{x}$ and a shape code $\mathbf{z}$ as input, and predicts the signed distance $s_g$ at $\mathbf{x}$ to the surface of shape represented by the latent code $\mathbf{z}$. 
We apply positional encoding~\cite{10.1145/3503250} to improve boundary accuracy by introducing high-frequency details.

\paragraph{Overfitting SDF.} 
To capture more high-fidelity geometric details, we employ an overfitting-based network $s_o=f^o_\theta (\mathbf{x},\mathbf{c},\mathbf{z})$ with a shared spatial feature grid. The vertices within the grid hold learnable features, which represent the shape characteristics of objects within that region. Given a 3D query point $\mathbf{x}$, we first interpolate its cube latent code $\mathbf{c}$ from the grid using the trilinear interpolation. Then, the overfitting network takes the concatenation of query point $\mathbf{x}$, the shared shape code $\mathbf{z}$, and the interpolated feature $\mathbf{c}$ as input, and predicts the signed distance $s_o$ at $\mathbf{x}$. 

\paragraph{Signed Distance Fusion.} 
For inference, we fuse the SDF predictions $s_g$ and $s_o$ from both branches into a fused signed distance $s$. We first use the generalization network $f^g_\theta$ to reconstruct a coarse 3D shape using the marching cubes. Then, we extract the surface, find the voxels occupied by the surface, and expand voxels by $n$ layers by both sides of the surface, forming a bandwidth $\mathbb{B}$ of the surface. We rerun the marching cubes with a signed distance fusion below,

\begin{equation}
\label{equ:final}
s =
  \begin{cases}
    s_o=f^o_\theta(\mathbf{x}, \mathbf{c}, \mathbf{z}), & \mathbf{x}\in\mathbb{B} \\
    s_g=f^g_\theta(PE(\mathbf{x}), \mathbf{z}), &  others
  \end{cases}
\end{equation}

\paragraph{Sampling with a balance constraint.} 
To work with our SDF representation, we also propose a sampling with a balance constraint to sample training queries for our overfitting-based network. 
\begin{figure}
  \centering 
  \includegraphics[width=1.0\linewidth]{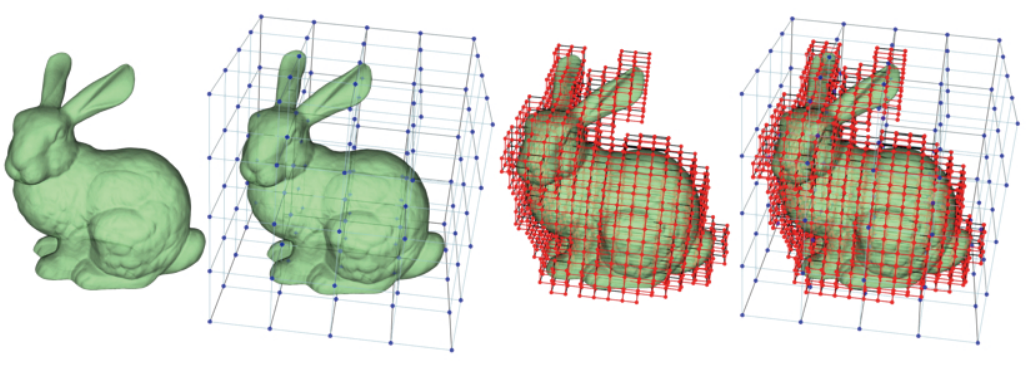}
  \makebox[0.9\linewidth]{\hfill (a)\hspace{5em}(b)\hspace{5em}(c)\hspace{5em}(d)\hfill}
  \caption{Sampling strategy of our method. Red points show dense samplings of the near surface, and blue points show sparse samplings of the far surface.}
  \label{fig:sampling}
\end{figure}

We use dense near-surface and uniform off-surface sampling in Fig.~\ref{fig:sampling} to boost training efficiency. Specifically, we first divide the entire space at low resolution to obtain sparse uniform sampling, as shown in Fig.~\ref{fig:sampling} (b). Next, we divide the space again at a higher resolution. We then find the voxels occupied by the surface of the object, expand them with $n$ more layers of voxels to determine the bandwidth $\mathbb{B}$ of the surface to obtain dense uniform sampling in bandwidth, as shown in Fig.~\ref{fig:sampling} (c). Finally, we merge the two sets and remove duplicates to obtain the training samples in Fig. 5(d).

\paragraph{Optimization.} 
For training, we supervise the signed distance predictions $s_g$ and $s_o$ with the ground truth at the same time. We simultaneously optimize the two networks by minimizing a composite loss function that includes the generalized SDF reconstruction loss with shape code regularization, and the overfitting SDF reconstruction loss with grid code regularization. Thus, the overall loss is formulated as, where the SDF losses $\mathcal{L}_{\text{gen}}$ and $\mathcal{L}_{\text{ovf}}$ are defined as the mean squared error (MSE) between the predicted SDF values $\hat{s}$ and the ground truth $s'$, measuring the discrepancy between the predicted and ground-truth signed distance values. $N$ is the number of sampled query points. 
\begin{equation}
\begin{split}
\mathcal{L} &= \mathcal{L}_{\text{gen}} + \mathcal{L}_{\text{ovf}} + \lambda_1 \mathcal{L}_z + \lambda_2 \mathcal{L}_c \\
&= \frac{1}{N} \sum_{i=0}^N \left\| \hat{s}^g_i - s'_i \right\|_2^2  +\frac{1}{N} \sum_{i=0}^N \left\| \hat{s}^o_i - s'_i \right\|_2^2 \\ 
& \quad + \lambda_1 \frac{1}{\sigma_{\mathbf{z}}^2} \left\| \mathbf{z} \right\|_2^2 
 + \lambda_2 \frac{1}{M} \sum_{j=0}^M \frac{1}{\sigma_{\mathbf{c}_j}^2} \left\| \mathbf{c}_j \right\|_2^2, 
\end{split}
\end{equation}
$\mathcal{L}_{\text{z}}$ and $\mathcal{L}_{\text{c}}$ are the regularization loss terms applied to the shape latent code $\mathbf{z}$ and each grid latent code $\mathbf{c}_j$ to prevent overfitting and encourage smoothness and compactness, where $M$ is the number of grid code. $\lambda_1$ and $\lambda_2$ are coefficients to balance the contributions of each term.

\section{Experiments}
\subsection{Implementation details}

\paragraph{Dataset.}
We conduct experiments on three datasets: The Stanford Models~\cite{10.1145/237170.237269}, ShapenetCore.v2~\cite{chang2015shapenetinformationrich3dmodel}, and D-FAUST~\cite{dfaust:CVPR:2017}. 
We select six detail-rich models from the Stanford 3D Scanning Repository for the reconstruction of single complex shapes.
For ShapenetCore.v2, we select six object categories with the same train/test splits as DeepSDF~\cite{Park_2019_CVPR}. On D-FAUST, we select models from three different human subjects, with each subject performing a variety of motion sequences, such as chicken\_wings, jiggle\_on\_toes, and light\_hopping\_loose. We follow the train/test splits from~\cite{10378388}. We preprocessed each mesh as follows: first, we scaled it to fit the unit cube $[-0.9, 0.9]^3$. Then, we used Manifold~\cite{huang2018robust} to make the mesh watertight before sampling.

\paragraph{Metrics.}
We use Chamfer Distance (CD) $\mathcal{L}$, F-Score, precision, and recall to provide a comprehensive evaluation of our method. Chamfer distance captures overall accuracy, while F-Score, precision, and recall focus on the quality and completeness of surface predictions. 

\paragraph{Setting.}

We employ a higher learning rate {$1.0 \times 10^{-1}$} to optimize the grid code (a $128^3$ grid with $128$-dim feature code for each grid), and a lower learning rate {$1.0 \times 10^{-3}$} for the shape code (a $256$-dim code) to ensure stable convergence and prevent overfitting to early-stage representations. The bandwidth of n=3 defines the near and far surfaces, which are sampled at $512^3$ and $128^3$, respectively. Both networks use 256-dim codes and eight hidden layers of 512 units. The learning rate decays by a factor of 0.5 every 1000 epochs, with a total of 4000 training epochs. 

\subsection{Reconstruction}

\begin{figure*}[htbp]
  \centering
\includegraphics[width=0.85\linewidth]{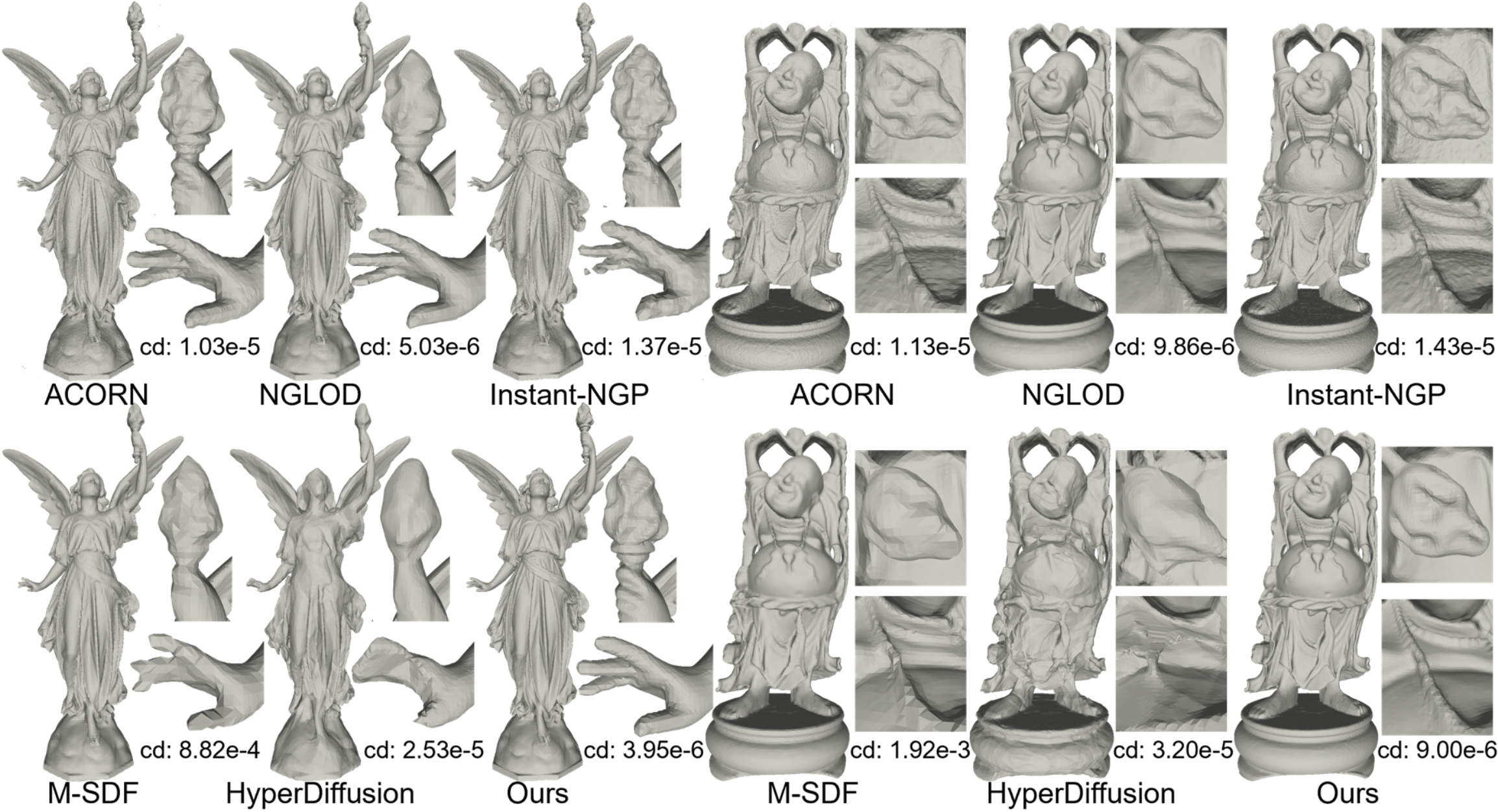} 
  \caption{Comparison in single complex object reconstruction on the Stanford models.}
  \label{fig:single-shape-lucy-happy-vrip}
\end{figure*}

\paragraph{Single complex objects.}
We demonstrate the superiority of our method in preserving fine-grained geometric details in surface reconstruction on complex objects, compared to five SOTA approaches, including Instant-NGP~\cite{mueller2022instant}, NGLOD~\cite{9578205}, ACORN~\cite{martel2021acorn}, MosaicSDF~\cite{yariv2023mosaicsdf}, and HyperDiffusion~\cite{erkoç2023hyperdiffusion}.
The results are reconstructed by marching cubes~\cite{10.1145/37401.37422} with a uniform resolution of $512$. As shown in Fig.~\ref{fig:single-shape-lucy-happy-vrip}, our method successfully reconstructs Lucy and Happy Buddha with high-fidelity geometry details and sharp edges, such as the hand, torch, skirt folds, petals, and prayer beads.
In addition to visual comparison, we sample 500K points from the reconstructed surfaces for quantitative evaluation. As shown in Table~\ref{tab:fitting-result-table}, our method consistently outperforms existing approaches across all metrics, accurately recovering high-fidelity shape details that other methods fail to capture.

\setlength{\tabcolsep}{1mm}
\begin{table}[h]
\small
\centering
\begin{tabular}{l ccccccc}
\toprule
Method & CD$(\downarrow)$  & F-Score$(\uparrow)$ & Precision$(\uparrow)$  & Recall $(\uparrow)$  \\
\midrule
ACORN & 6.76e-05 & 0.982 & 0.967 & 0.998 \\
NGLOD & 6.77e-05 & 0.980 & 0.968 & 0.994 \\
Instant-NGP & 7.37E-05 & 0.976 & 0.962 & 0.990 \\
M-SDF & 1.30e-03 & 0.902 & 0.846 & 0.972 \\
HyperDiffusion & 1.20e-04 & 0.835 & 0.847 & 0.823\\
\textbf{Ours} & \textbf{6.56e-05} & \textbf{0.983} & \textbf{0.969} & \textbf{0.998} \\
\bottomrule
\end{tabular}
\caption{Quantitative comparison on the Stanford models.}
\label{tab:fitting-result-table}
\end{table}

\paragraph{Multiple 3D objects.}

\setlength{\tabcolsep}{1mm}
\begin{table}
    \small
    \centering
    \begin{tabular}{lccccccc}
    \toprule
    Method  & Bench & Chair & Plane & Table & Lamp & Sofa  & Avg. \\
    \midrule
    DeepSDF  &  4.890 & 8.630 & 2.660 & 6.330 & 14.63 & 5.040 & 7.030 \\
    IF-NET  & 1.340 & 1.000 & 0.225  & 0.857  & 0.817  & 1.100 & 0.890\\
    3DILG    & 0.872 & 1.380 & 0.355 & 1.280 & 4.700 & 1.510 & 1.683 \\
    InsNGP   & 0.881 & 1.210 & 0.664 & 1.030 & 1.880 & 1.100 & 1.128\\
    M-SDF    & 2.005 & 15.918 & 4.101 & 6.630  & 18.428 & 26.813 & 12.316\\
    HyperDiff & 1.640  & 1.490 & 2.310  & 1.470  & 6.990  & 2.830 & 2.788 \\
    \textbf{Ours} & \textbf{0.463} & \textbf{0.898} & \textbf{0.223} &\textbf{0.790} & \textbf{0.517} & \textbf{0.751} & \textbf{0.607}\\
    \bottomrule
    \end{tabular}
\caption{Comparison in multi-objects reconstruction on ShapeNet (Note: CD $\times \small 10^{-4}$, InsNGP denotes Instant-NGP, and HyperDiff denotes HyperDiffusion).}
\label{tab:evaluations-multiple-3d-objects}
\end{table}

As presented in Table~\ref{tab:evaluations-multiple-3d-objects}, our approach consistently outperforms six state-of-the-art methods across all evaluation metrics on ShapeNet. Furthermore, Fig.~\ref{fig:multi-shapes-reconstruction-results} illustrates its ability to reconstruct complex lamp geometries without introducing artifacts in empty regions, while faithfully recovering fine details, particularly in objects containing small holes.

Compared to the existing methods, our method leverages spatially-aware grid features combined with global latent codes, enabling it to capture both coarse structure and fine-grained geometric details. 
Especially, since IF-Net relies on convolutional architectures that require dense spatial context, when we trained with global query points but tested on partial inputs (e.g., half of the query points), IF-Net struggles to extract meaningful features, and fails to recover complete human bodies. 

\begin{figure}[H]
  \centering
\includegraphics[width=1\linewidth]{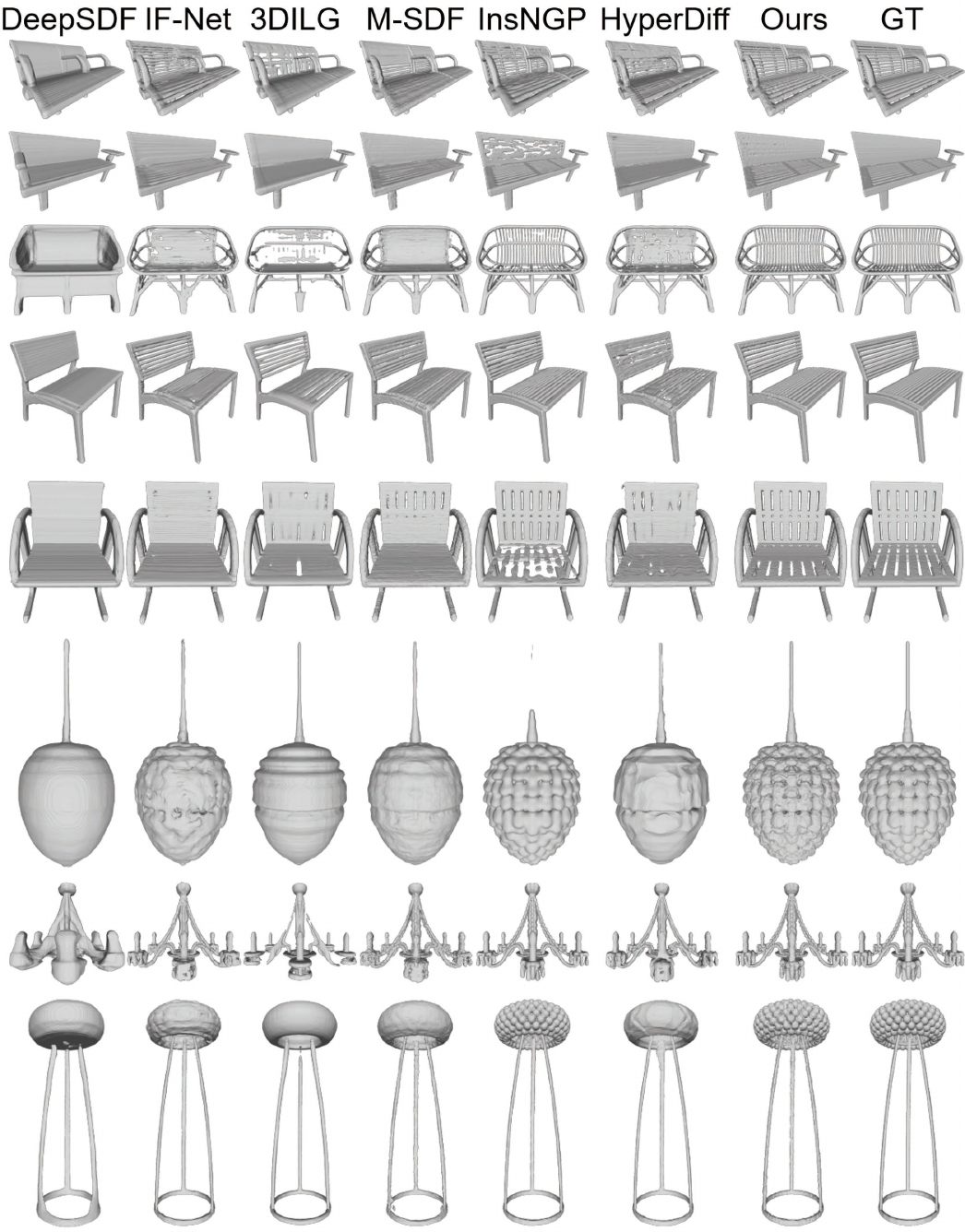} 
  \caption{Comparison in multiple object reconstruction on ShapeNet.)}
  \label{fig:multi-shapes-reconstruction-results}
\end{figure}

Like Instant-NGP~\cite{mueller2022instant}, our method employs grid-based features to encode local shape information. However, while Instant-NGP demonstrates remarkable performance in single-object reconstruction, its effectiveness diminishes notably in multi-object reconstruction scenarios, even when trained with our sampled query points. This performance degeneration is caused by its limited capacity to distinguish between different shapes within a shared latent space. Specifically, hash collisions and the finite resolution of the hash table undermine the fidelity of feature representations, degenerating geometry details in reconstructions.

\subsection{Shape completion}
\begin{figure}[H]
  \centering
  \includegraphics[width=\linewidth]{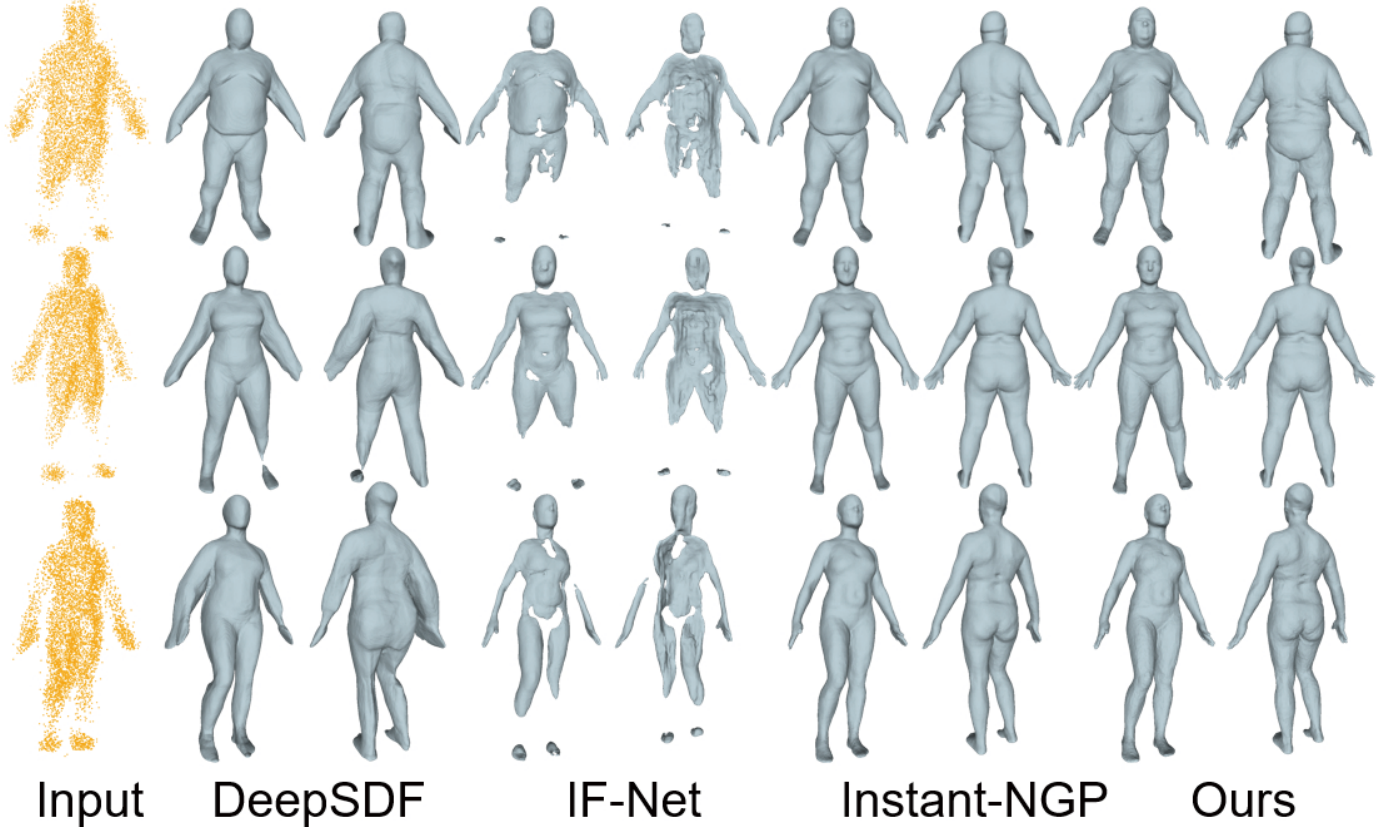}  
  \caption{ \small Comparison results in shape completion on DFAUST.}
  \label{fig:shape completion result}
\end{figure}

We evaluate our method on shape completion using a subset of the DFAUST dataset~\cite{dfaust:CVPR:2017}. As shown in Table~\ref{tab:shape-completion}, our method achieves a superior quantitative performance compared to existing approaches, demonstrating its effectiveness in handling incomplete input data. Furthermore, Fig.~\ref{fig:shape completion result} shows that our method successfully reconstructs partial human models while preserving fine-grained geometric details, such as the sharp structures of the fingers.
\begin{table}[h]
    \centering
    \begin{tabular}{lcccc}
    \toprule
    Method  & 50002 & 50022 & 50025  & Avg.\\
    \midrule
    DeepSDF  &  3.87 & 3.39 & 3.48& 3.58 \\
    IF-NET   &  391.79  & 486.36 & 544.60 & 474.25 \\
    Instant-NGP   & 0.342 & 0.523 & 0.345 & 0.403\\
    \textbf{Ours} & \textbf{0.134} & \textbf{0.124} & \textbf{0.115}& \textbf{0.124}\\
    \bottomrule
    \end{tabular}
\caption{\small Shape completion results on DFAUST (CD $\times$ {\small $10^{-4}$}).}
\label{tab:shape-completion}
\end{table}

\FloatBarrier
\subsection{Shape interpolation}
\begin{figure}[h]
  \centering
\includegraphics[width=1\linewidth]{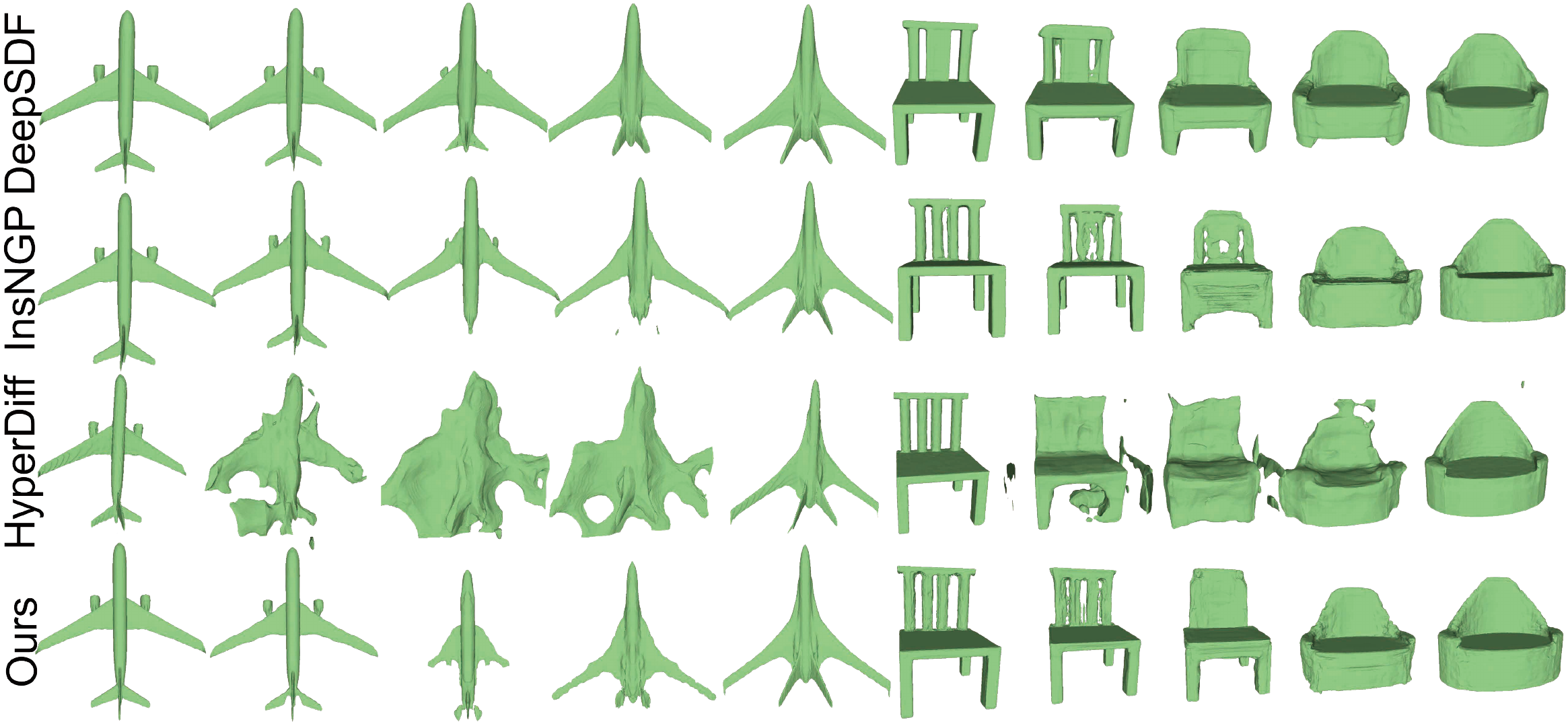} 
  \caption{Comparison of shape interpolation on ShapeNet.}
  \label{fig:plane-shape-interpolation-result}
\end{figure}

We report the results of shape interpolation between the latent codes of two shapes to demonstrate the better continuity and semantic significance of our learned shape embedding, compared to existing methods. The length of latent shape code used in our method, DeepSDF, and Instant-NGP is 256-dimensional, while HyperDiffusion adopts a 3-layer 128-dimensional MLP.
As shown in Fig.~\ref{fig:plane-shape-interpolation-result}, compared to DeepSDF~\cite{Park_2019_CVPR} and Instant-NGP~\cite{mueller2022instant}, our method produces more pronounced shape changes when interpolating between shapes. While DeepSDF has smoother transitions, especially in regions like airplane wings, our method allows for more noticeable shape transformations, capturing more complex geometry changes.
For HyperDiffusion~\cite{erkoç2023hyperdiffusion}, interpolating directly in their latent space results in unrealistic geometries or structural artifacts, since each object is represented by a separate MLP and the complexity of the MLP weights leads to nonlinear transitions.
Moreover, our method learns more compact latent codes than HyperDiffusion. Each latent code is a one-column vector, and its dimension is flexible to set. While HyperDiffusion uses parameters of a neural network to represent a shape, which is not compact. 
While DeepSDF and Instant-NGP adopt similar codes, their bias toward low-frequency signals and resulting artifacts hinder accurate representation of multiple neural SDFs with detailed geometry.

\subsection{Ablation study}
\paragraph{Overfitting/Generalization.}

\begin{figure}[H]
  \centering
  \includegraphics[width=0.6\linewidth]{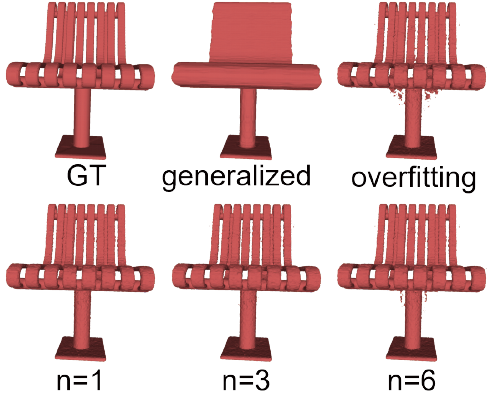} 
 \begin{tabular}{ccc|c|c}
        \hline
        \multicolumn{3}{c|}{Dual (Ours)} & \multirow{2}{*}{Gene-only} & \multirow{2}{*}{Over-only} \\ \cline{1-3}
        n=1       & n=3      & n=6      &                            &                            \\ \hline
        4.01      & 4.05      & 11.6     & 4.29                       & 14.6                       \\ \hline
        \end{tabular}
    
  \caption{Comparison with single network under different bandwidths $n$ in CD $\times$ \small $10^{-4}$.}
   \label{fig:over-gene}
\end{figure}

We report our ablation experiments on 100 randomly selected chair models. To validate the effectiveness of our dual-branch architecture, we separately evaluated the generalization-based and overfitting-based networks using the CD metric in Fig.~\ref{fig:over-gene}. 
All experiments use the settings in Sec setting.

The generalization branch captures coarse geometric features to establish a reasonable shape prior, while the overfitting branch focuses on recovering high-frequency details. The reconstructions show that generalization branch can recover a surface with low-frequency geometry while the overfitting branch can recover a sharp surface but have some artifacts caused by imbalanced sampling outside the bandwidth. Eventually, the signed distance fusion can address the issues of reconstruction from both branches. 

This demonstrates that combining signed distances from both branches not only removes artifacts of single-branch reconstruction but also captures fine surface geometry.

\paragraph{Bandwidth.}
We report the impact of bandwidth in Fig.~\ref{fig:over-gene}. With the trained networks, we try other sizes of bandwidth in signed distance fusion, including 1, 3, and 6. The results show that the artifacts will inherit from the overfitting branch if the bandwidth is too large.

\paragraph{Shared shape code dimensions.}
To explore the impact of the shape code dimensions on shape reconstruction fidelity, we compare the reconstruction results using different dimensions ($d=8, 16, 32, 64, 128$) in Table~\ref{tab:ablation-latent}. The performance keeps getting improved with higher dimensions until $d=64$, and maintains afterward. Note that we only use 100 shapes in this experiment, and we may need higher dimensions for larger datasets.

\paragraph{Sampling strategy.} We highlight our sampling by comparing with uniform sampling on grids. We try to use vertices of grids with different resolutions including $32^3$, $64^3$, $128^3$, and $256^3$. We report the numerical comparison with ours in Table.~\ref{tab:different-grid-resolution-result}. We see that the performance can get improved with higher resolutions but is still worse than ours. Our sampling can produce more samples near the surface, which improves the recovery of geometry details, and relatively sparser sampling outside the bandwidth, which reduces the training burden by using much fewer samples and ensures the performance by the generalization of our method.
\begin{table}[h]
    \centering
    \begin{tabular}{lccccc}
    \toprule
    Dim. & 8 & 16 & 32 & 64 & 128\\
    \midrule
    Ours & 0.443 & 0.433 & 0.410 & 0.397 & 0.409 \\
    \bottomrule
    \end{tabular}
\caption{Impact of latent Code Dimensions on reconstruction in (CD $\times$ \small{$10^{-4}$}).}
\label{tab:ablation-latent}
\end{table}

\begin{figure}
  \centering
  \includegraphics[width=0.8\linewidth]{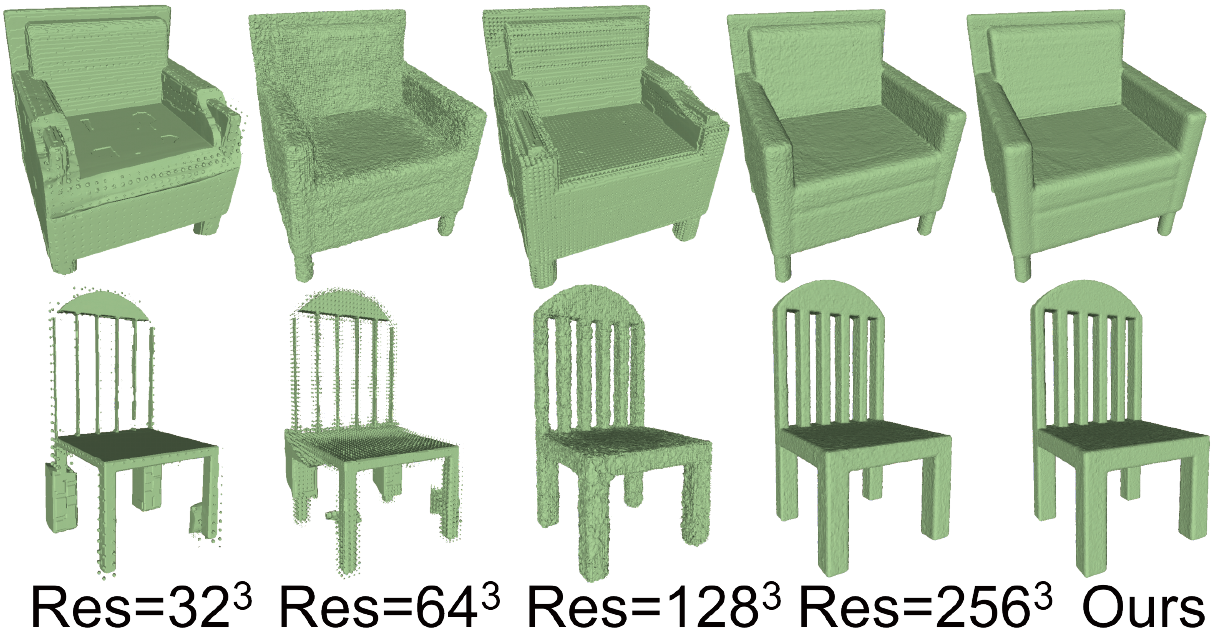}  
  \caption{Reconstruction results with different grid resolution sampling $res$.}
  \label{fig:grid resolution sampling}
\end{figure}

We compare reconstruction results in Fig.~\ref{fig:grid resolution sampling}. At low resolution (res=32), the sampling density is insufficient to supervise the $128^3$ spatial feature grid, leading to incomplete surfaces and poor reconstruction. As the resolution increases to 64 and 128, more query points fall within each grid cell, resulting in improved reconstructions with smoother surfaces and better preservation of fine details such as holes and sharp edges. In particular, our results show sharper surfaces.

\begin{table}[h]
    \centering
    \begin{tabular}{lccccc}
    \toprule
    Res. & $32^3$ & $64^3$ & $128^3$ & $256^3$ & Ours\\
    \midrule
    CD & 35.900 & 13.100 & 0.890 & 0.417 & 0.401 \\
    Samples& $32^3$ & $64^3$ & $128^3$ & $256^3$ & \textbf{$\approx 220^3$}\\
    \bottomrule
    \end{tabular}
  \caption{Different grid resolution on reconstruction in (CD $\times$ \small$10^{-4}$).}
\label{tab:different-grid-resolution-result}
\end{table}

\section{Conclusion}
We propose a method for learning a compact latent space to represent multiple neural SDFs with details. We leverage the strengths of both overfitting-based and generalization-based strategies for SDF representation. The generalization branch effectively captures the signed distance priors in regions far from the surface, even with sparse training samples. In contrast, the overfitting branch accurately reconstructs high-fidelity geometric details and mitigates interference from other shapes via a novel query sampling strategy, which achieves a favorable trade-off between learning efficiency and reconstruction quality. Extensive comparisons show our superiority over SOTA methods.

\section{Acknowledgement}
This work was supported by the Young Scientists Fund of the National Natural Science Foundation of China (Grant No. 62206106).

\bibliography{aaai2026}

\appendix

\setcounter{secnumdepth}{0}
\section{Limitation}
Although our method effectively recovers fine details in reconstruction tasks, it also has certain limitations. Our method employs a supervised strategy to learn neural SDFs, which requires large-scale signed distances ground truth. This fact limits our capability of learning to infer geometry without signed distance supervision, which is the unsupervised learning strategy. In addition, the setting of hyperparameters will also affect the reconstruction results, e.g., different bandwidth layers will lead to residual noise or discontinuity of shape.

\section{Computational Resources}
Experiments were conducted on a Linux server running Ubuntu 20.04 (kernel 5.15), equipped with an Intel Xeon Gold 6258R CPU @ 2.70GHz,  512 GiB of RAM, and 4 $\times$ NVIDIA A40 GPUs (48 GB). We train our model on a single A40 GPU for each training task with memory utilization of around 70\%.

\section{Video Demonstration and Code}
Please watch our video for more visualization on Stanford scan repo dataset, ShapeNet, and D-FAUST.
We also provide the key code in the supplementary materials for reviewing our method.

\section{Implementation Details of Baseline Methods}
For fairness and reproducibility, all baseline methods are trained using their publicly available implementations and default hyperparameters. For both ACORN~\cite{martel2021acorn} and NGLOD~\cite{9578205}, we use the official implementations and default training configurations provided by the authors. For 3DILG ~\cite{zhang2022dilg}, we directly evaluate the pretrained model provided by the authors on multiple categories, using their official code and data setup. Others are as follows:

\paragraph{Mosaic-SDF~\cite{yariv2023mosaicsdf}}
We use a third-party implementation of Mosaic-SDF, available at \url{https://github.com/cortwave/MosaicSDF} to produce its results. For the reconstruction task, we use the model in its default configuration and the mesh as input, without any additional supervision.

\paragraph{IF-NET~\cite{chibane20ifnet}}
IF-Net is capable of predicting implicit shape representations based on various input modalities, such as voxels or point clouds.  In the reconstruction task, we use a latent grid of resolution $32^3$ to represent the shape.
For the human shape completion task, we use point clouds as input, following the same training strategy as our method. During training, the model is supervised using full point clouds sampled from normalized watertight meshes. At test time, only half of the point cloud is provided to simulate partial observations, and the model is evaluated on its ability to reconstruct the complete human shape from this limited input.

\paragraph{Instant-NGP~\cite{mueller2022instant}} 
we adopt the third-party implementation based on tiny-cuda-nn ~\cite{tiny-cuda-nn}. To enable multi-object reconstruction across all methods, we introduce a shape-specific latent code for each object during training. In our configuration of Instant-NGP, we reduce the $per\_level\_scale$ parameter of the hash grid encoder from 2 to 1.174, since we did not produce plausible results with the default 2.

\paragraph{HyperDiffusion~\cite{erkoç2023hyperdiffusion}}
Following the official code, each 3D object in HyperDiffusion is modeled by a separate MLP, which is trained independently and used directly for shape reconstruction.

\begin{figure*}[b!]
  \centering
\includegraphics[width=1\linewidth]{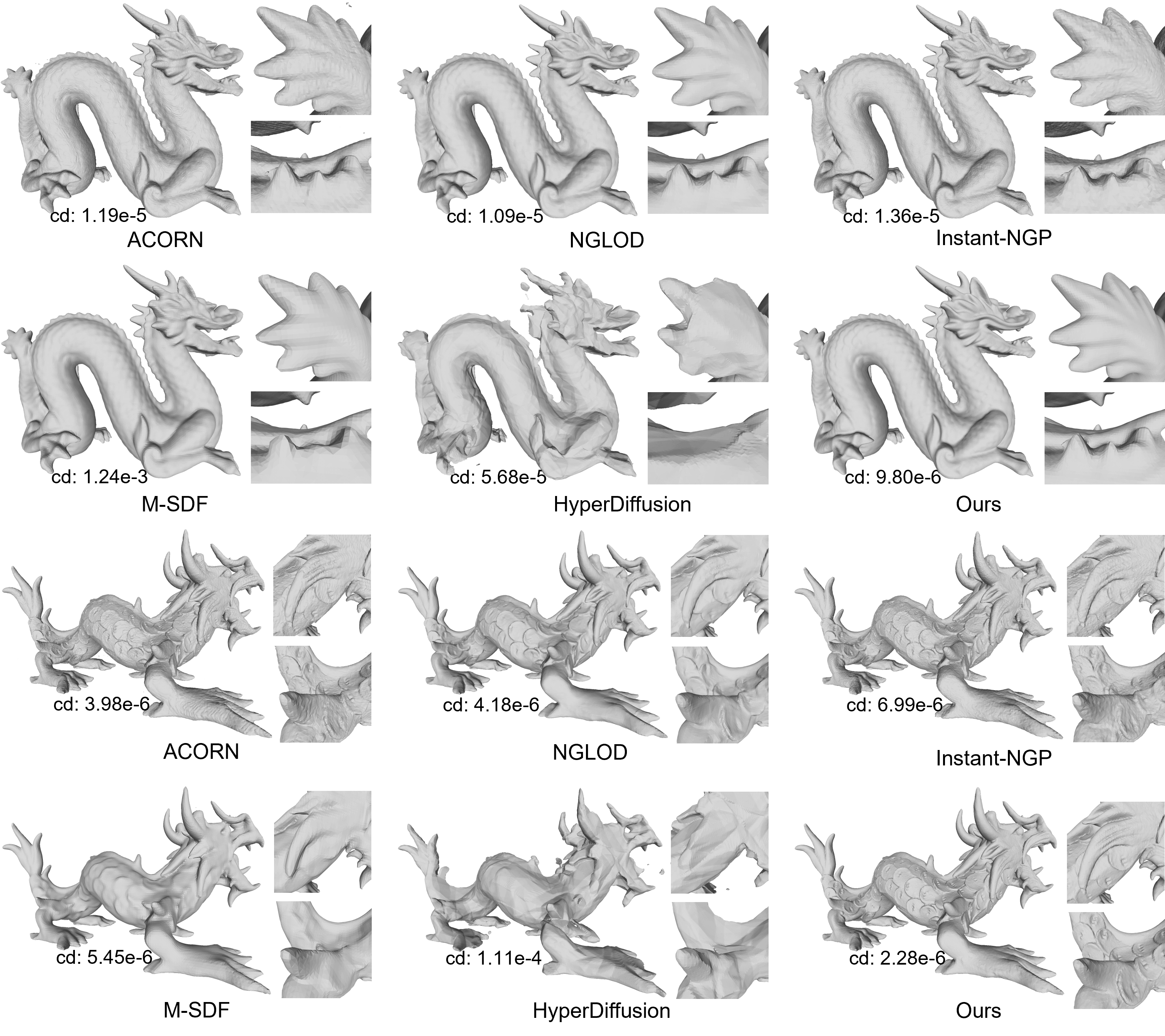} 
  \caption{Visualization results of the Stanford Dragon and Asian Dragon model}.
  \label{fig:supp_complex_dragon}
\end{figure*}

\section{More Results}

\subsection{Single Complex Object Reconstruction}

\begin{figure*}
  \centering
\includegraphics[width=1.0\linewidth]{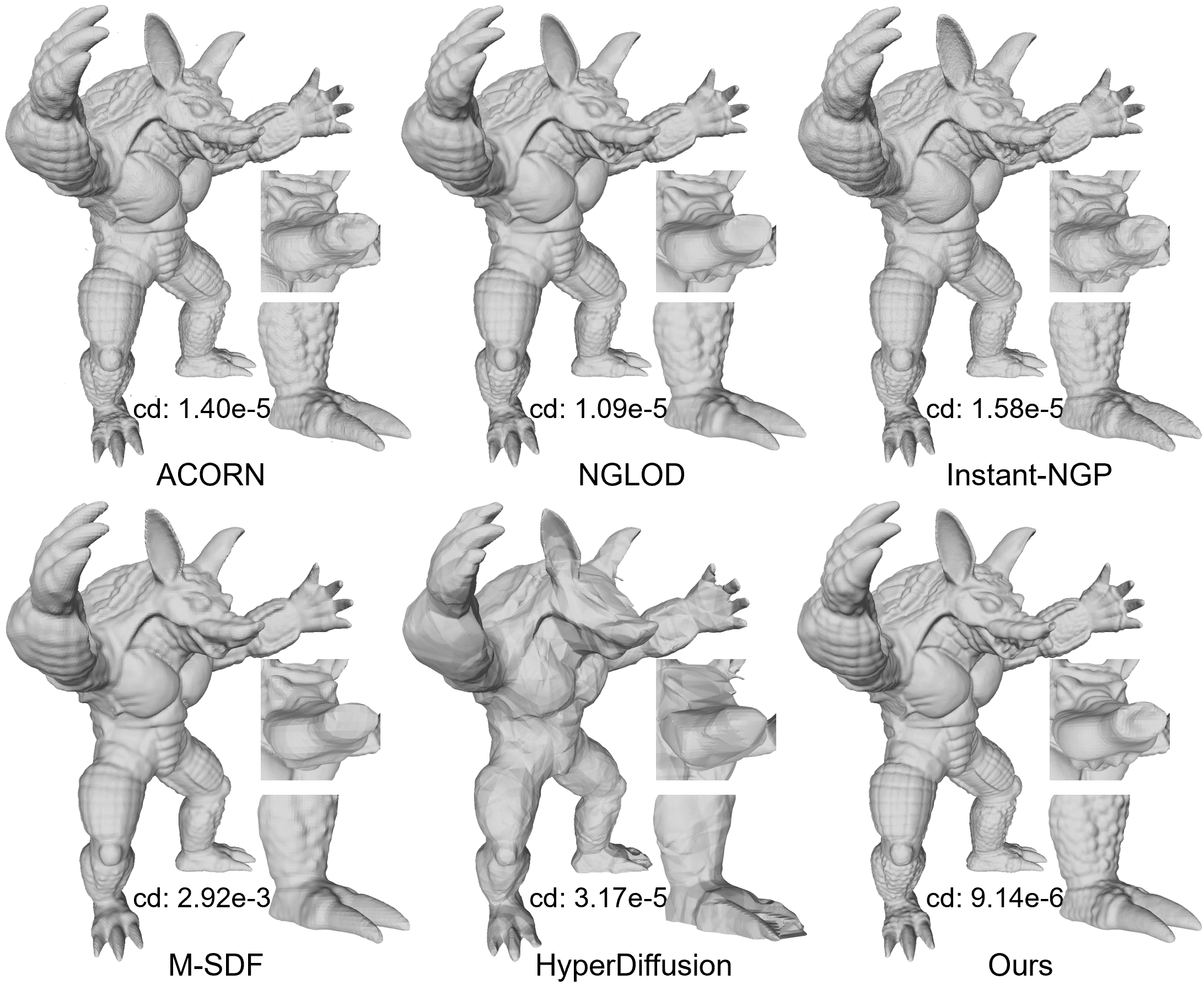} 
  \caption{Visualization results of the Stanford Armadillo model}.
  \label{fig:supp_complex_armadillo}
\end{figure*}

We report a visual comparison in reconstruction on the other three objects in the Stanford scan repo, as shown in Fig.~\ref{fig:supp_complex_dragon} and ~\ref{fig:supp_complex_armadillo}.
For example, in the Dragon model reconstruction, our method achieves smoother surfaces and better details, especially around the teeth and scales. And for the Armadillo model, it captures fine textures and indentations in areas like the nose and feet.
In contrast, although M-SDf shows better surface quality than HyperDiffusion, it has a higher Chamfer Distance due to internal pseudo-surfaces caused by third-party implementation issues.

\begin{figure*}
  \centering
\includegraphics[width=1.0\linewidth]{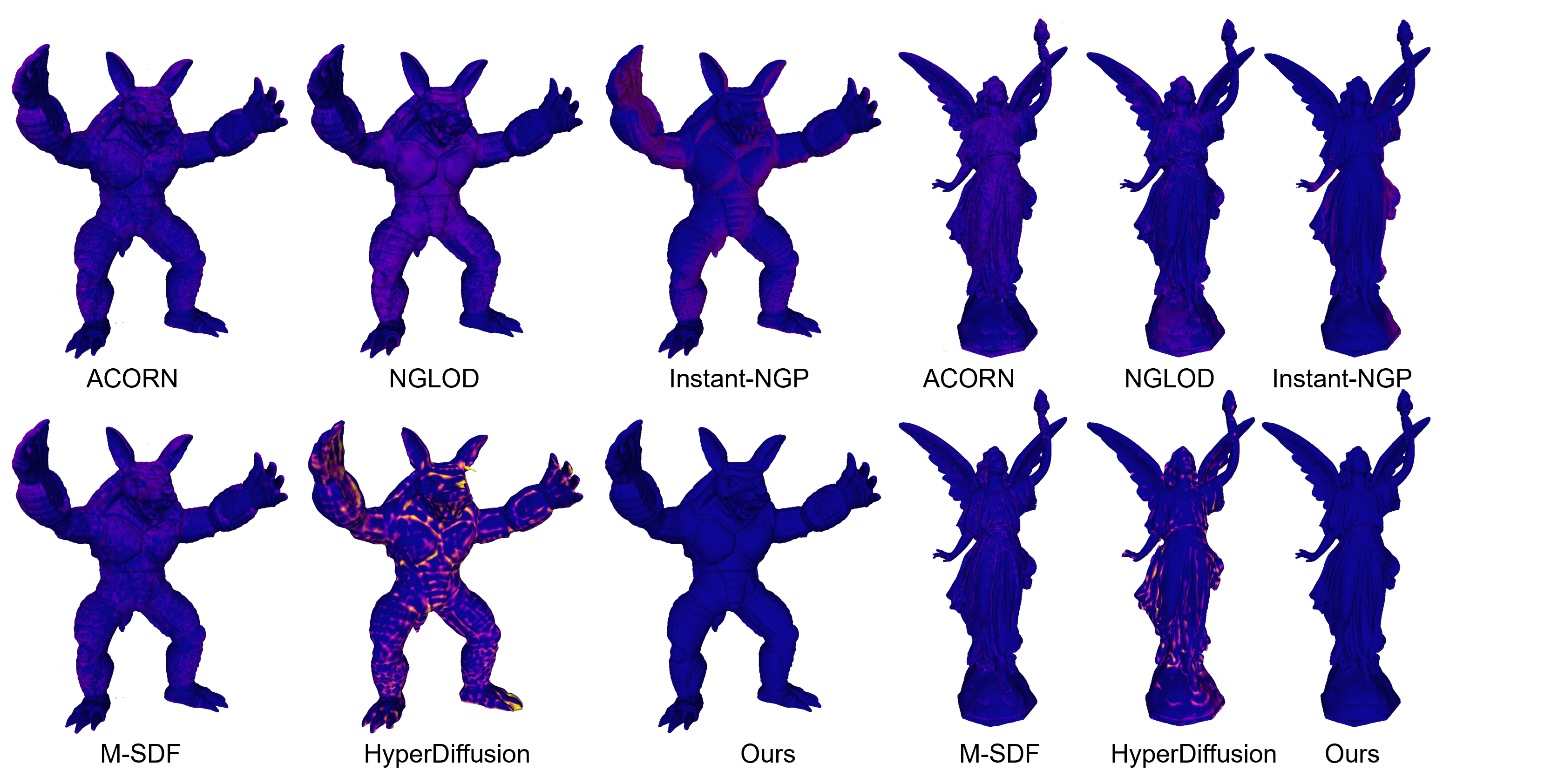}
  \caption{Error map comparison on Armadillo and Lucy.}
  \label{fig:supp_armadillo_lucy_errormap}
\end{figure*}

We visualize the reconstruction accuracy using error maps. Specifically, the error is computed by querying the ground-truth SDF at each vertex of the reconstructed mesh. The resulting values indicate the reconstruction accuracy at each point, as shown in the Fig.~\ref{fig:supp_armadillo_lucy_errormap}

\begin{figure*}
  \centering
\includegraphics[width=1.0\linewidth]{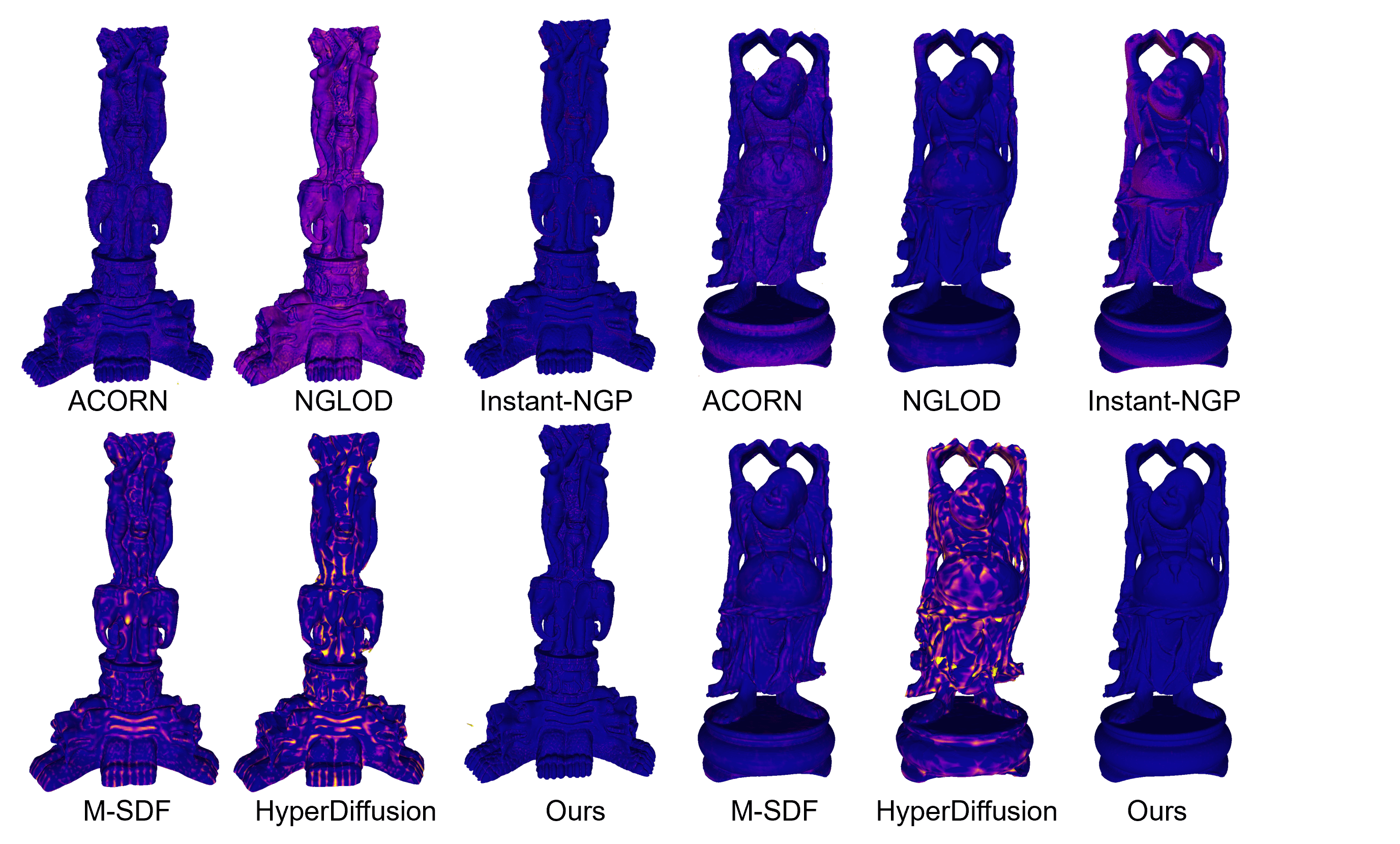} 
  \caption{Error map comparison on Happy Buddha and Thai Statue. }
  \label{fig:supp_happy_errormap}
\end{figure*}

\begin{figure*}
  \centering
\includegraphics[width=1.0\linewidth]{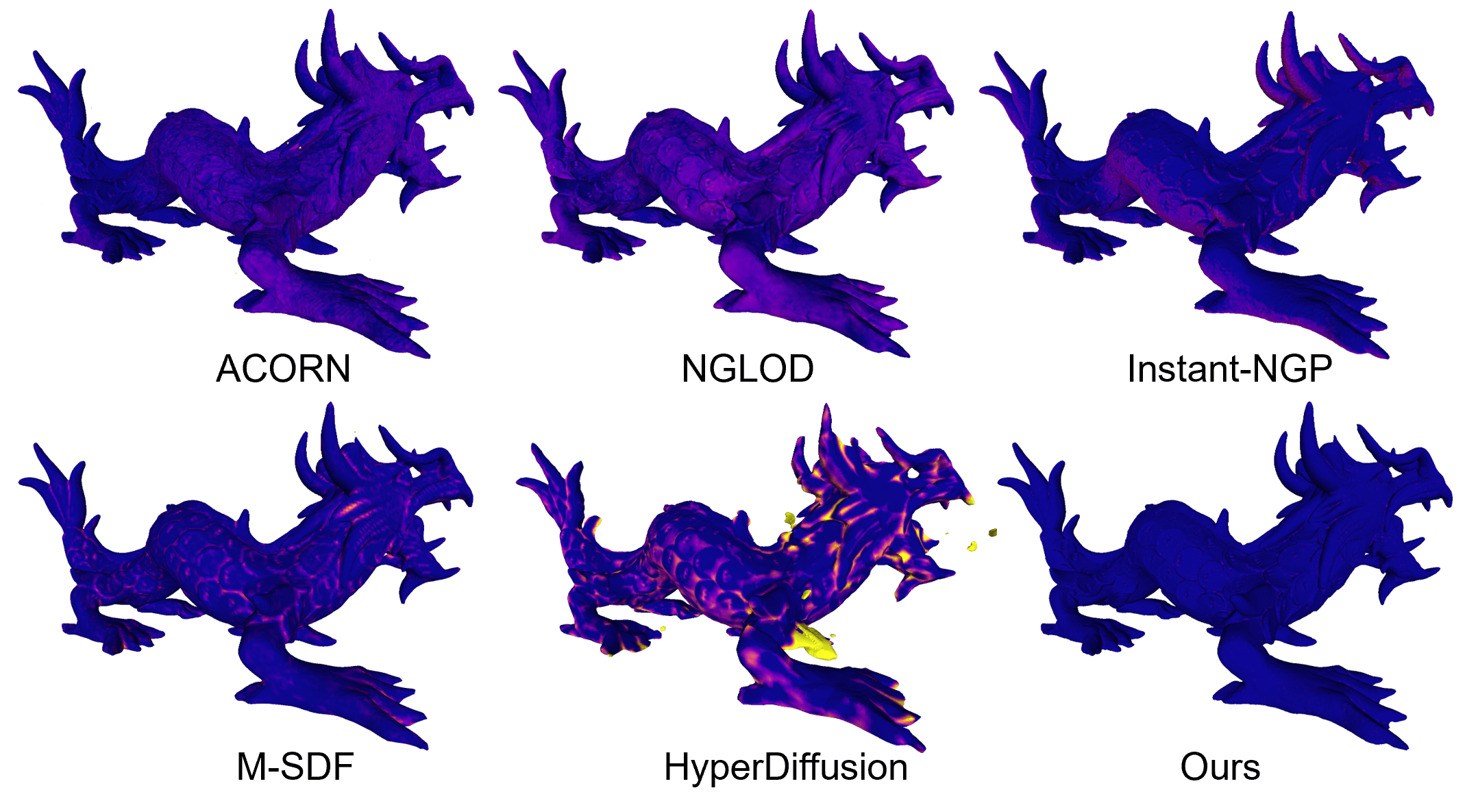} 
  \caption{Error map comparison on Dragon.}
  \label{fig:supp_dragon_errormap}
\end{figure*}

\begin{figure*}
  \centering
\includegraphics[width=1.0\linewidth]{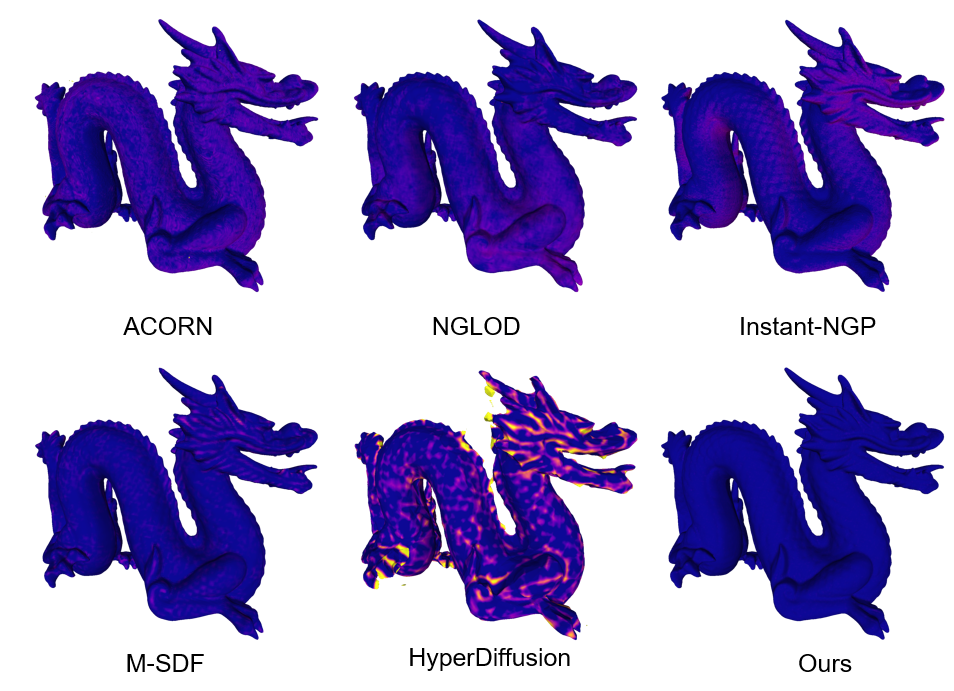} 
  \caption{Error map comparison on Asian Dragon.}
  \label{fig:supp_asian_dragon_errormap}
\end{figure*}

To quantitatively evaluate the reconstruction quality, we report metrics including CD,  F-Score, precision, and recall. As shown in Table~\ref{tab:supp_stanford_quantitative_comparison}, our method consistently outperforms baselines across all metrics.
Overall, our method adapts well to varying surface complexities and delivers high-quality reconstructions.

\subsection{Multiple 3D Objects Reconstruction}
We provide additional reconstruction results on the ShapeNet dataset in Fig.~\ref{fig:multi-shapes-reconstruction-results}, further demonstrating the superiority of our method in preserving fine geometric details.

\begin{figure*}
  \centering
  \begin{tabular}{@{}c@{}c@{}c@{}c@{}c@{}c@{}c@{}c@{}}

     \scalebox{0.9}{DeepSDF} & \scalebox{0.9}{IF-Net} & \scalebox{0.9}{3{DILG}} &  \scalebox{0.9}{Instant-NGP}&\scalebox{0.9}{M-SDF} & 
     \scalebox{0.9}{HyperDiffusion} & \scalebox{0.9}{Ours} & \scalebox{0.9}{GT}  \\
  
\includegraphics[width=\imgwidth]{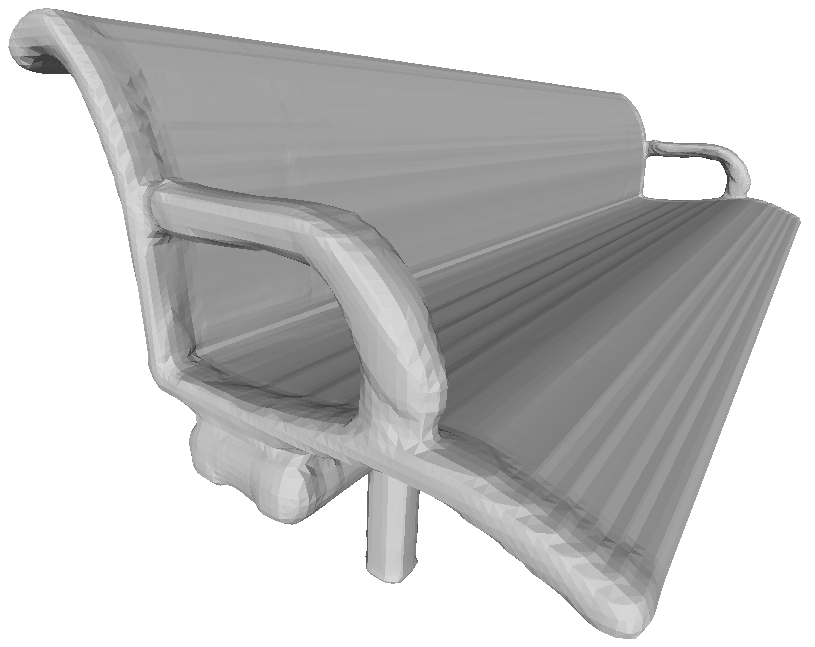} &
\includegraphics[width=\imgwidth]{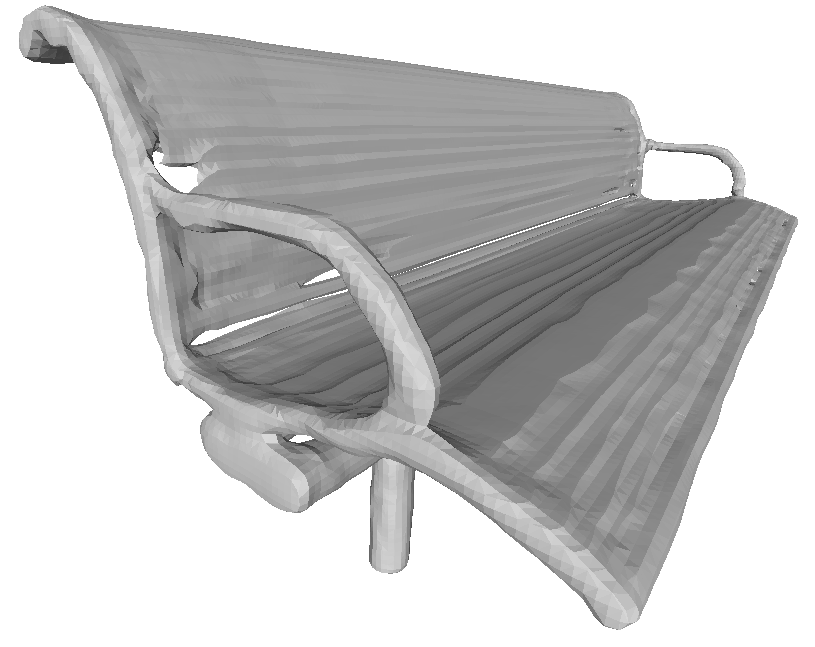} &
\includegraphics[width=\imgwidth]{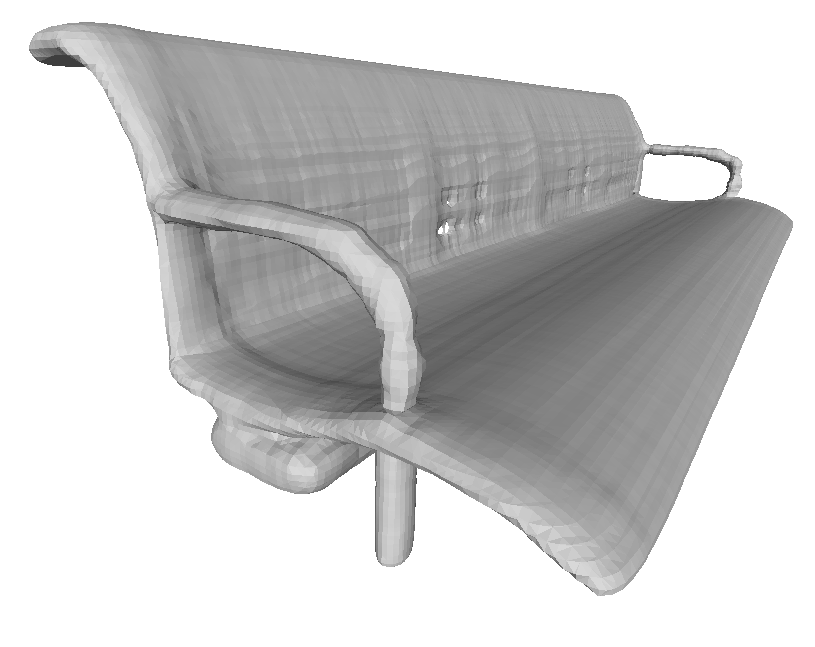} &
\includegraphics[width=\imgwidth]{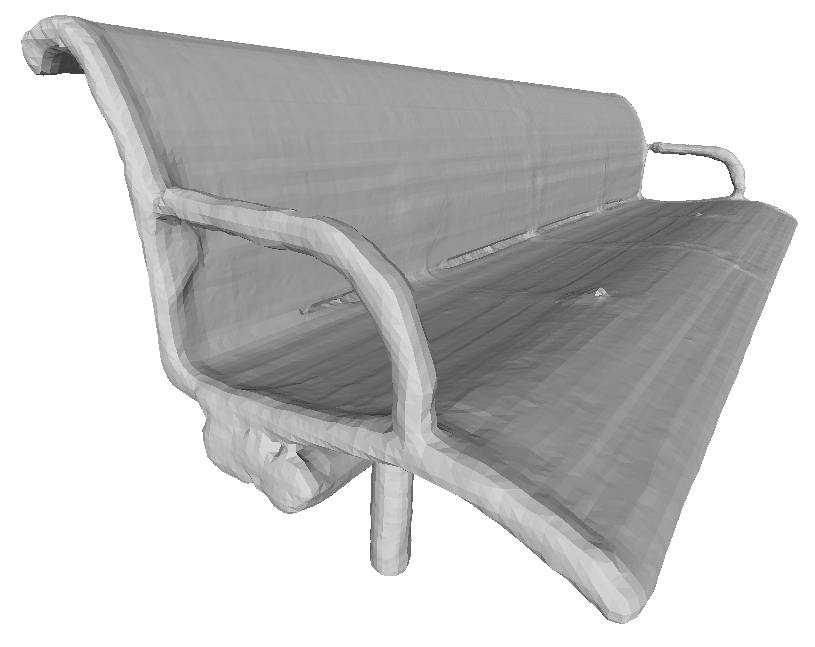} &
\includegraphics[width=\imgwidth]{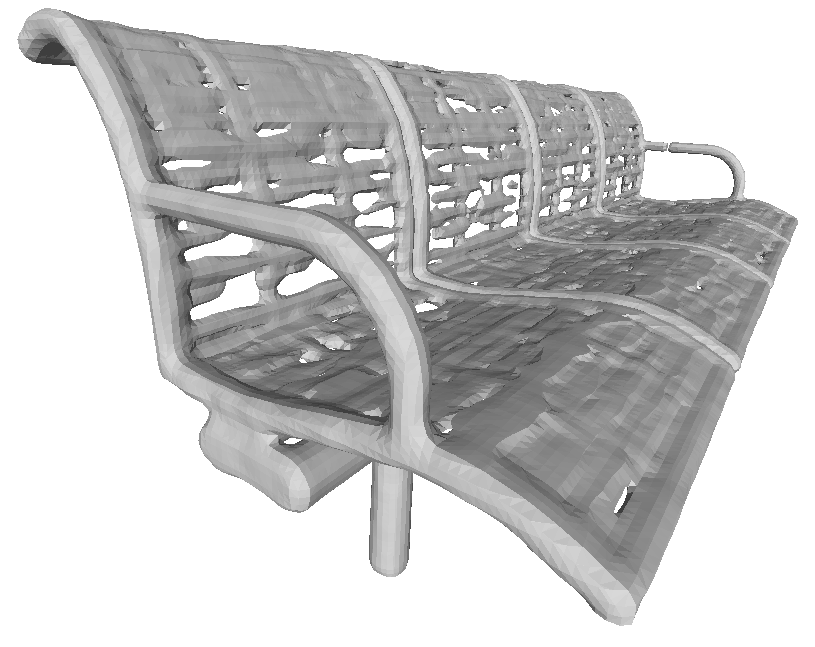} &
\includegraphics[width=\imgwidth]{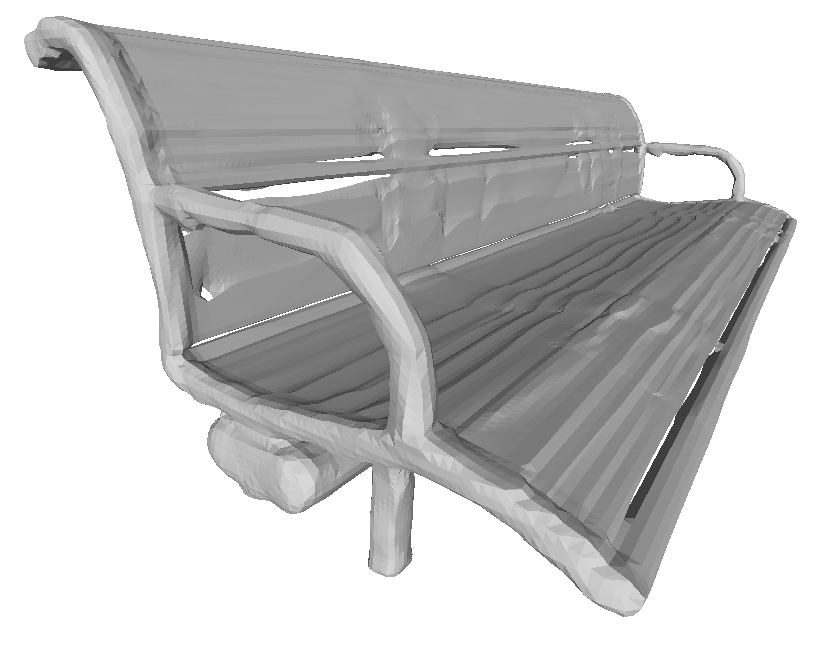} &
\includegraphics[width=\imgwidth]{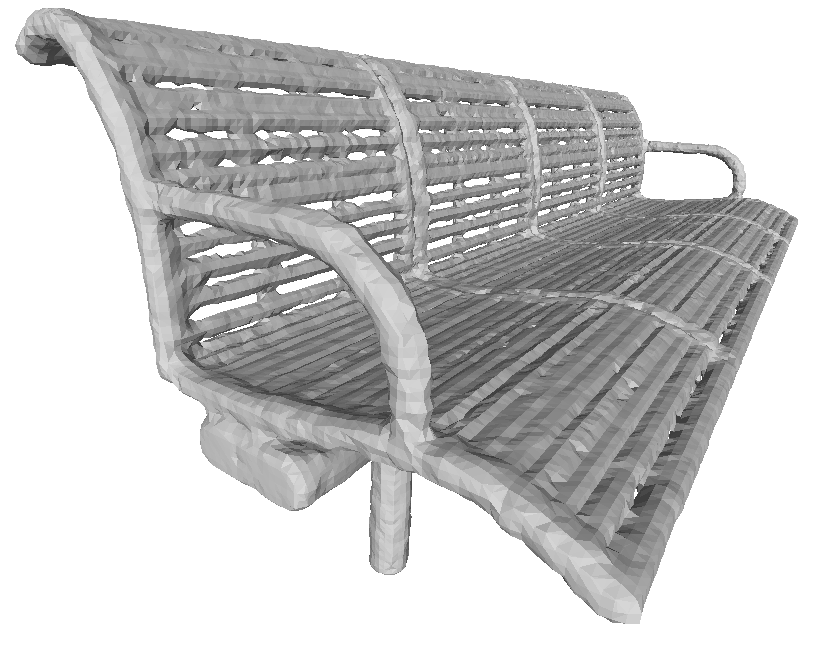} &
\includegraphics[width=\imgwidth]{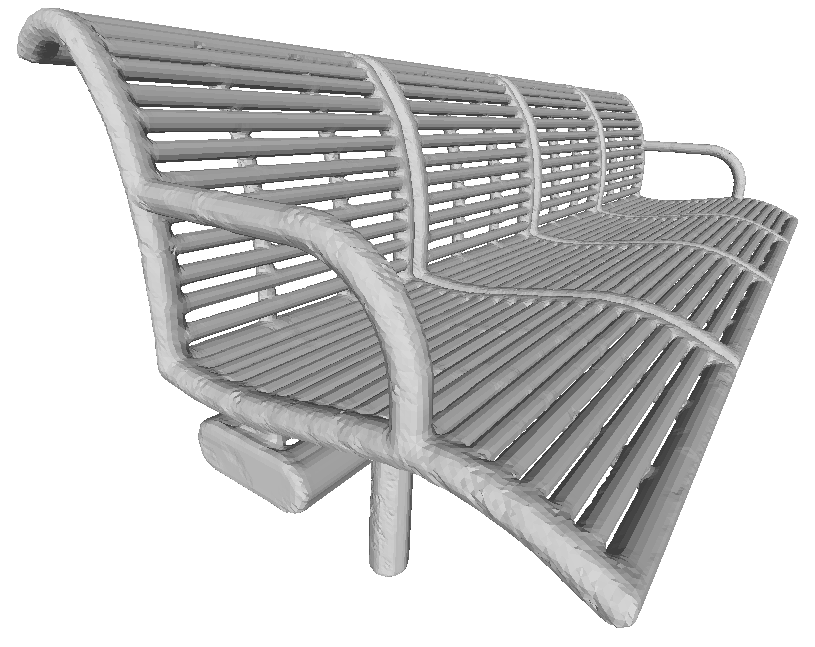} \\

\includegraphics[width=\imgwidth, height=3cm, keepaspectratio]{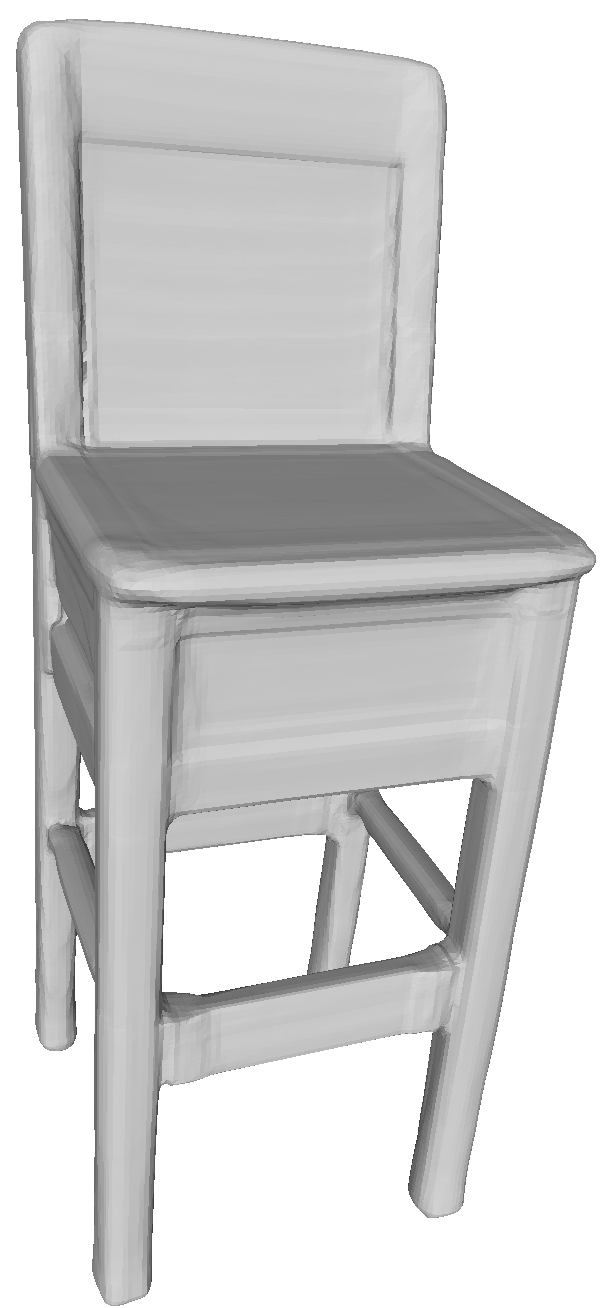} &
\includegraphics[width=\imgwidth, height=3cm, keepaspectratio]{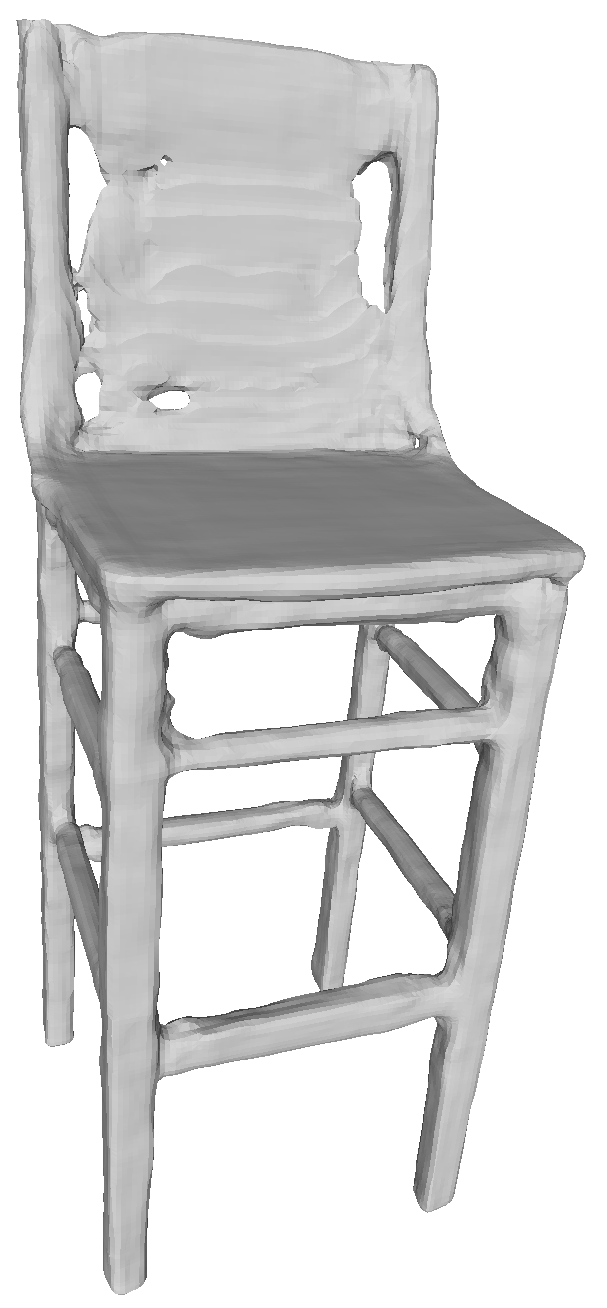} &
\includegraphics[width=\imgwidth, height=3cm, keepaspectratio]{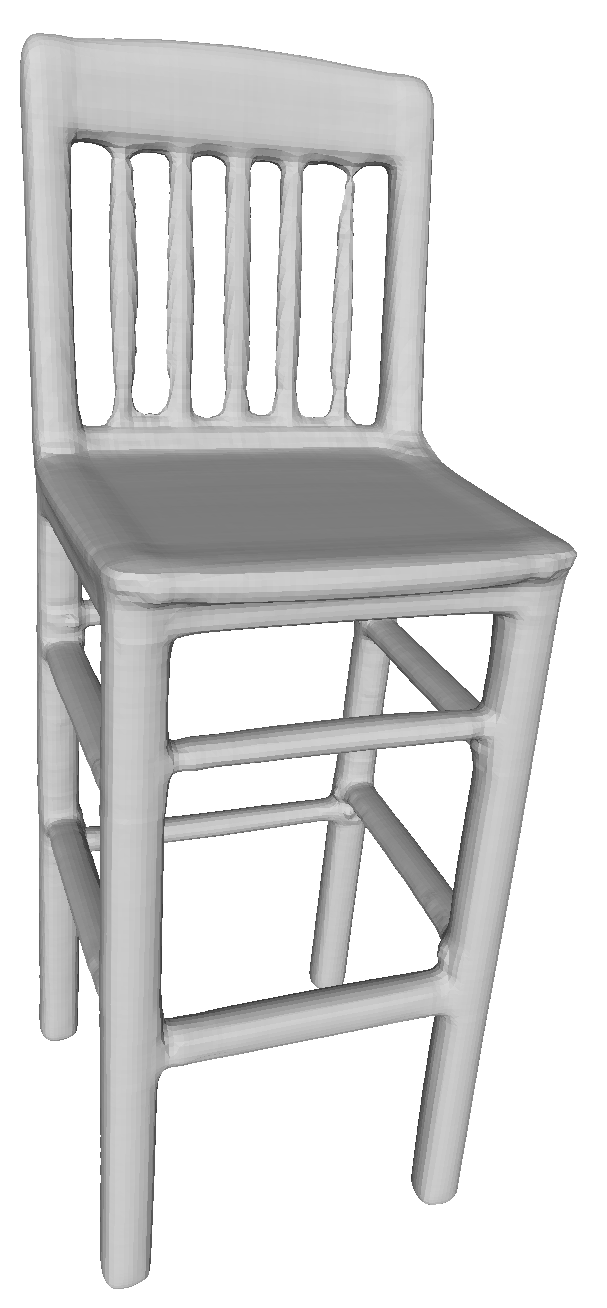} &
\includegraphics[width=\imgwidth, height=3cm, keepaspectratio]{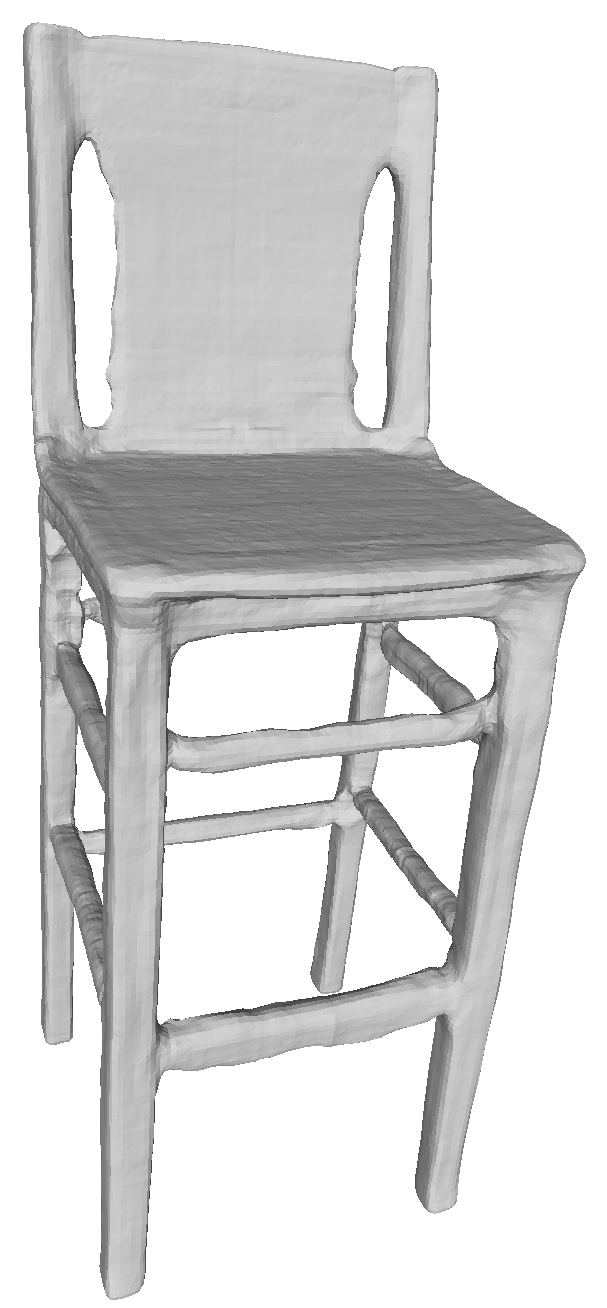} &
\includegraphics[width=\imgwidth, height=3cm, keepaspectratio]{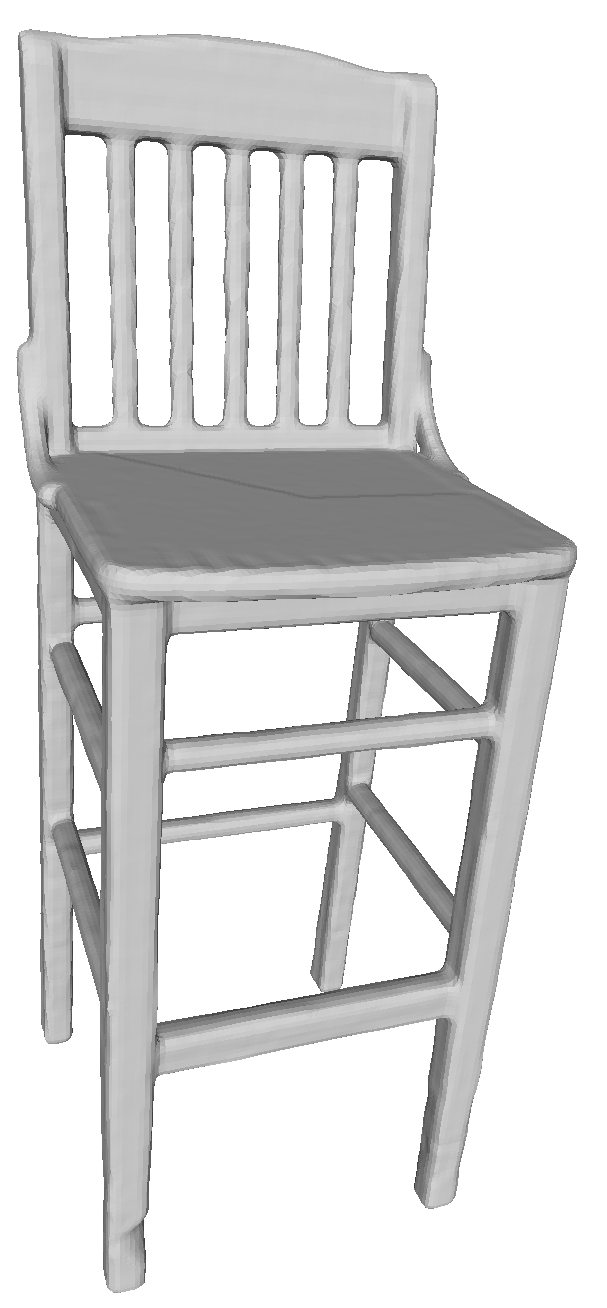} &
\includegraphics[width=\imgwidth, height=3cm, keepaspectratio]{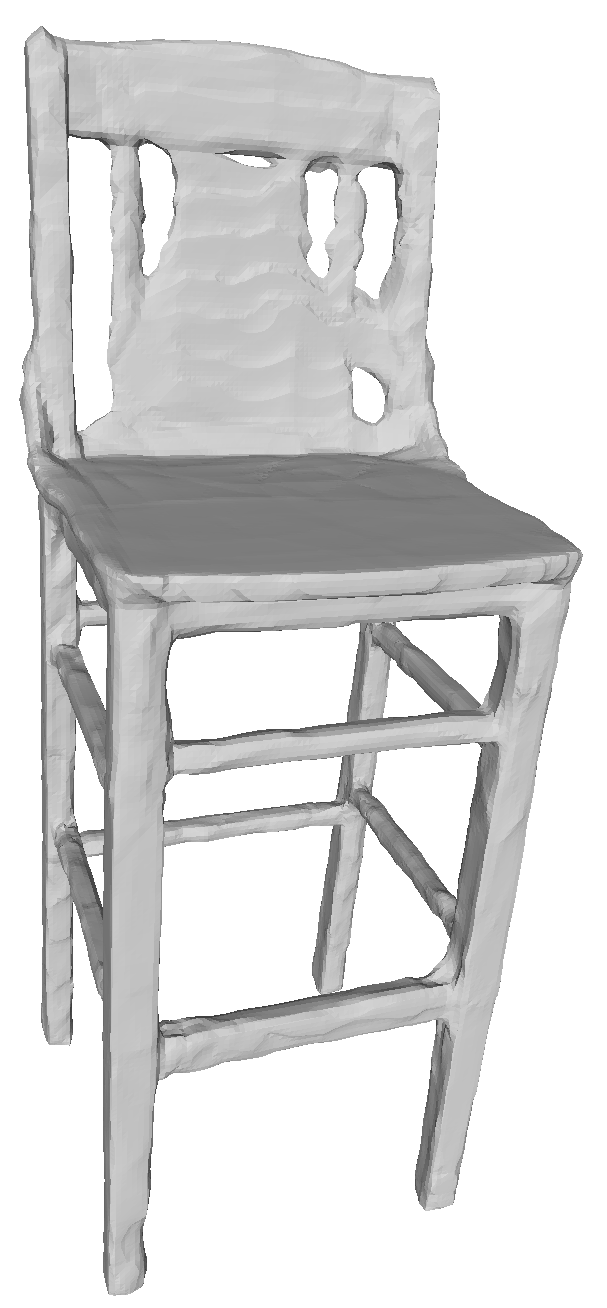} &
\includegraphics[width=\imgwidth, height=3cm, keepaspectratio]{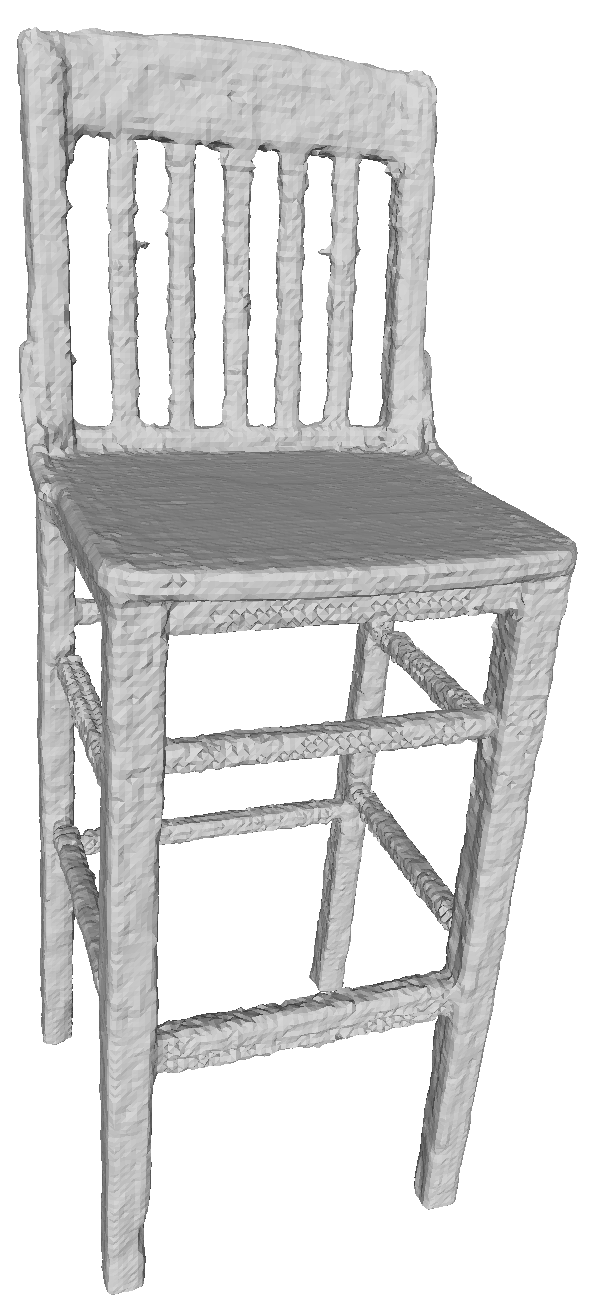} &
\includegraphics[width=\imgwidth, height=3cm, keepaspectratio]{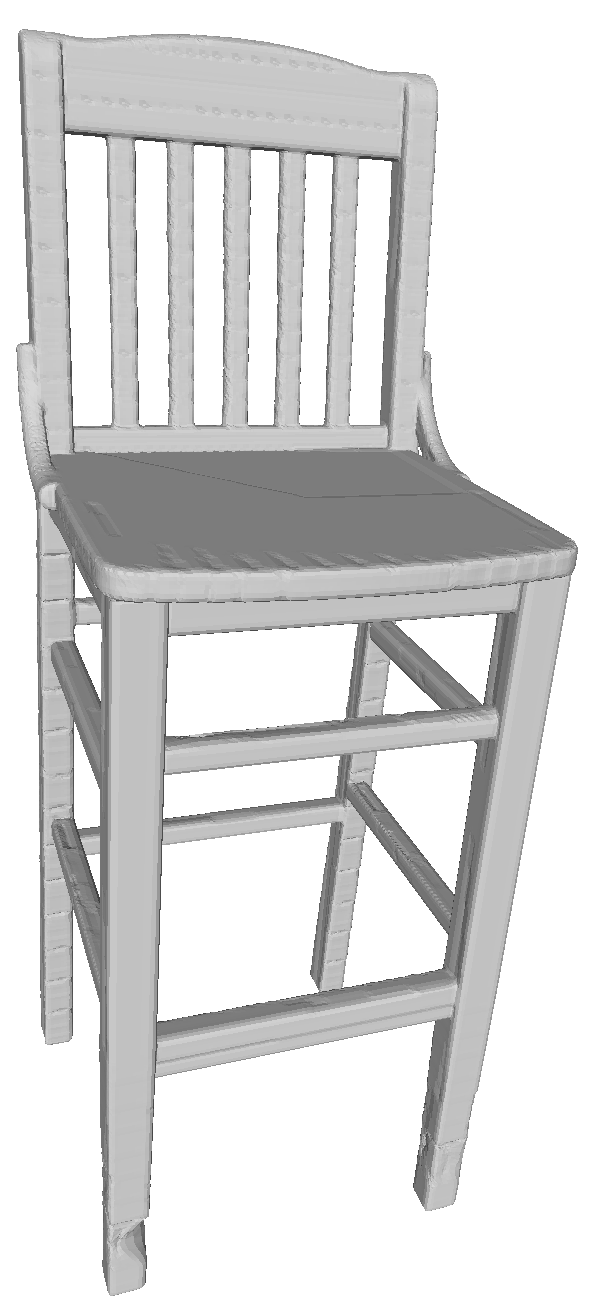} \\

\includegraphics[width=\imgwidth, height=3cm, keepaspectratio]{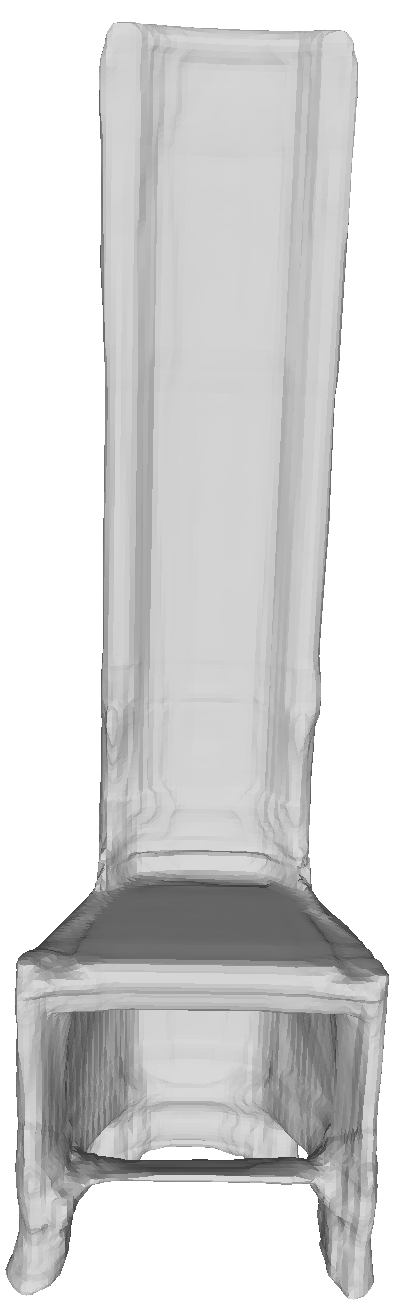} &
\includegraphics[width=\imgwidth, height=3cm, keepaspectratio]{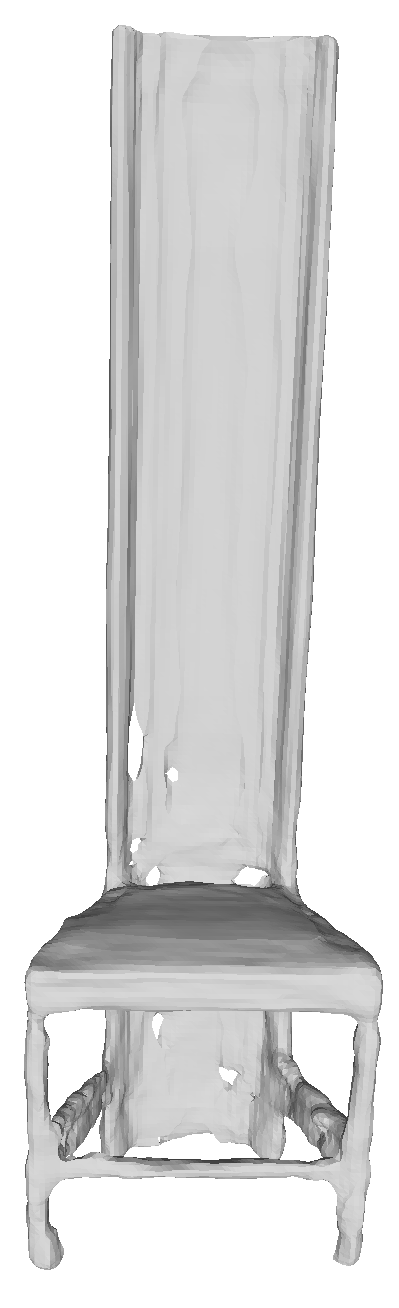} &
\includegraphics[width=\imgwidth, height=3cm, keepaspectratio]{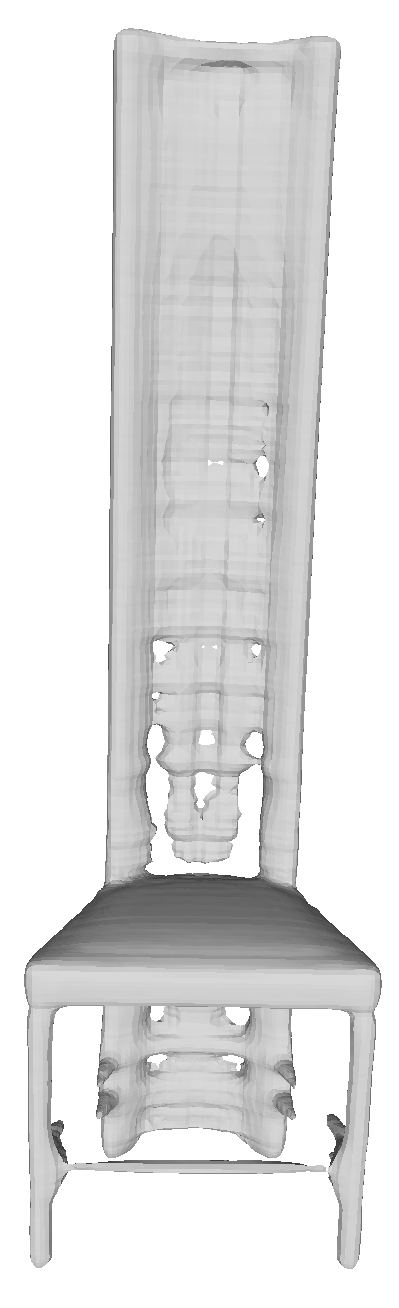} &
\includegraphics[width=\imgwidth, height=3cm, keepaspectratio]{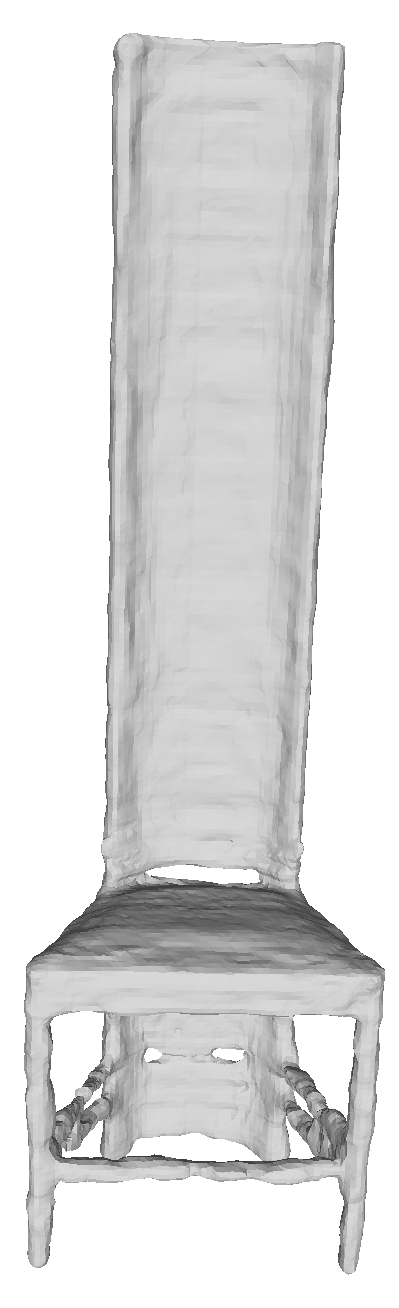} &
\includegraphics[width=\imgwidth, height=3cm, keepaspectratio]{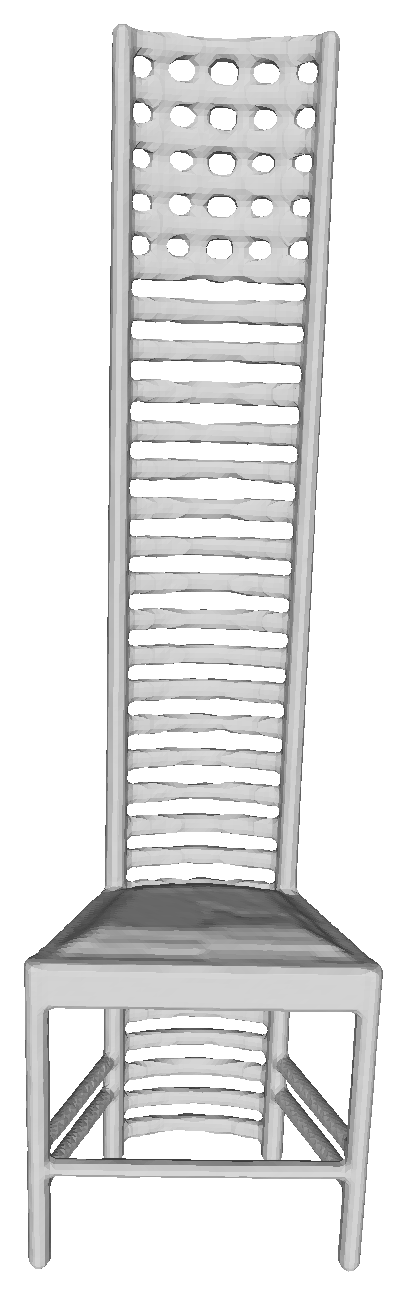} &
\includegraphics[width=\imgwidth, height=3cm, keepaspectratio]{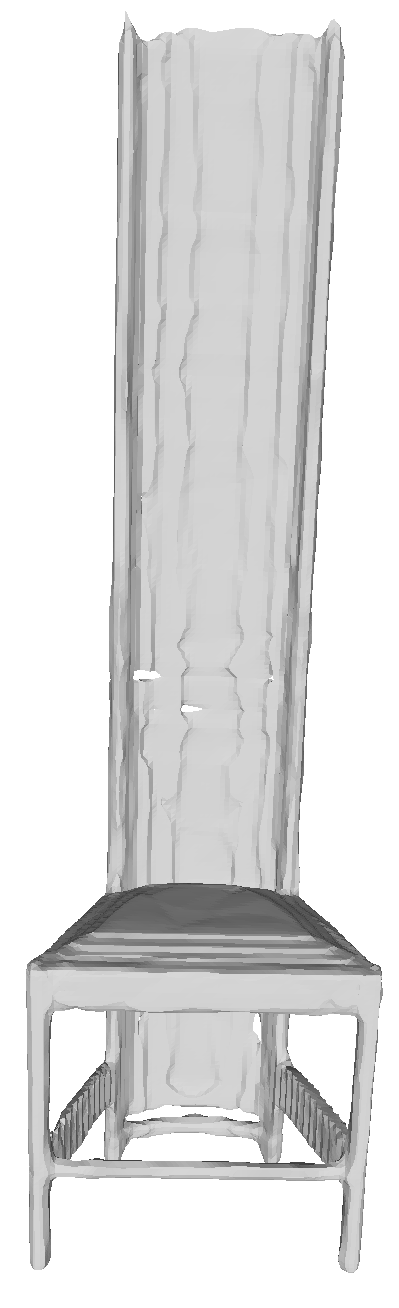} &
\includegraphics[width=\imgwidth, height=3cm, keepaspectratio]{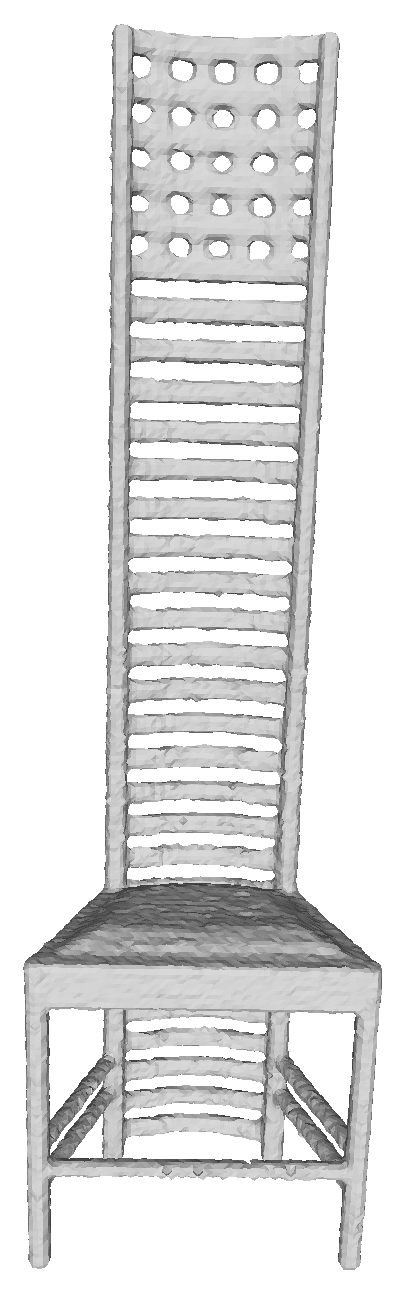} &
\includegraphics[width=\imgwidth, height=3cm, keepaspectratio]{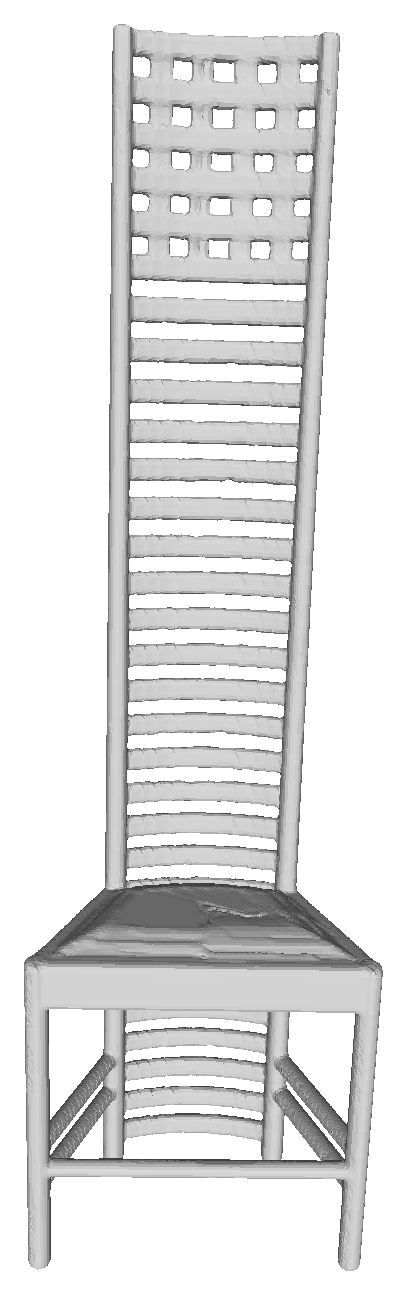} \\

\includegraphics[width=\imgwidth]{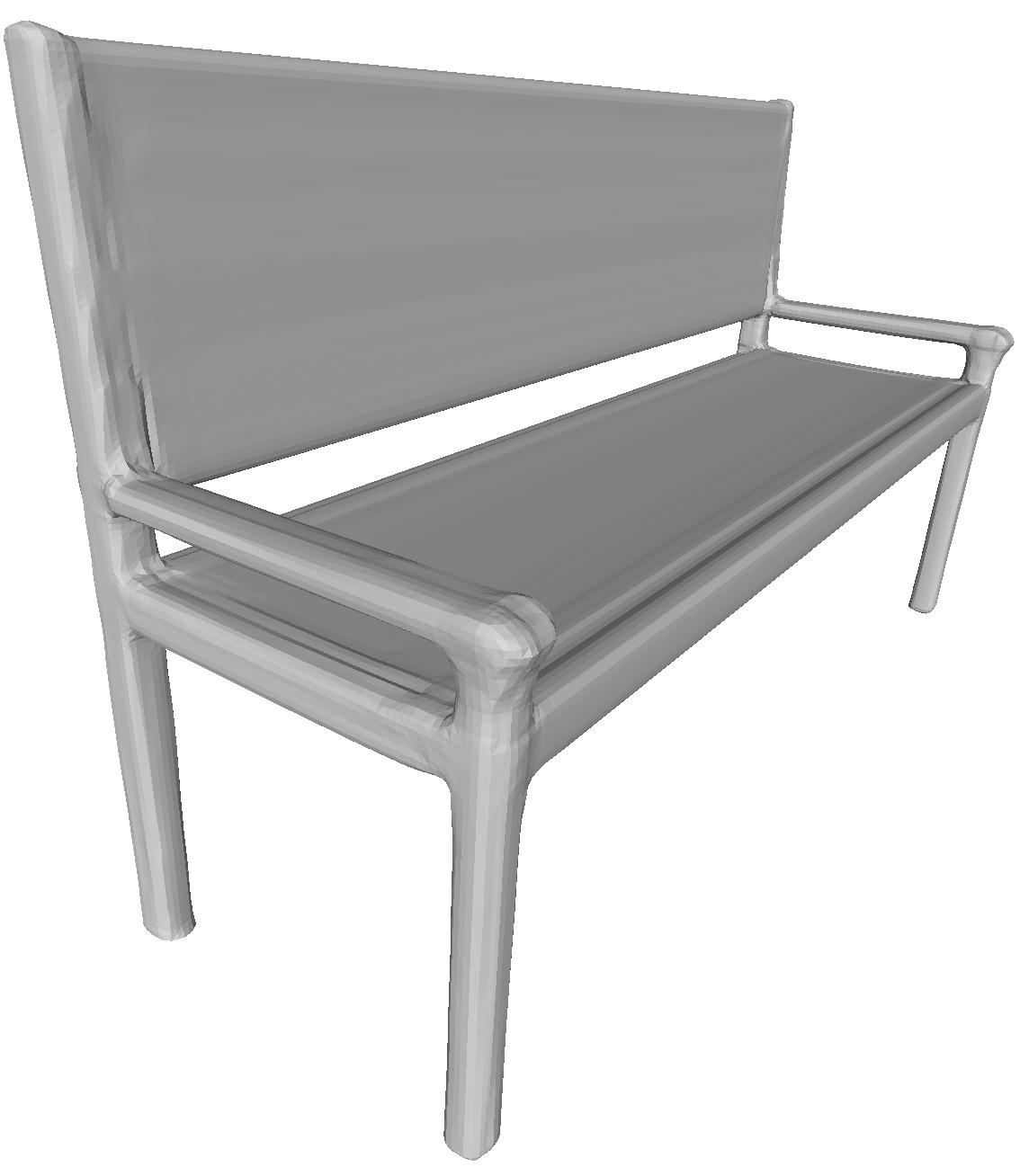} &
\includegraphics[width=\imgwidth]{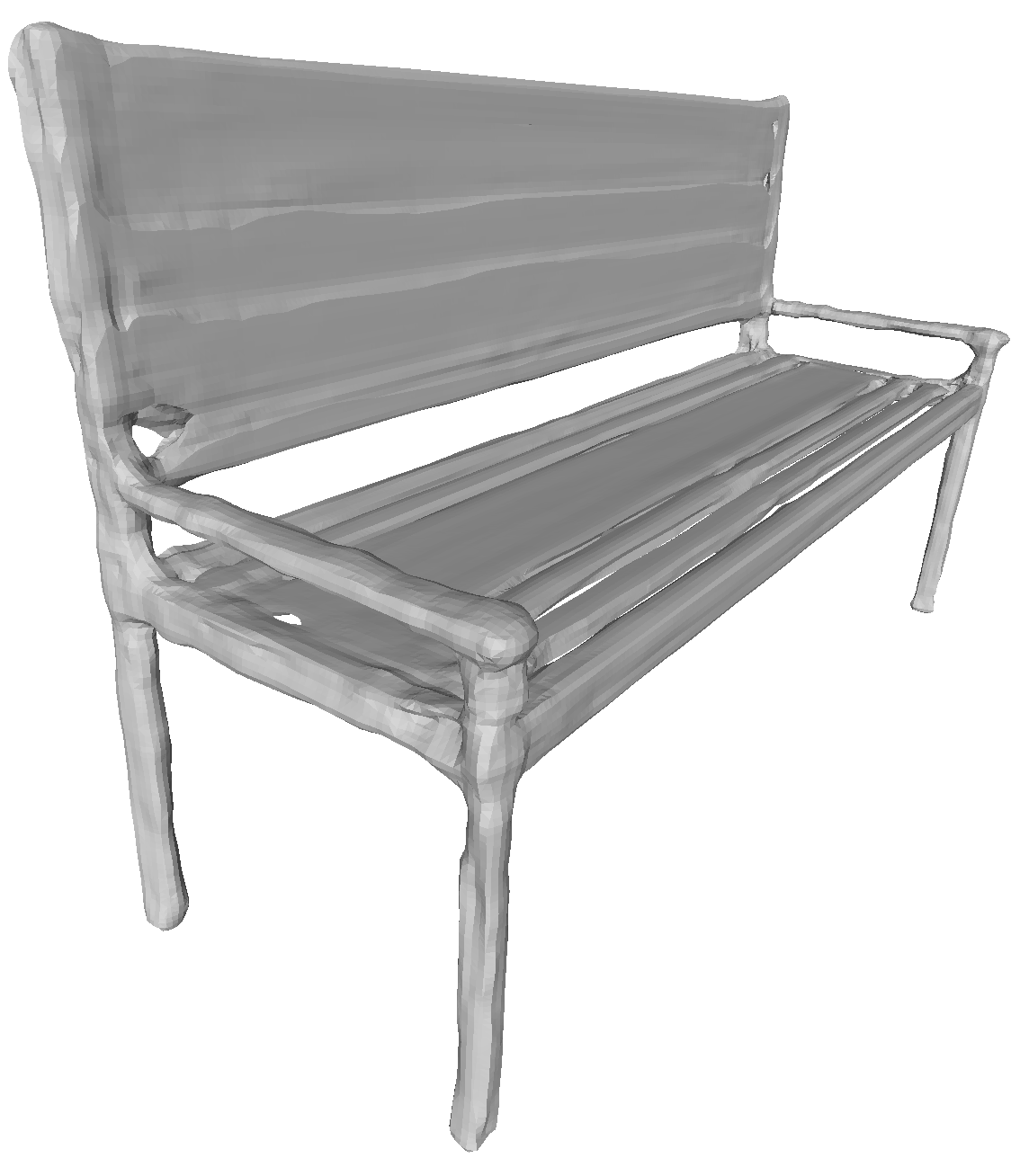} &
\includegraphics[width=\imgwidth]{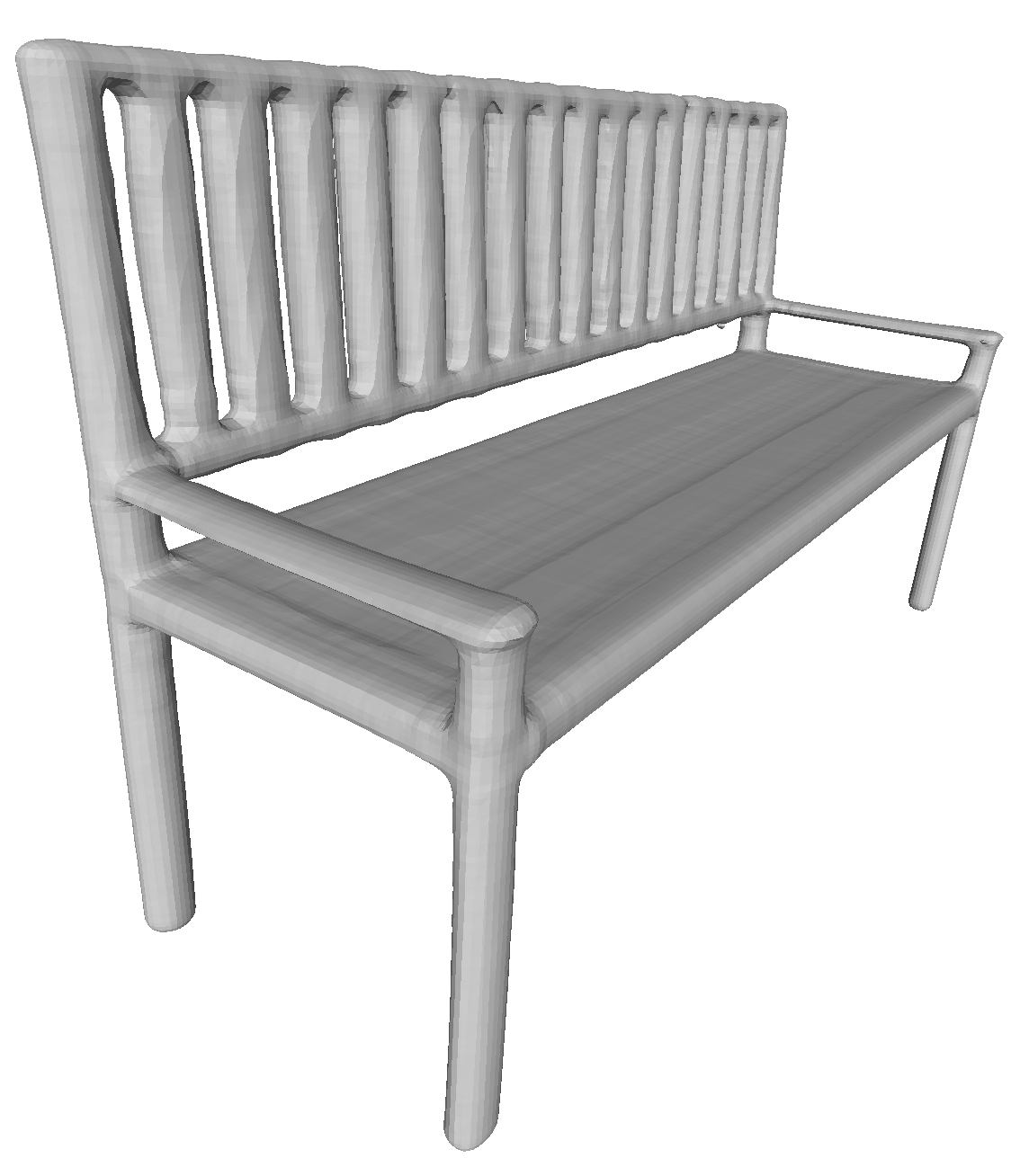} &
\includegraphics[width=\imgwidth]{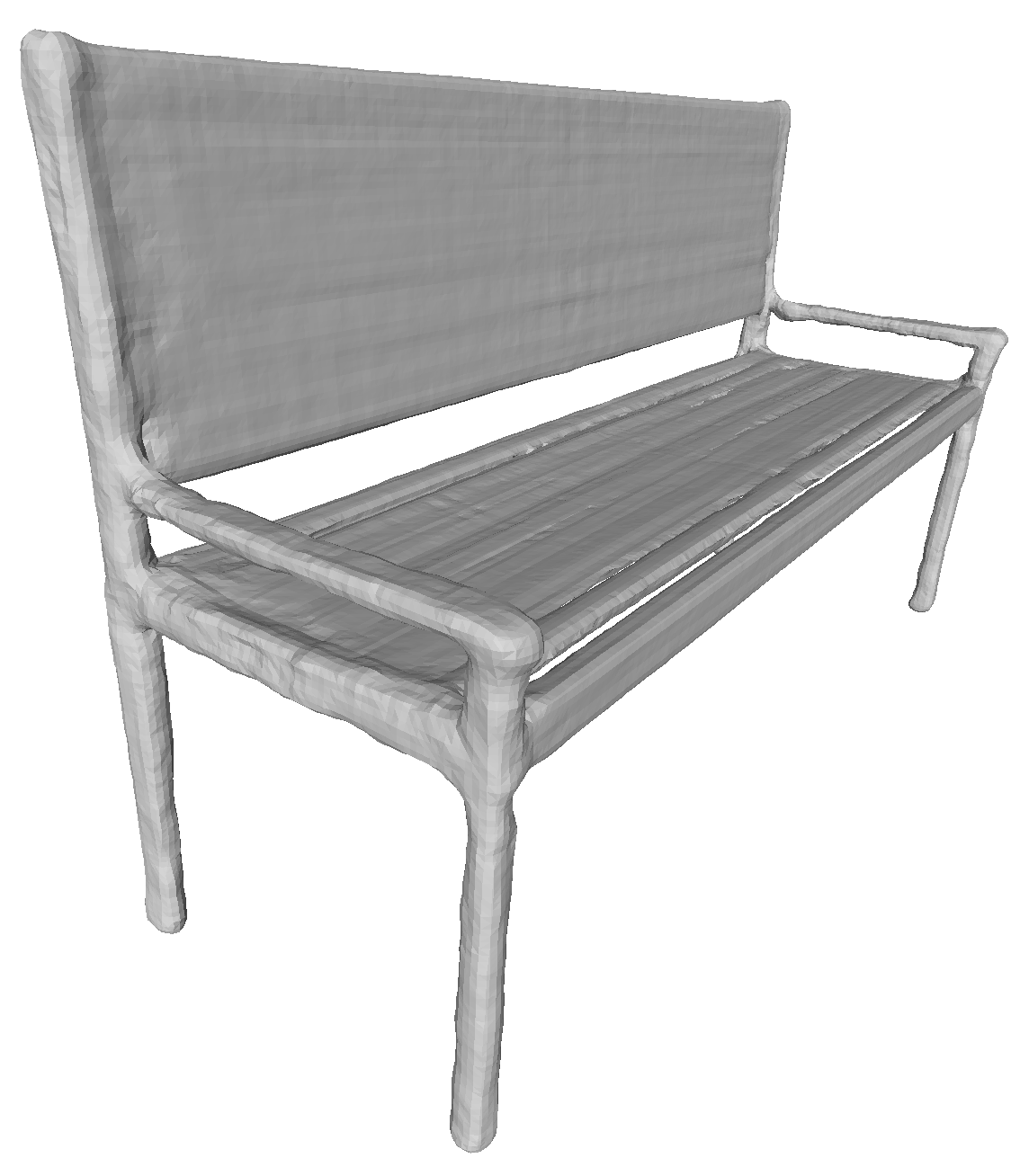} &
\includegraphics[width=\imgwidth]{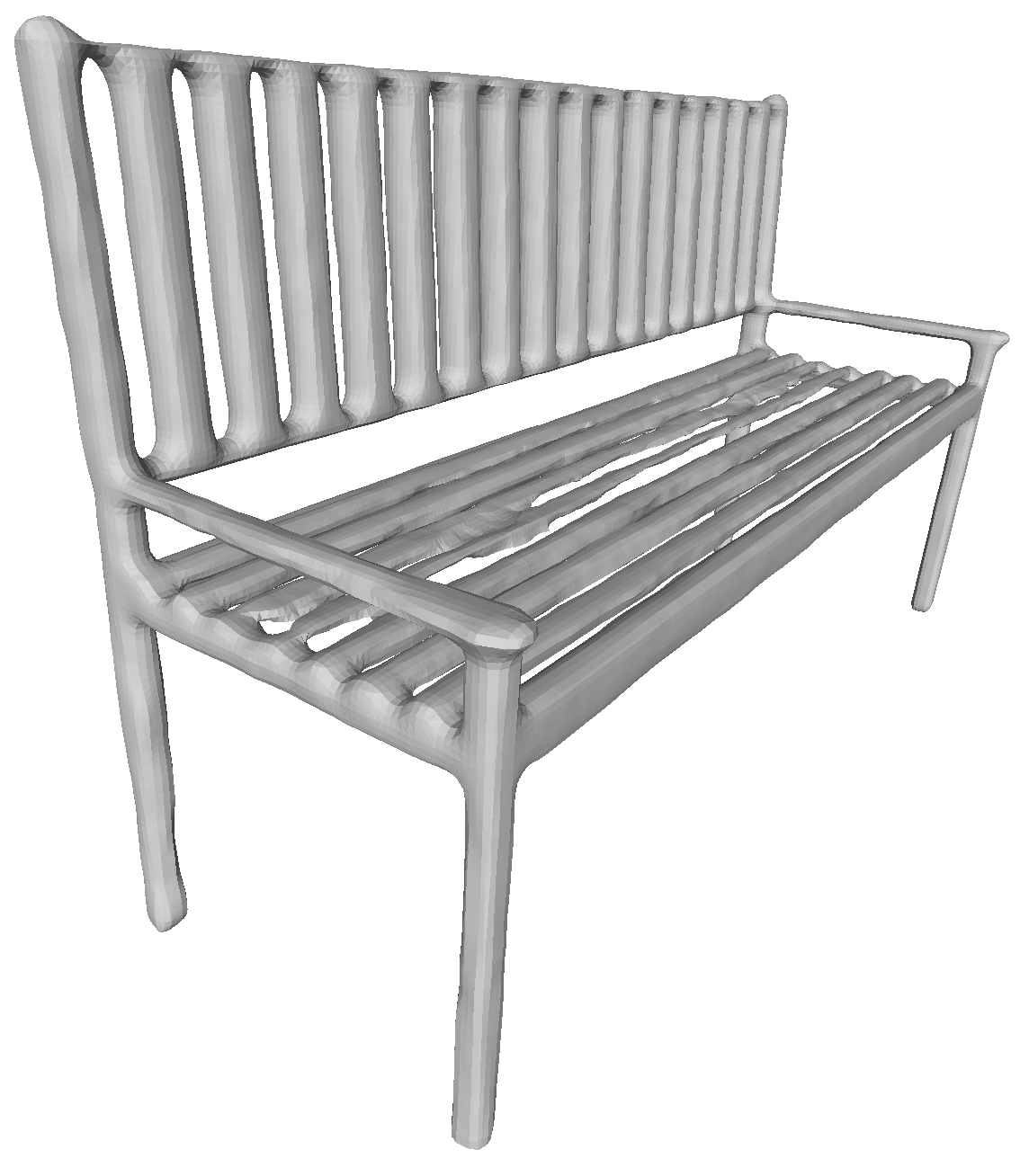} &
\includegraphics[width=\imgwidth]{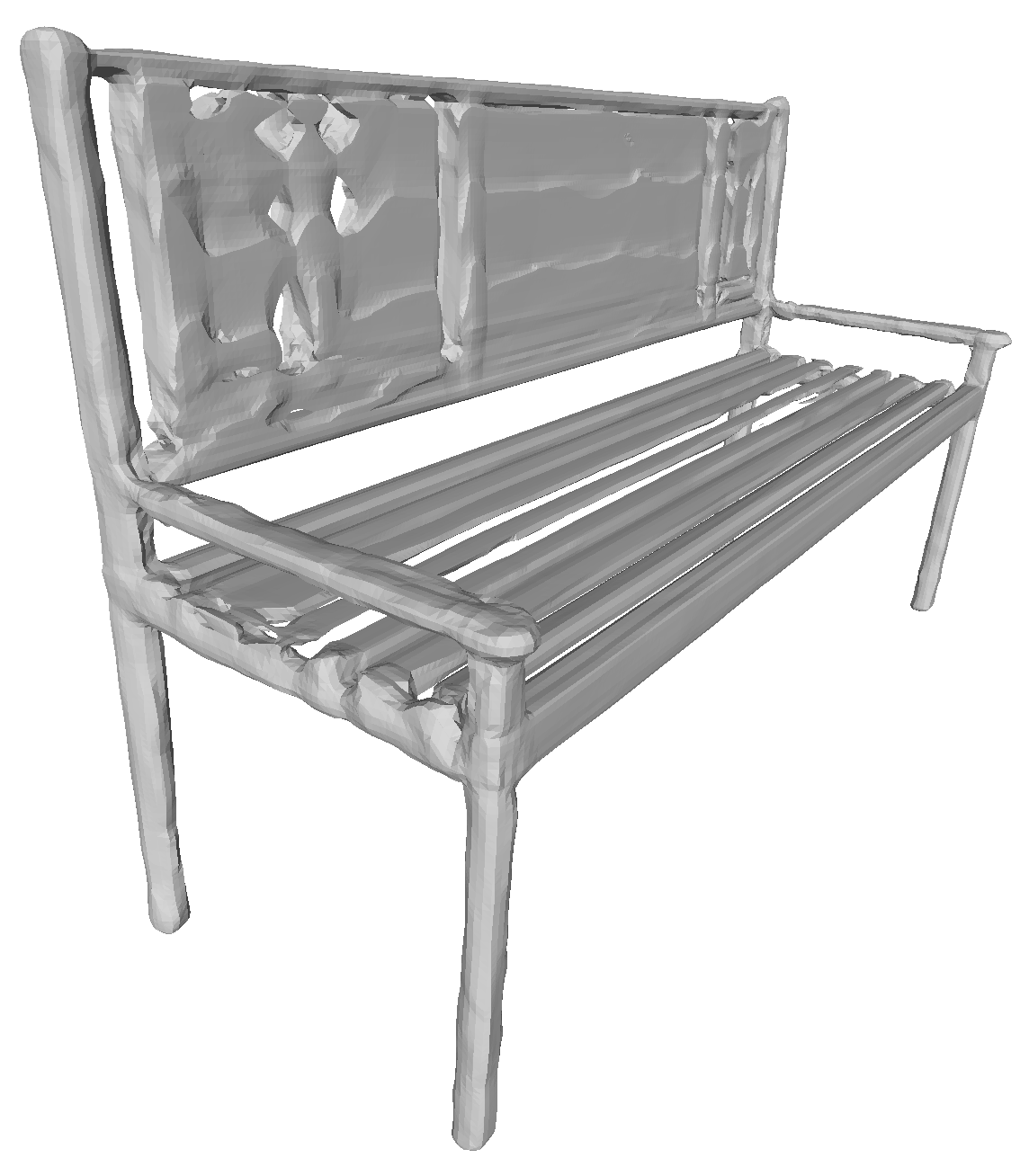} &
\includegraphics[width=\imgwidth]{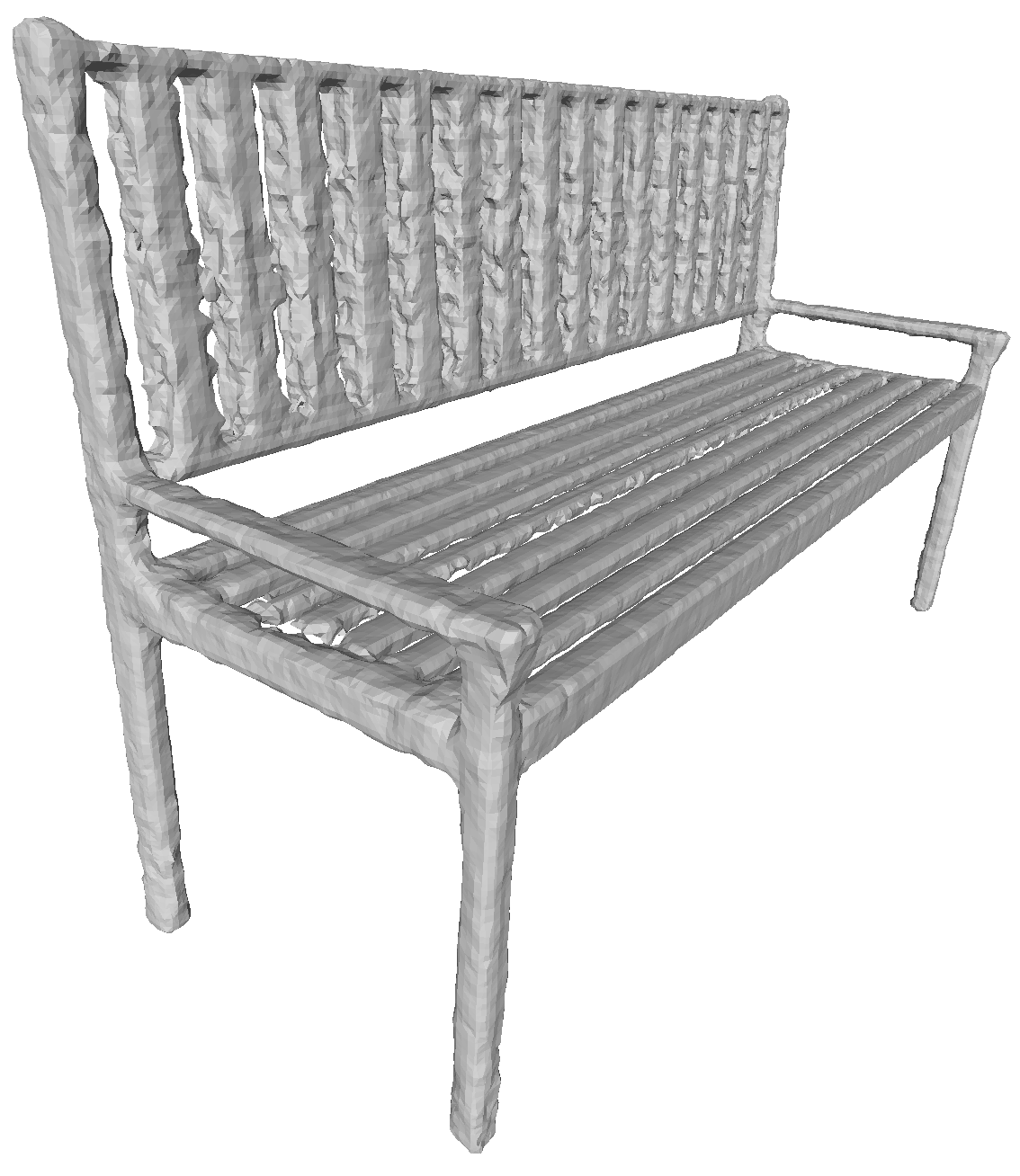} &
\includegraphics[width=\imgwidth]{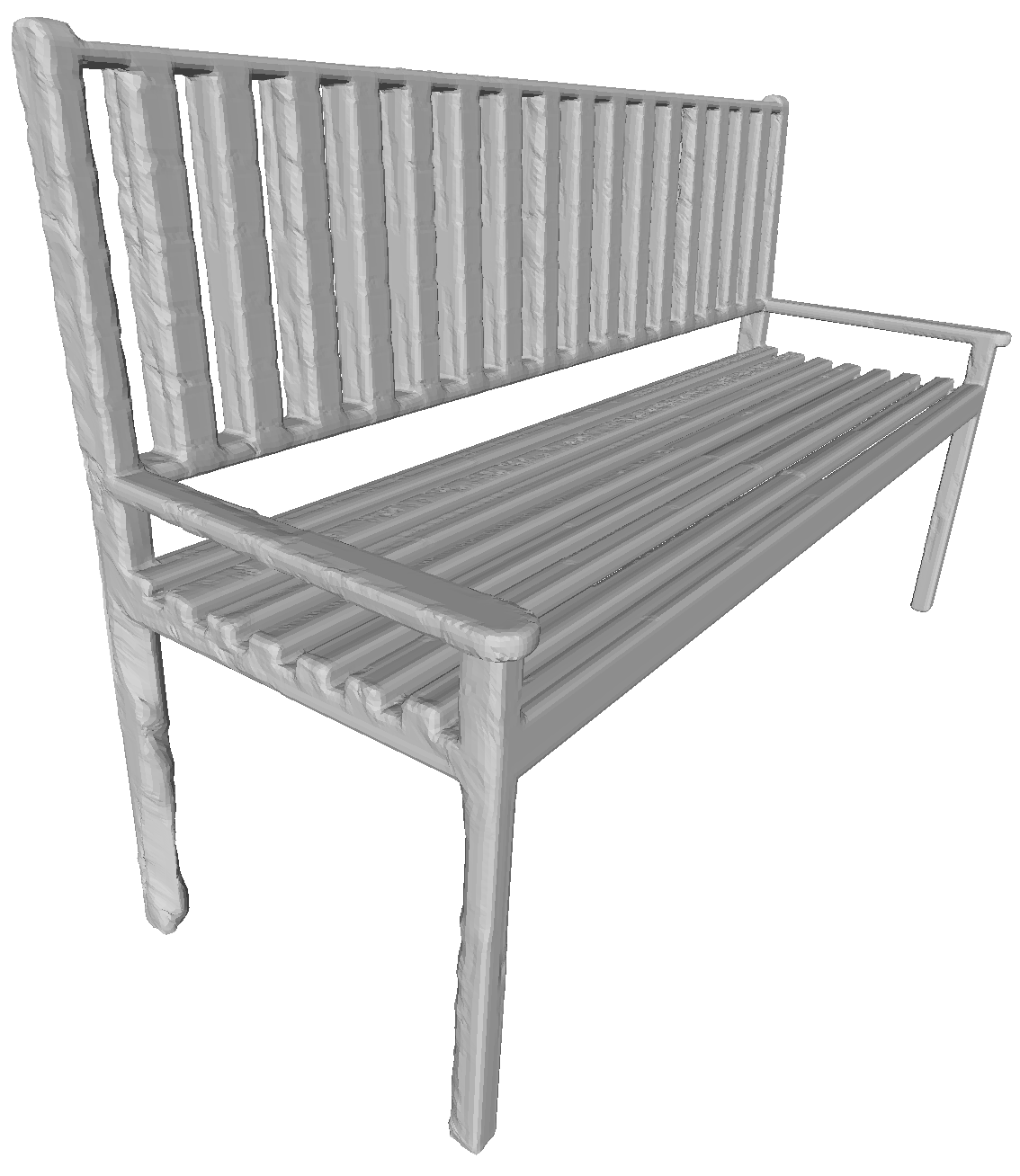} \\

\includegraphics[width=\imgwidth]{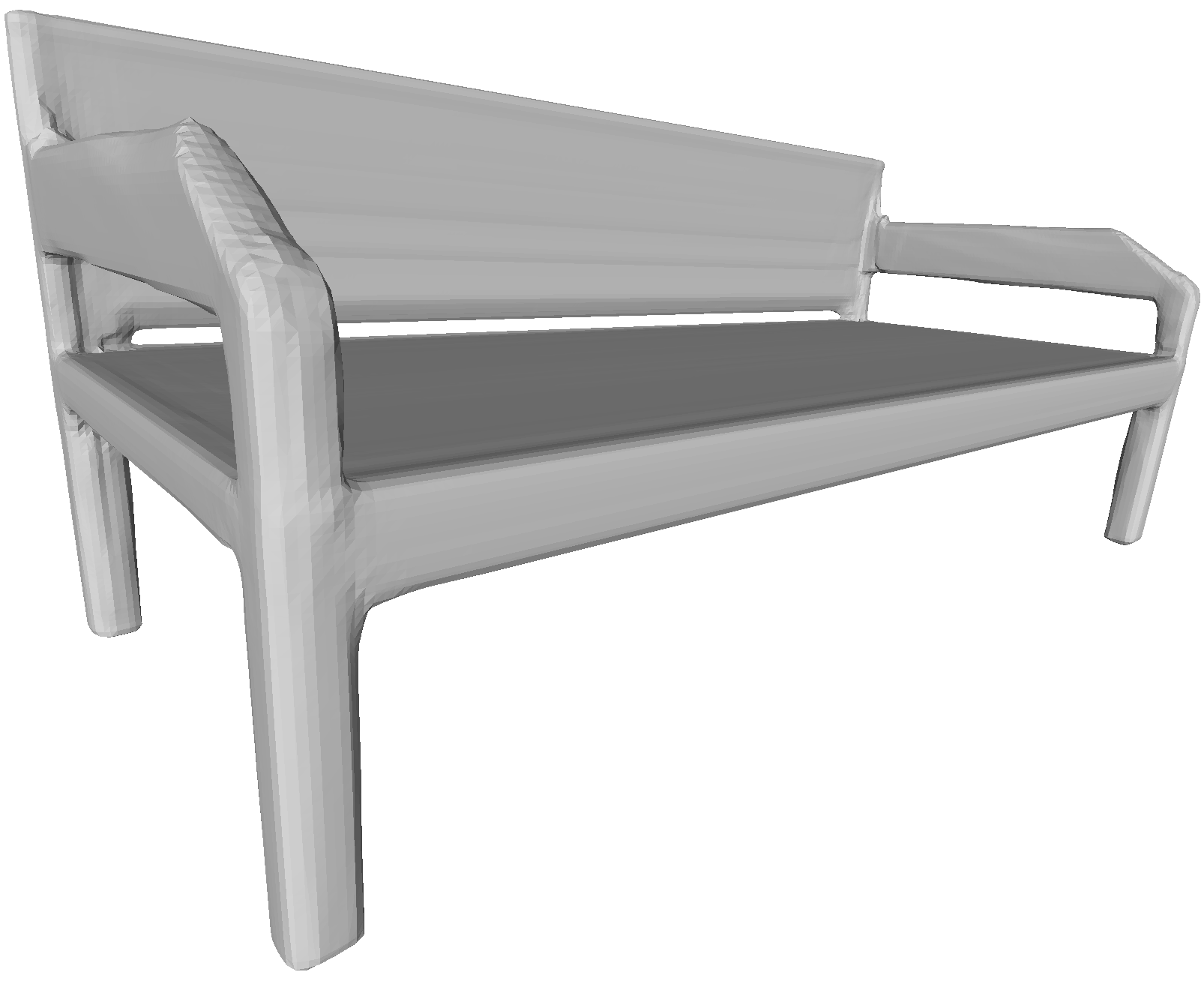} &
\includegraphics[width=\imgwidth]{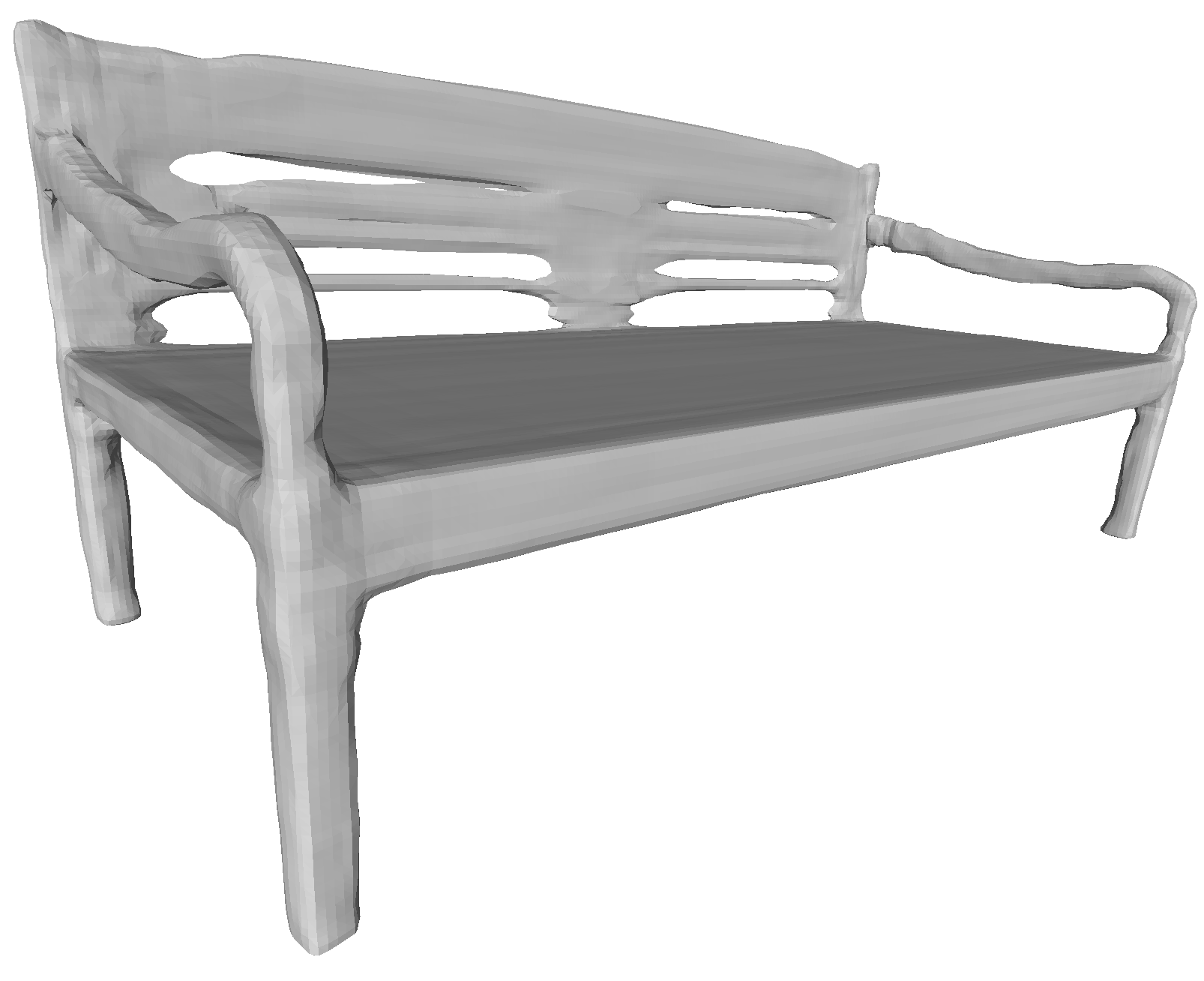} &
\includegraphics[width=\imgwidth]{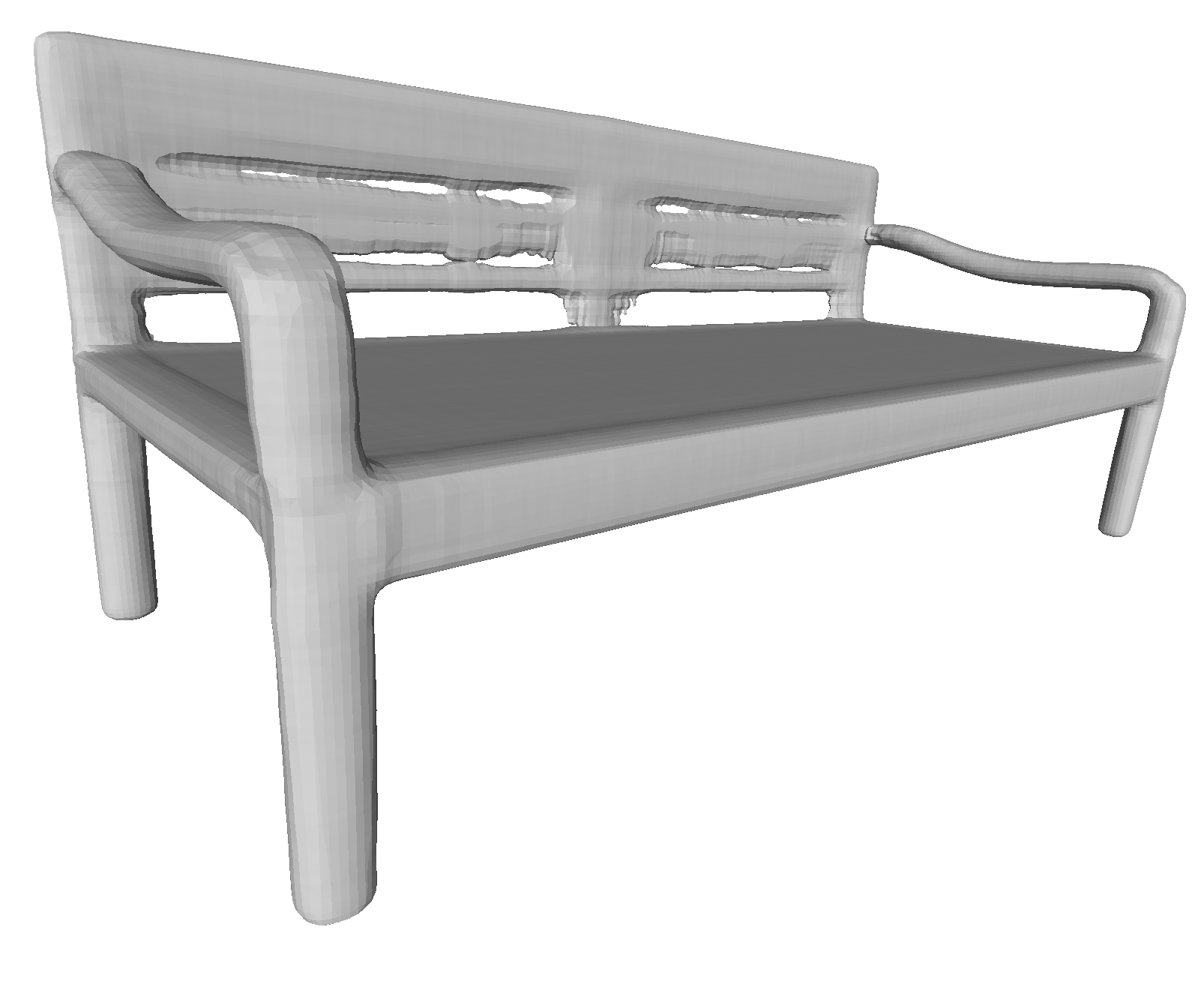} &
\includegraphics[width=\imgwidth]{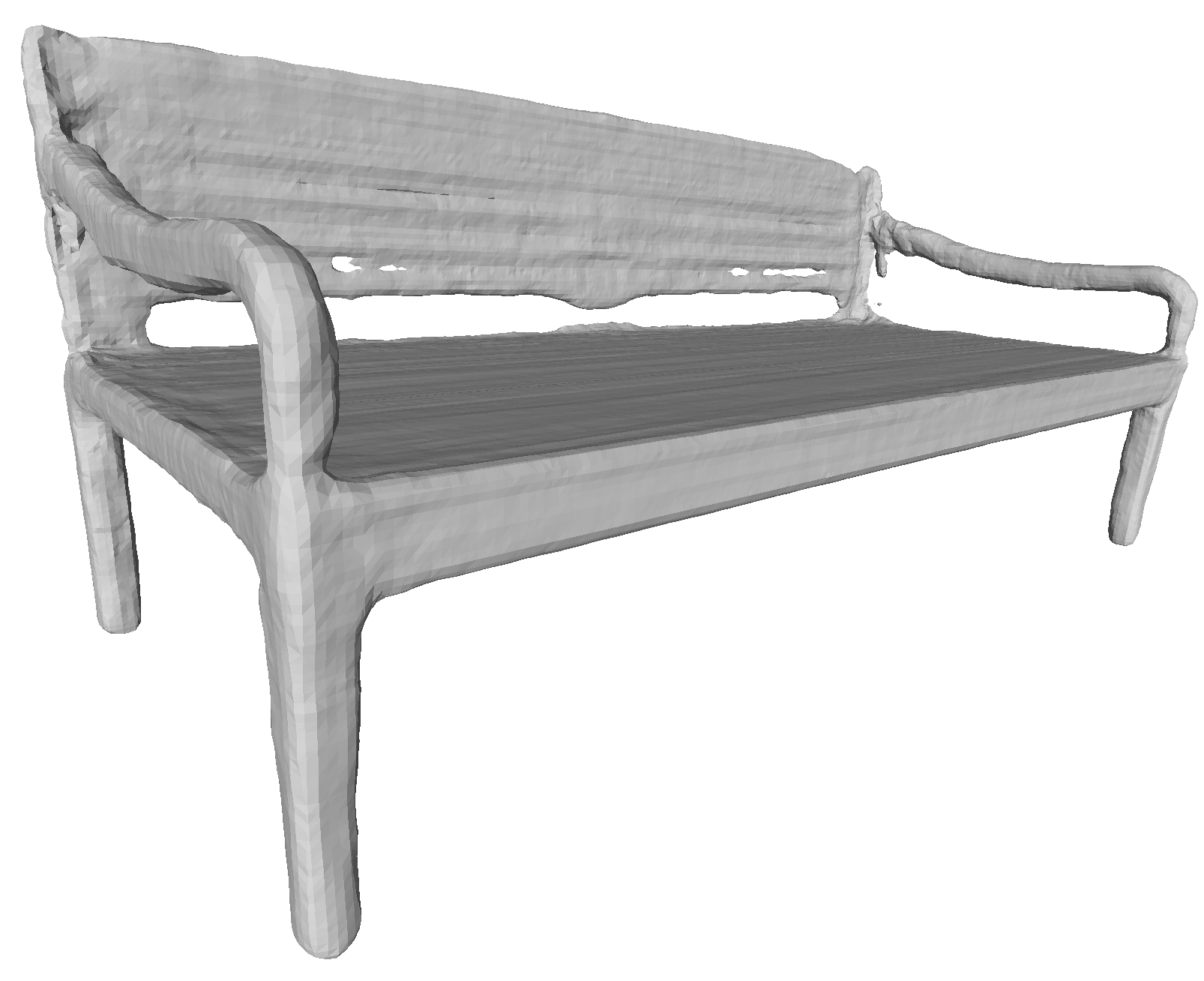} &
\includegraphics[width=\imgwidth]{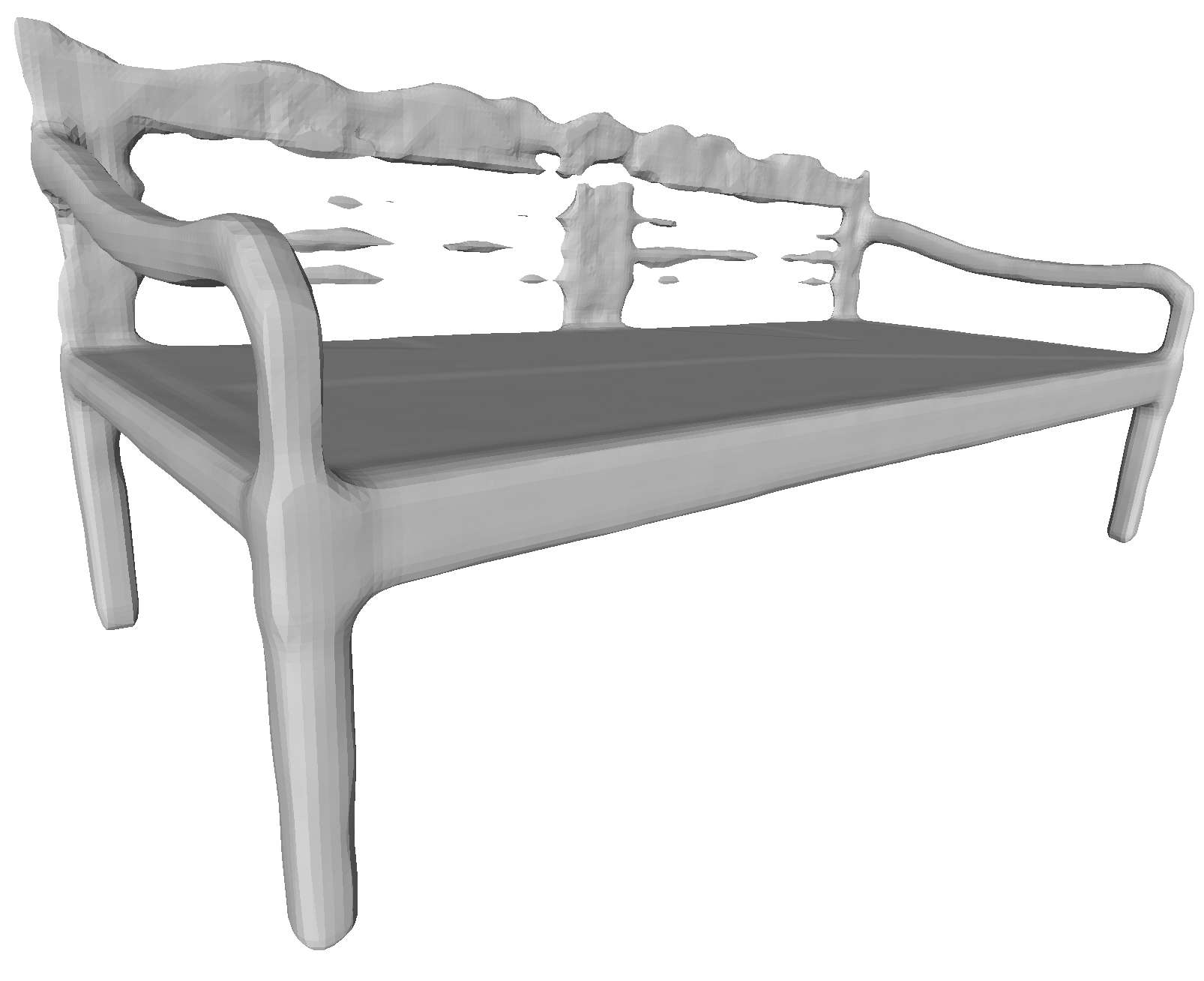} &
\includegraphics[width=\imgwidth]{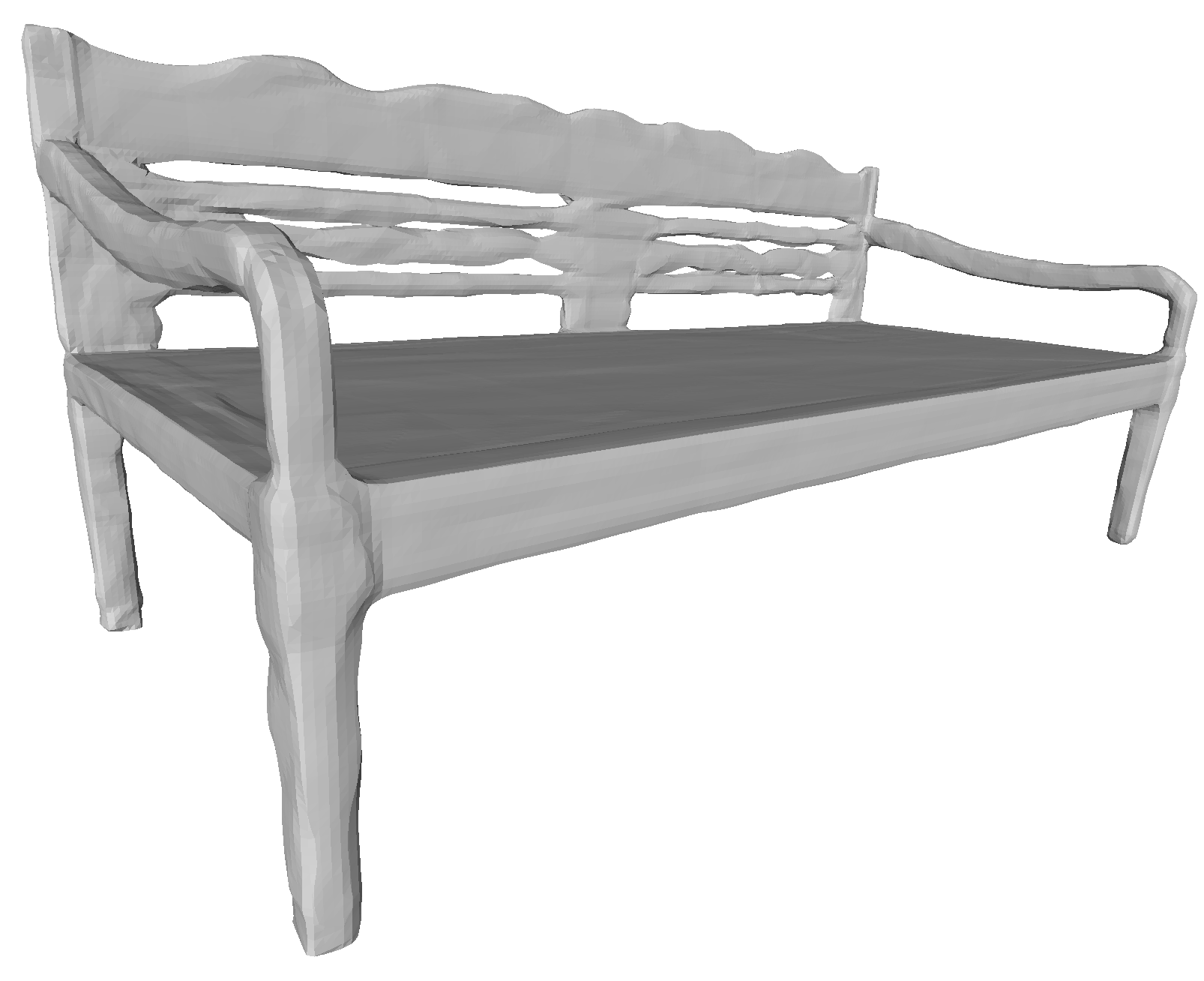} &
\includegraphics[width=\imgwidth]{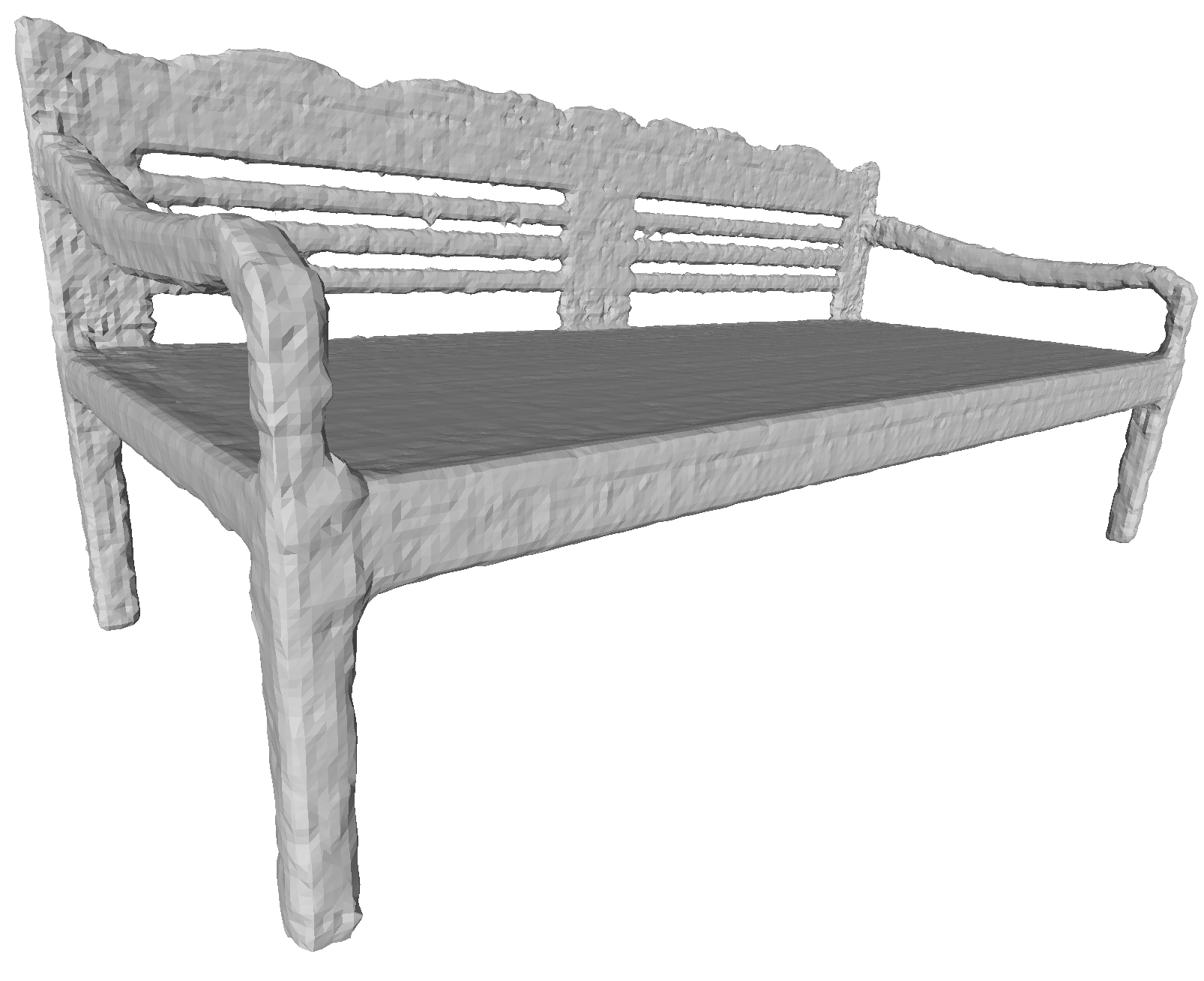} &
\includegraphics[width=\imgwidth]{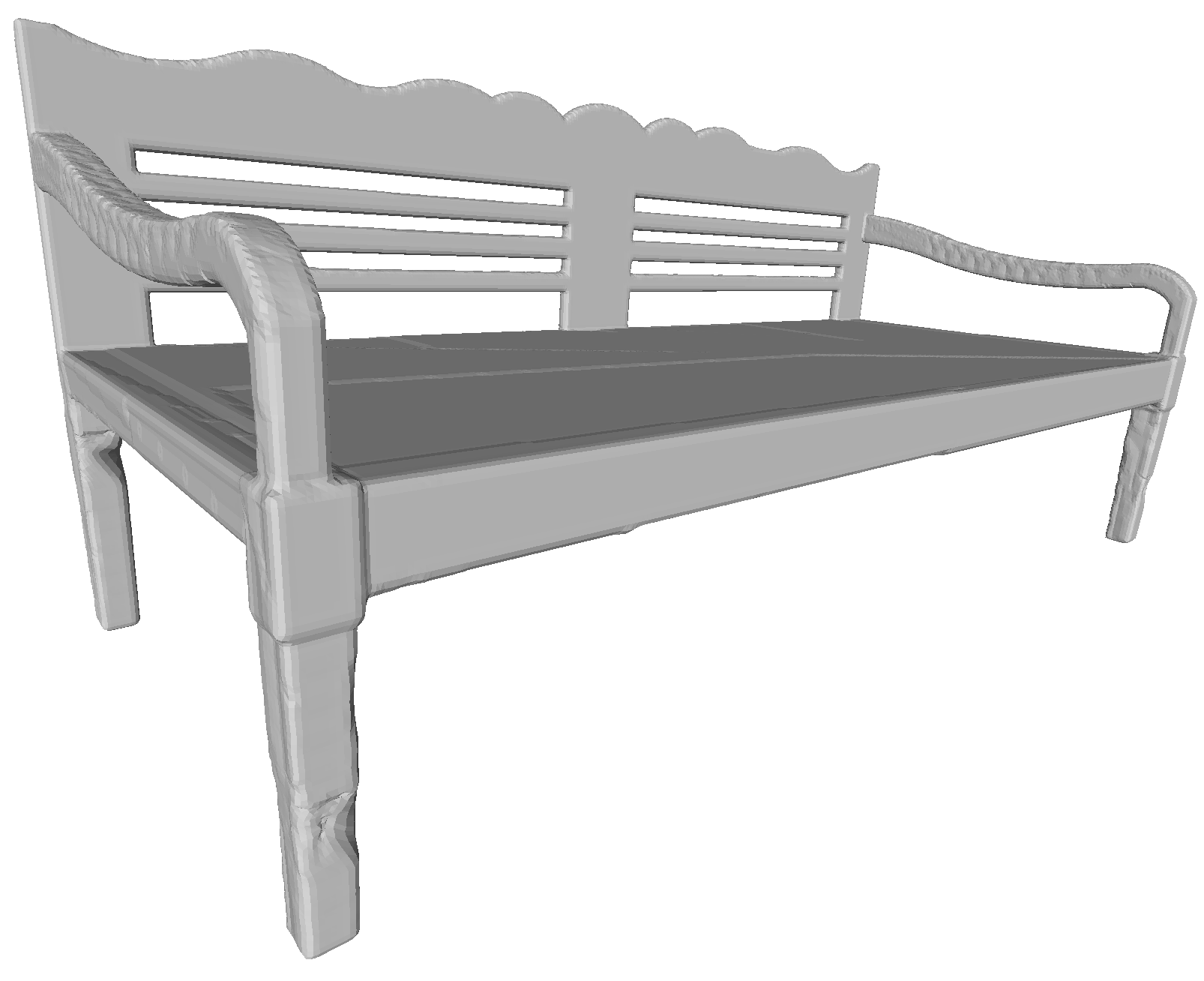} \\
\includegraphics[width=\imgwidth]{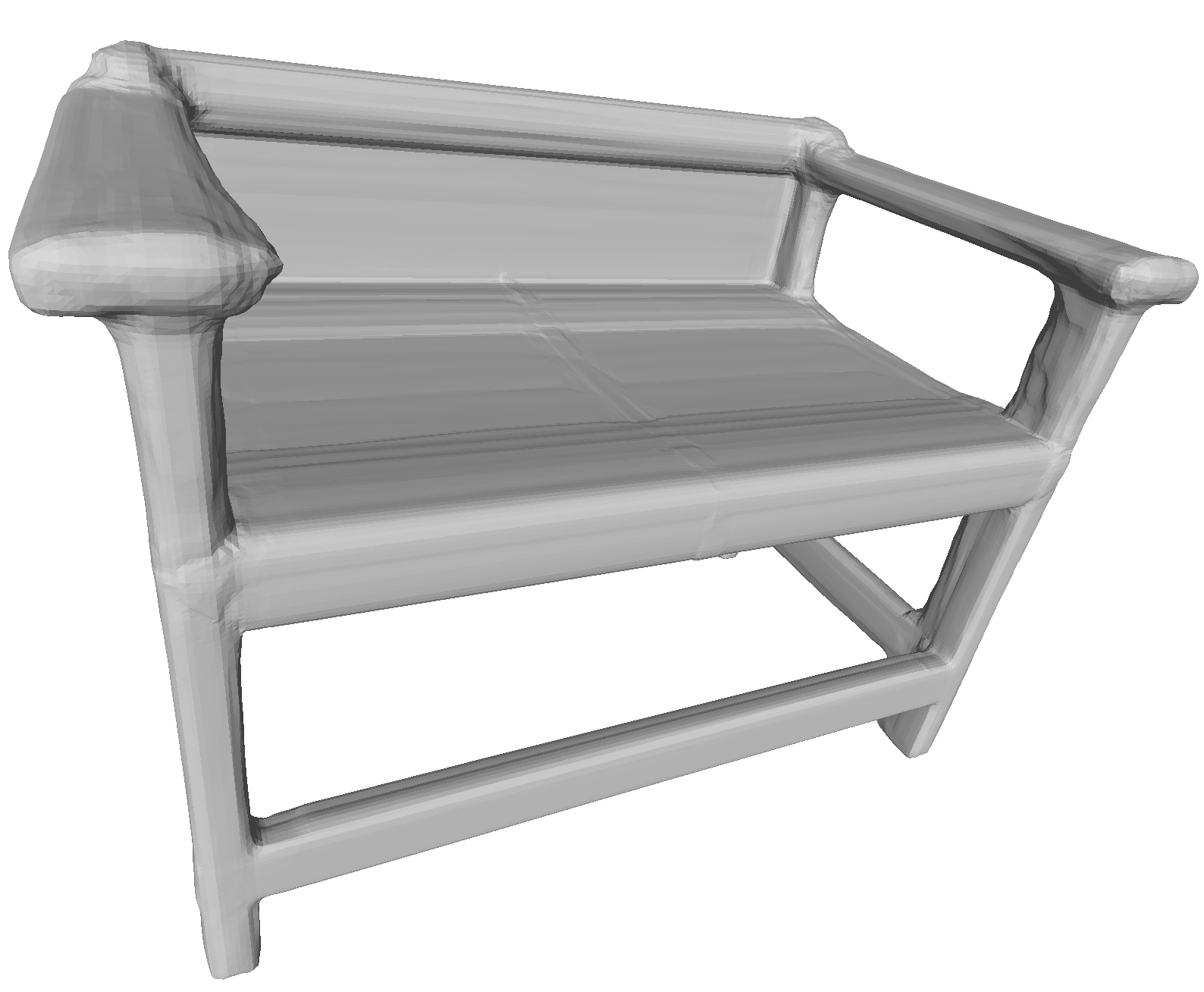} &
\includegraphics[width=\imgwidth]{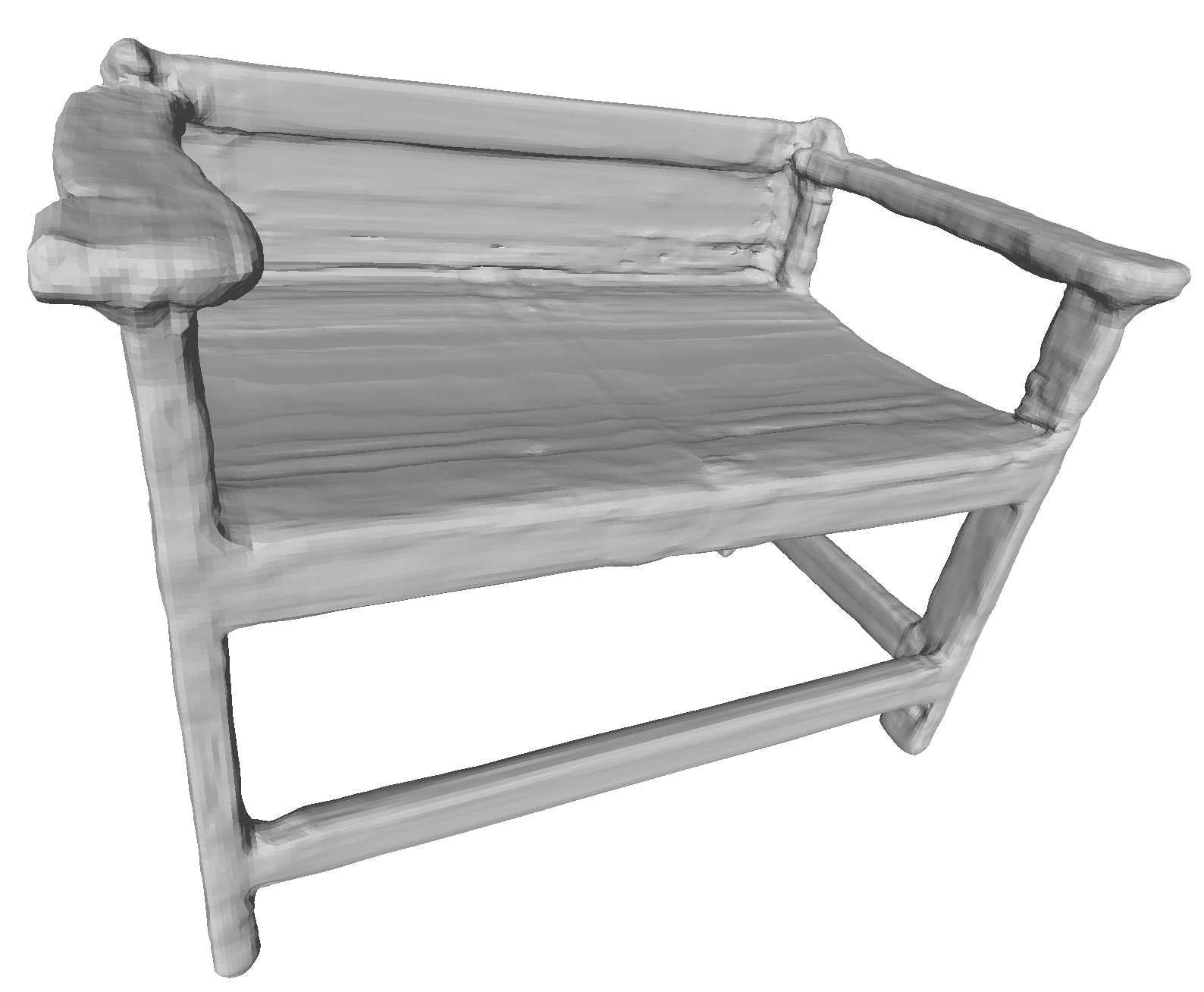} &
\includegraphics[width=\imgwidth]{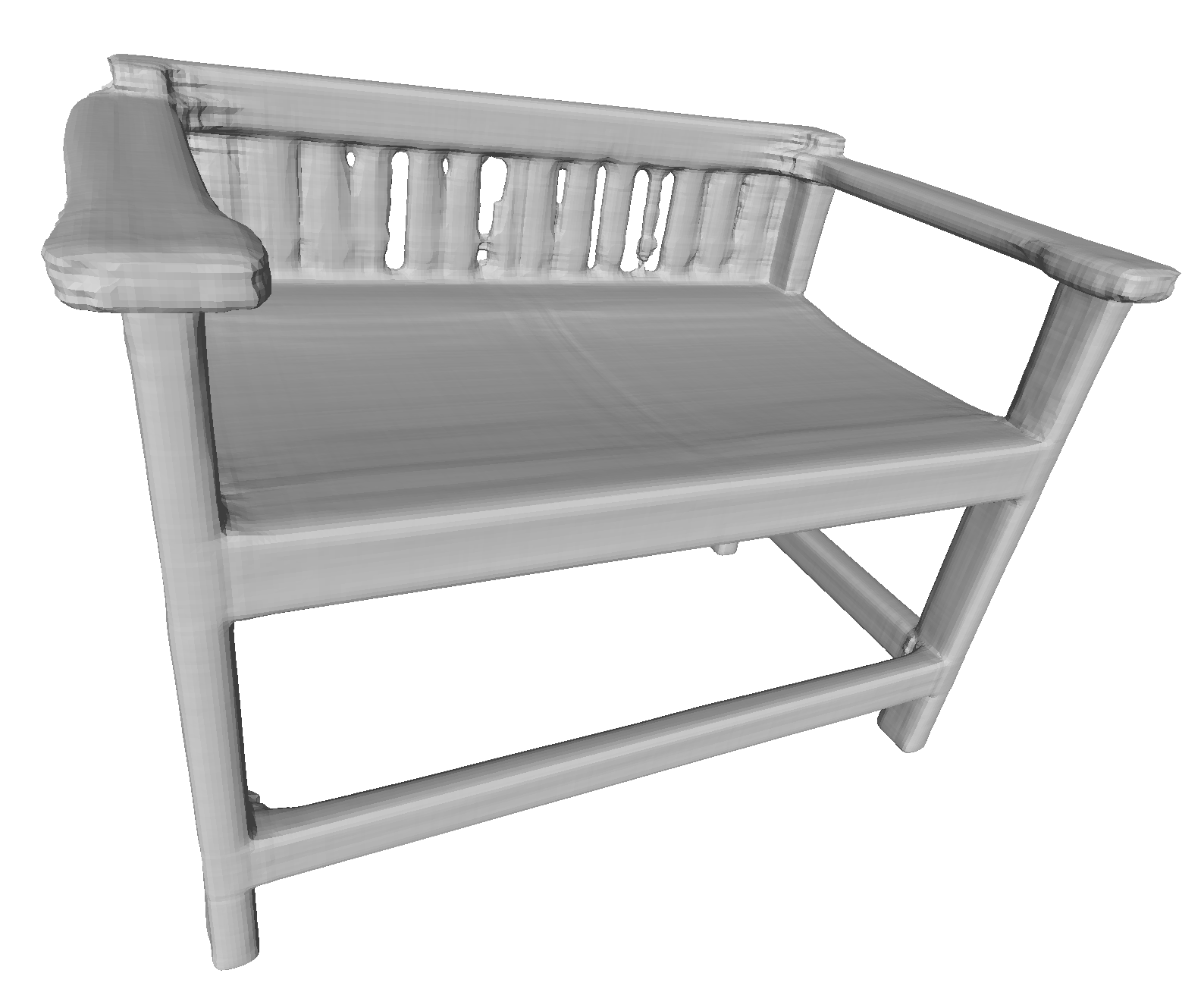} &
\includegraphics[width=\imgwidth]{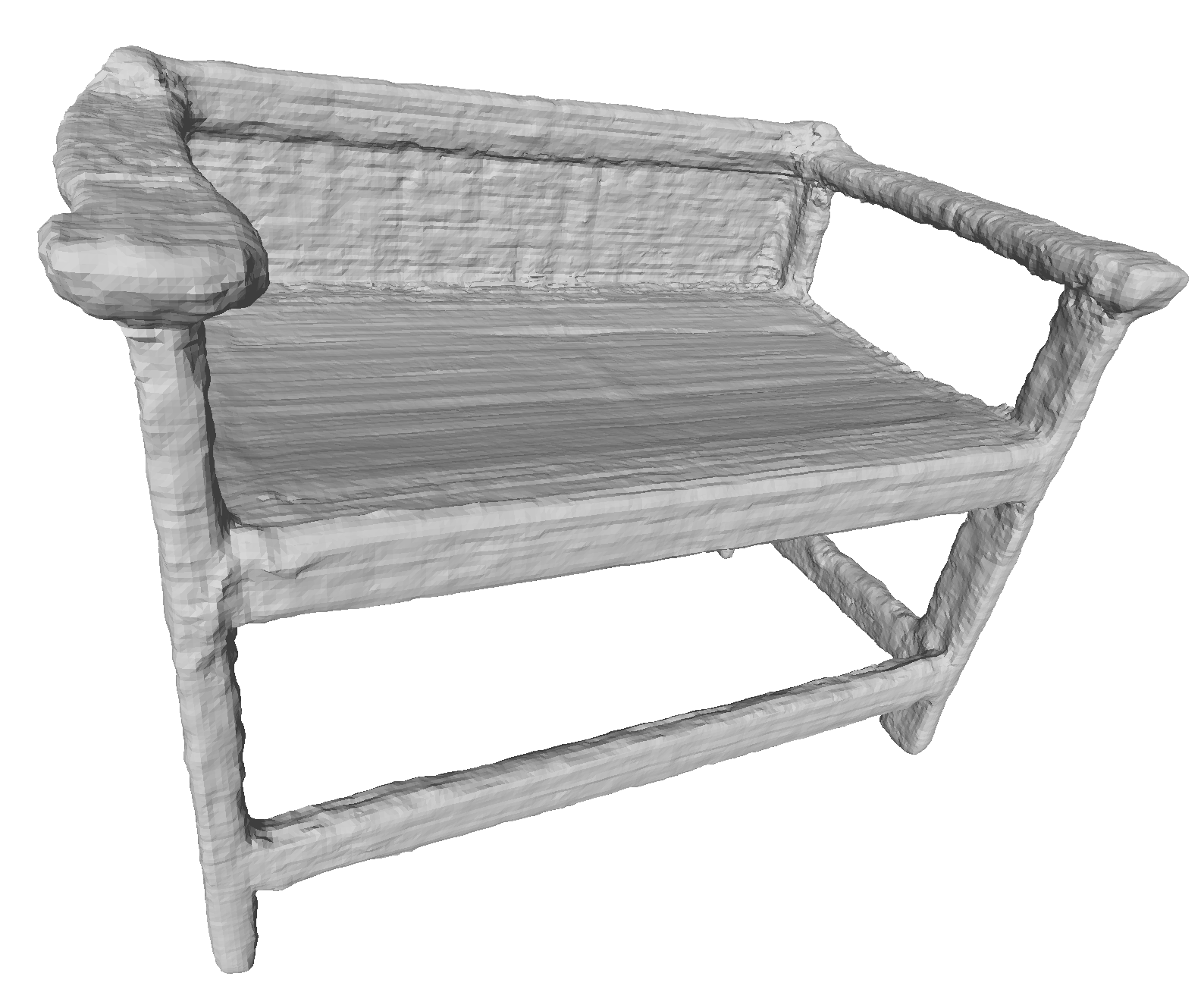} &
\includegraphics[width=\imgwidth]{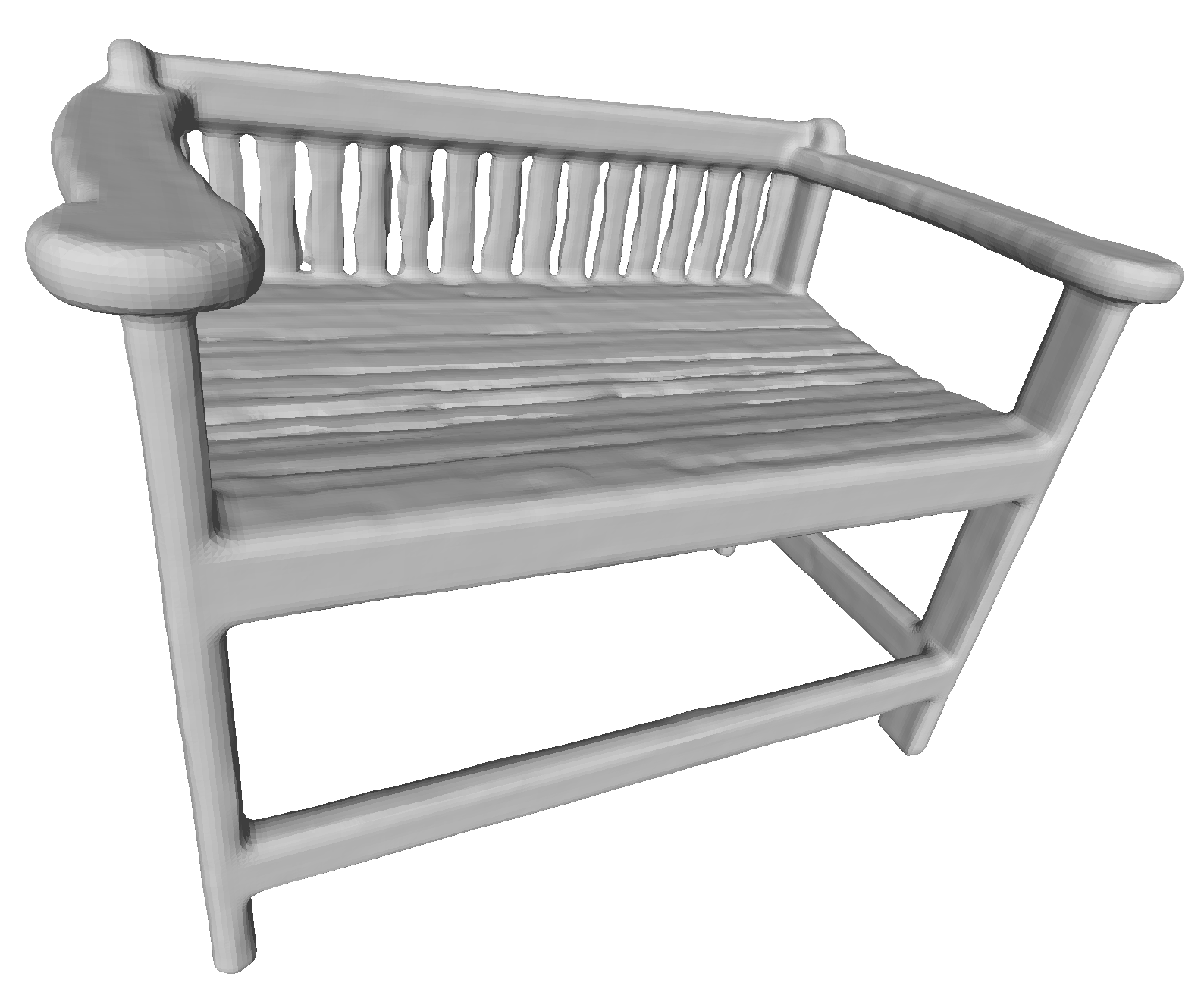} &
\includegraphics[width=\imgwidth]{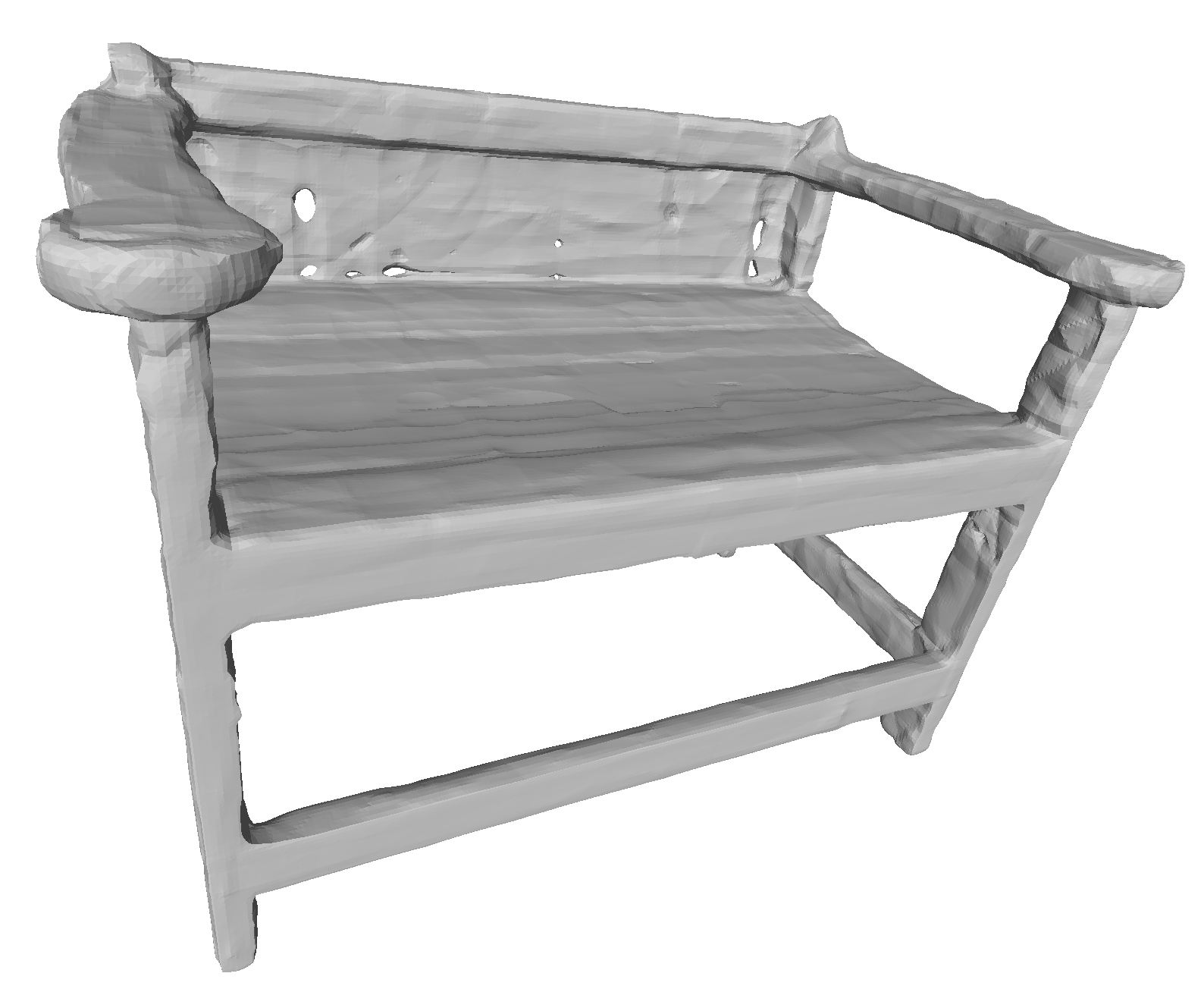} &
\includegraphics[width=\imgwidth]{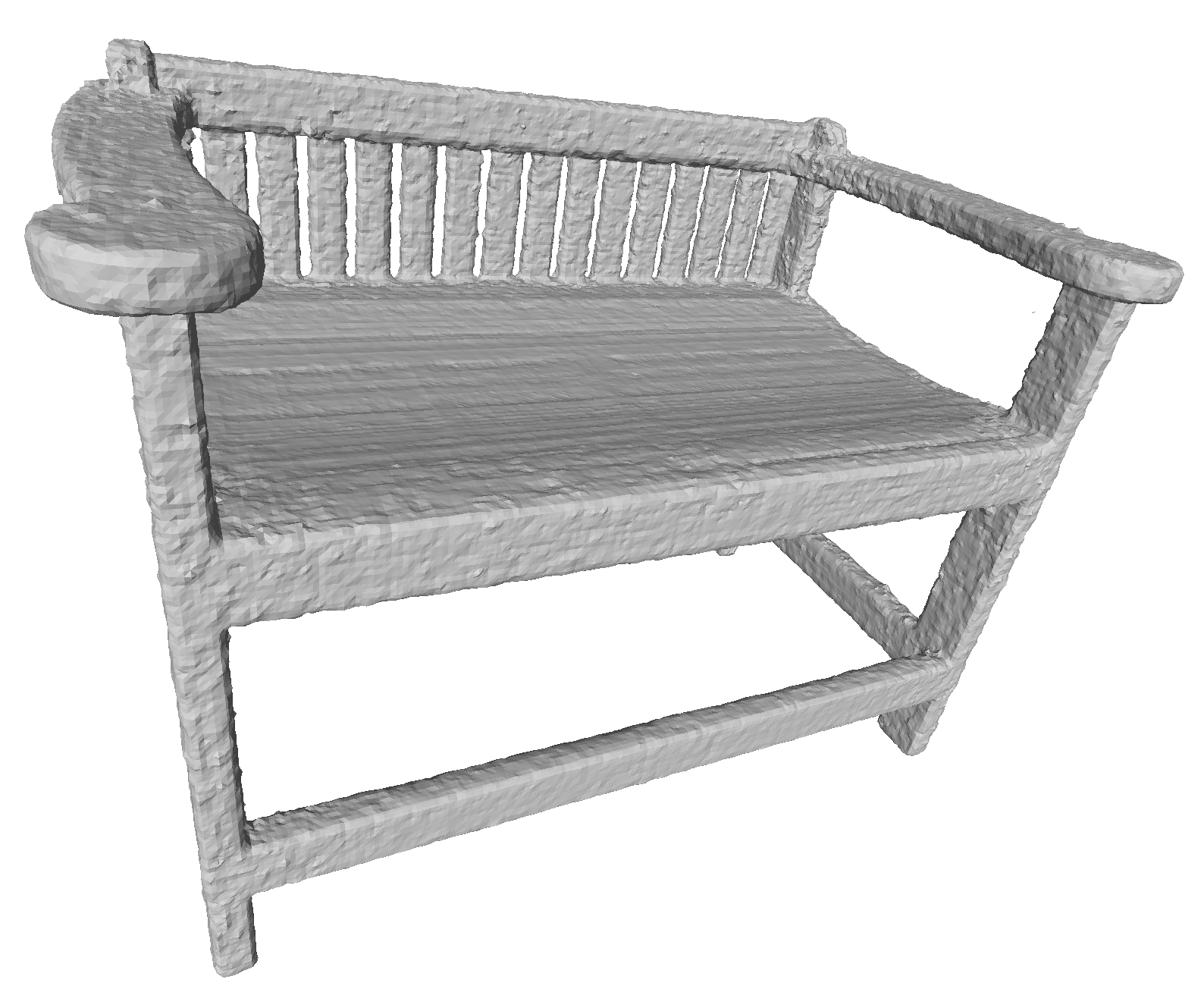} &
\includegraphics[width=\imgwidth]{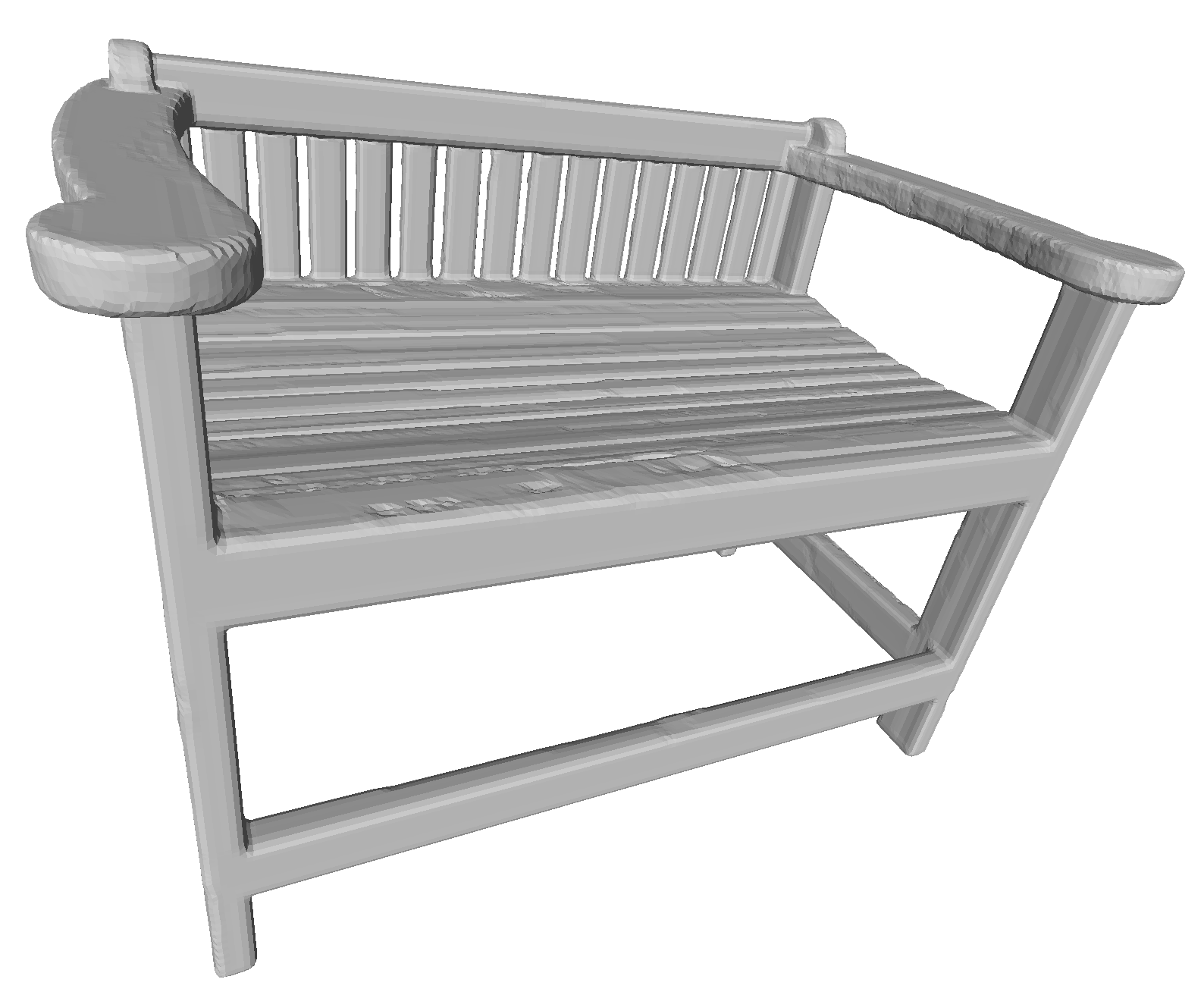} \\

\includegraphics[width=\imgwidth]{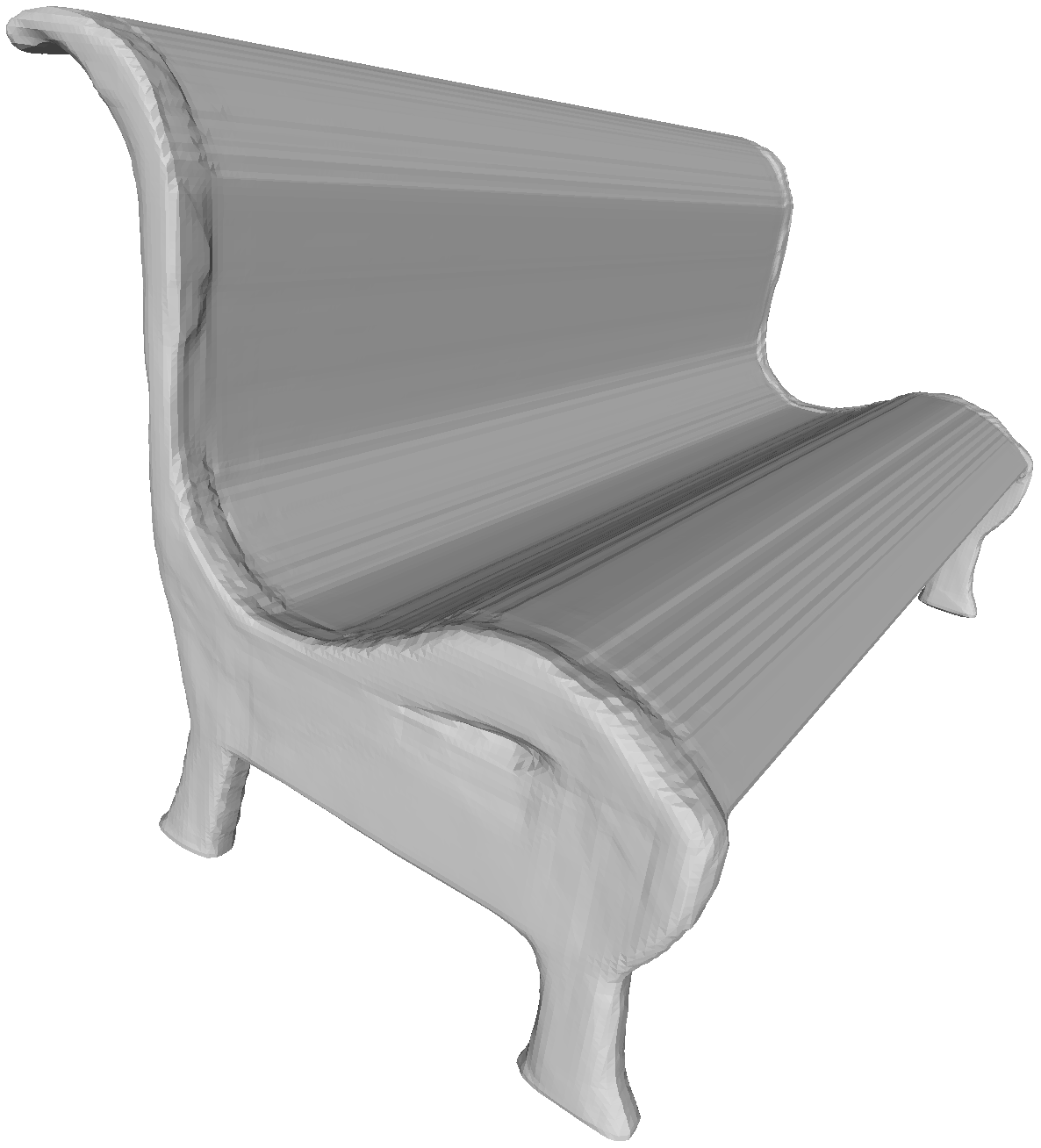} &
\includegraphics[width=\imgwidth]{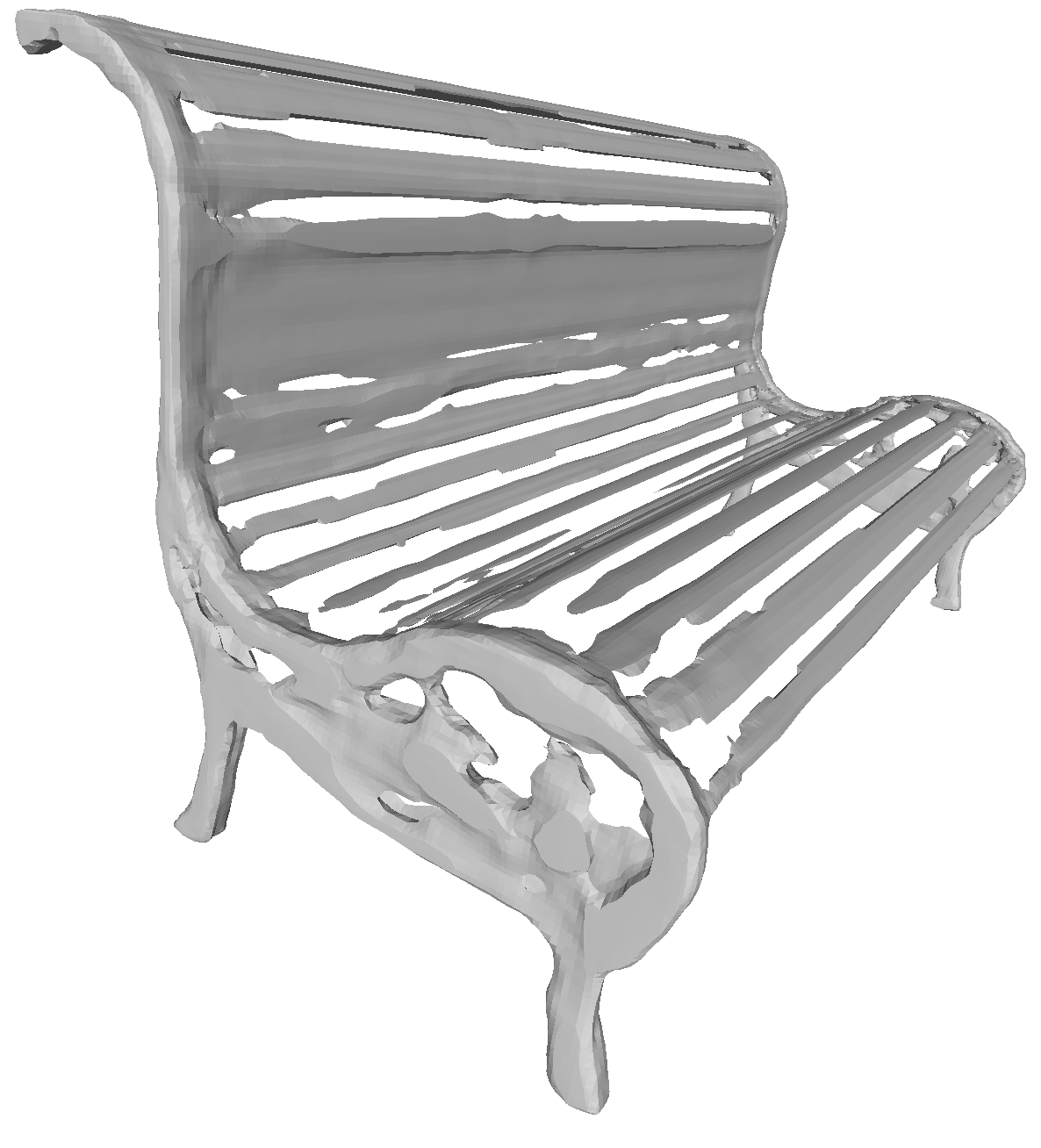} &
\includegraphics[width=\imgwidth]{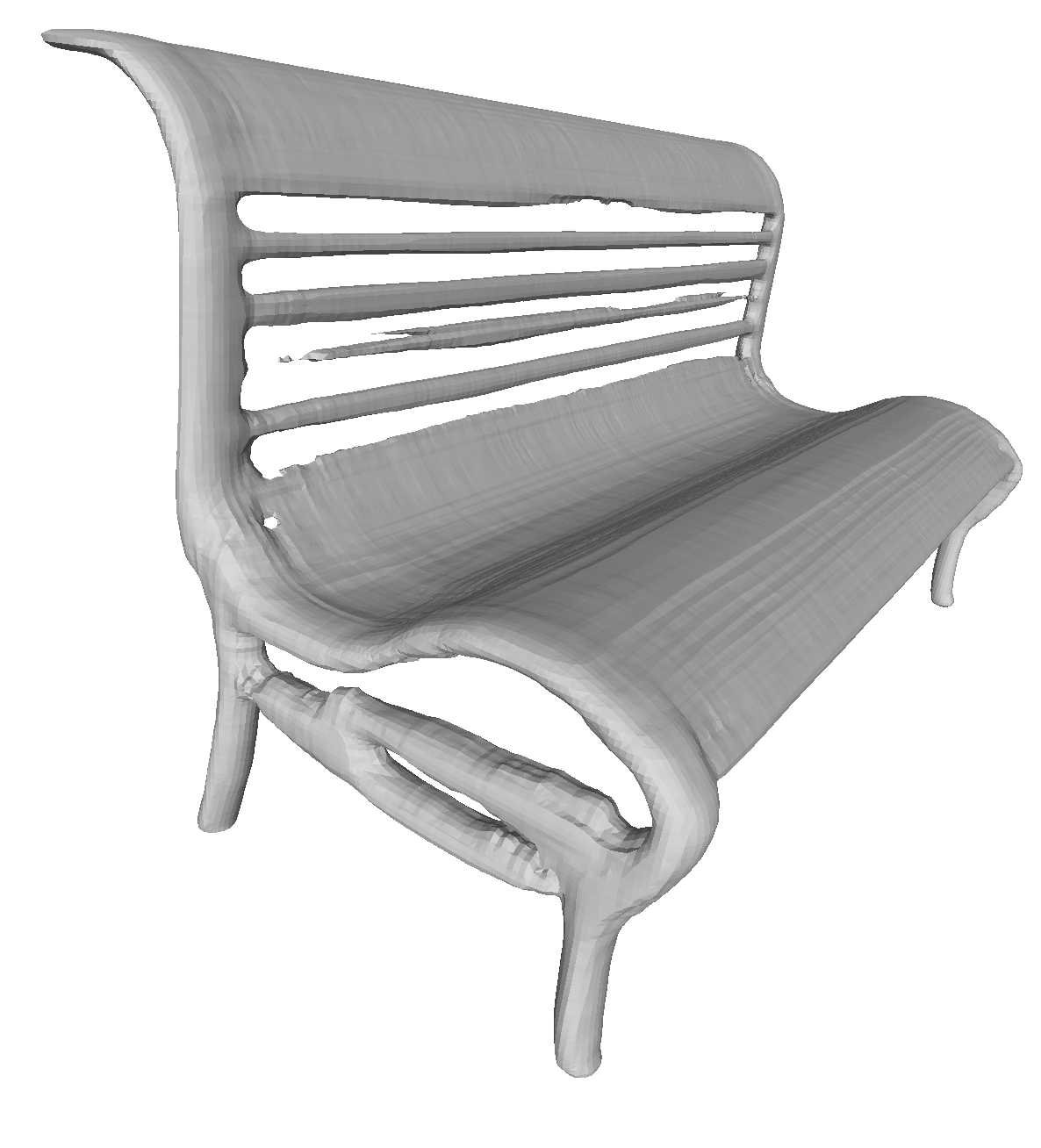} &
\includegraphics[width=\imgwidth]{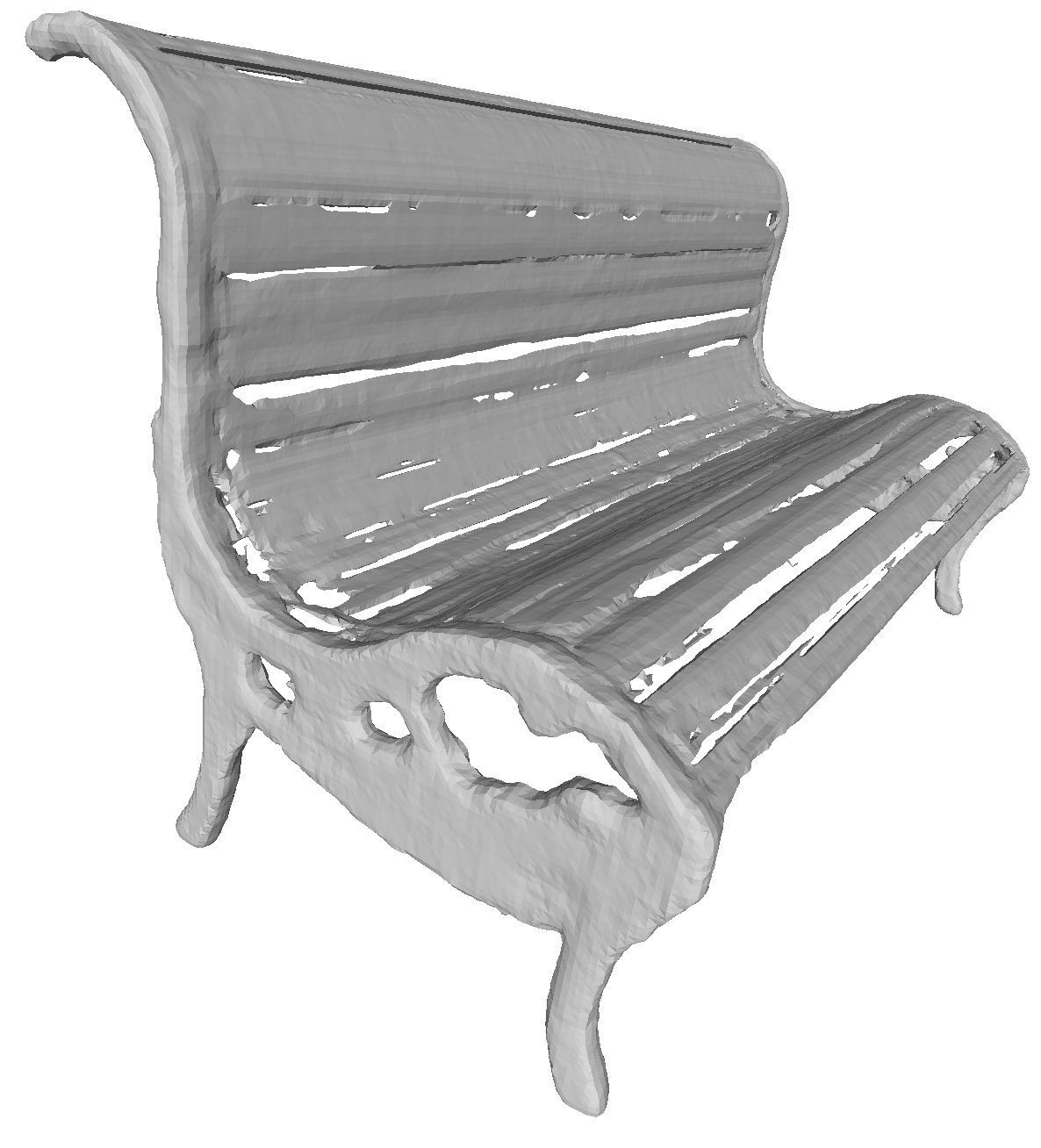} &
\includegraphics[width=\imgwidth]{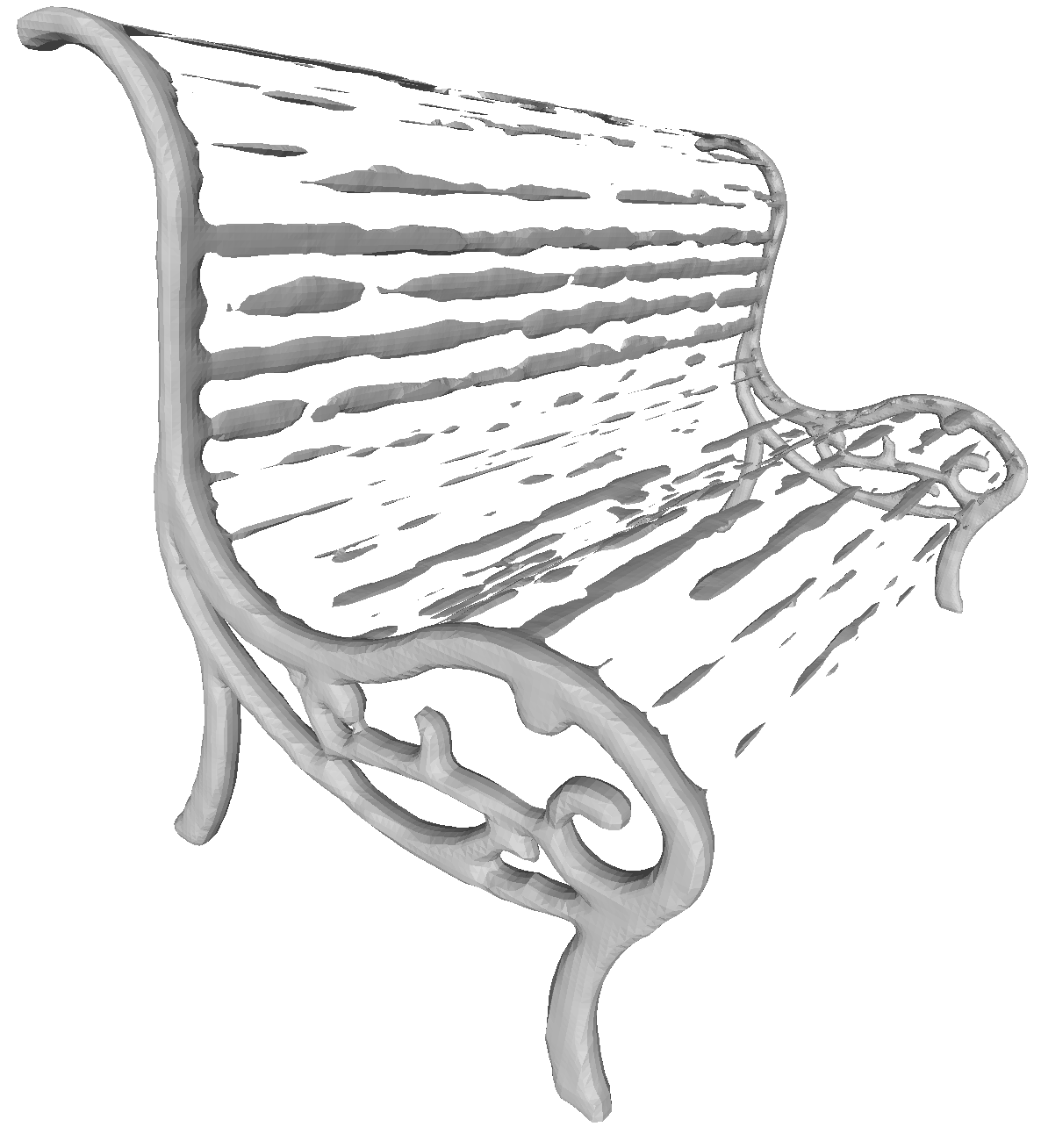} &
\includegraphics[width=\imgwidth]{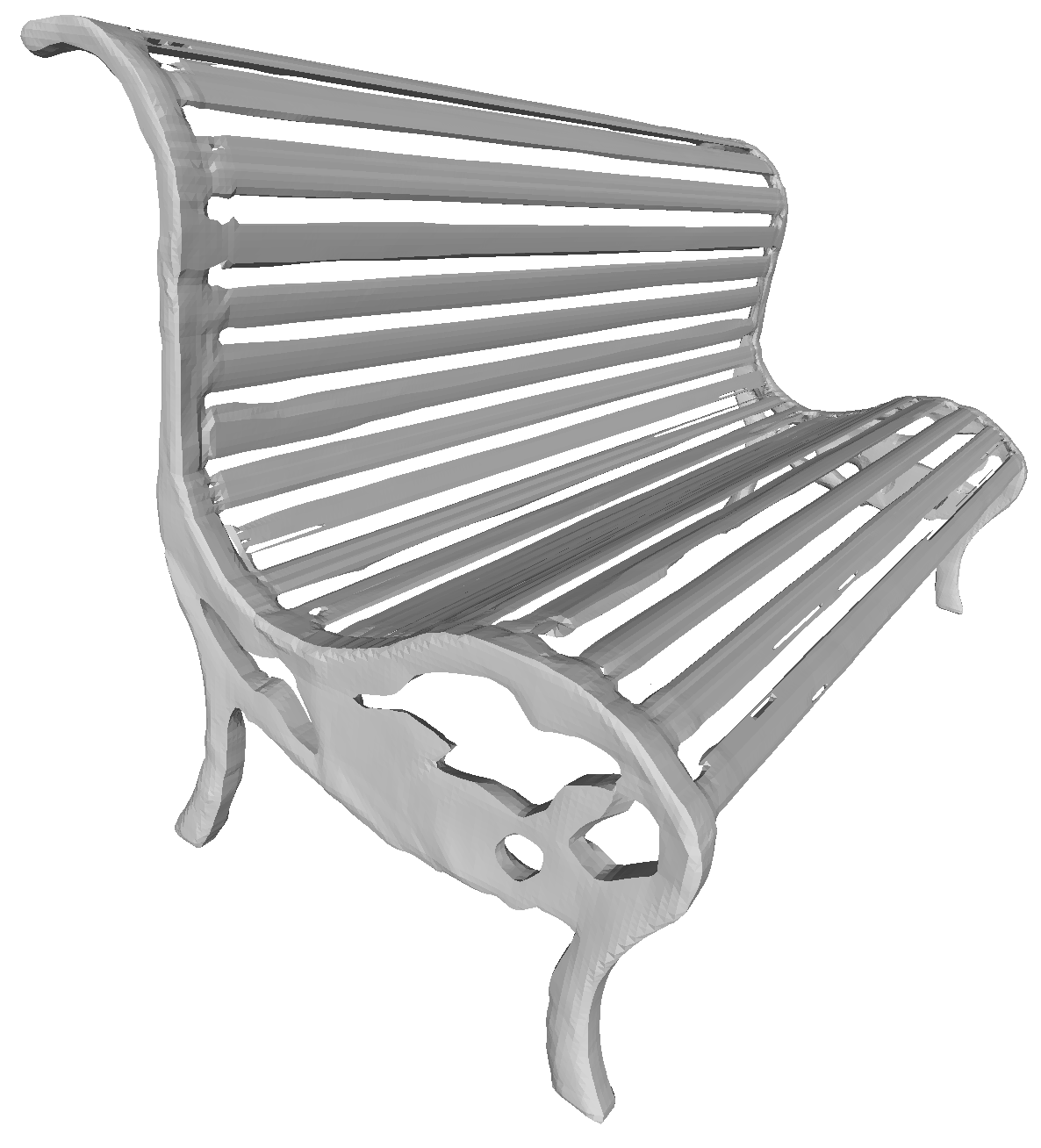} &
\includegraphics[width=\imgwidth]{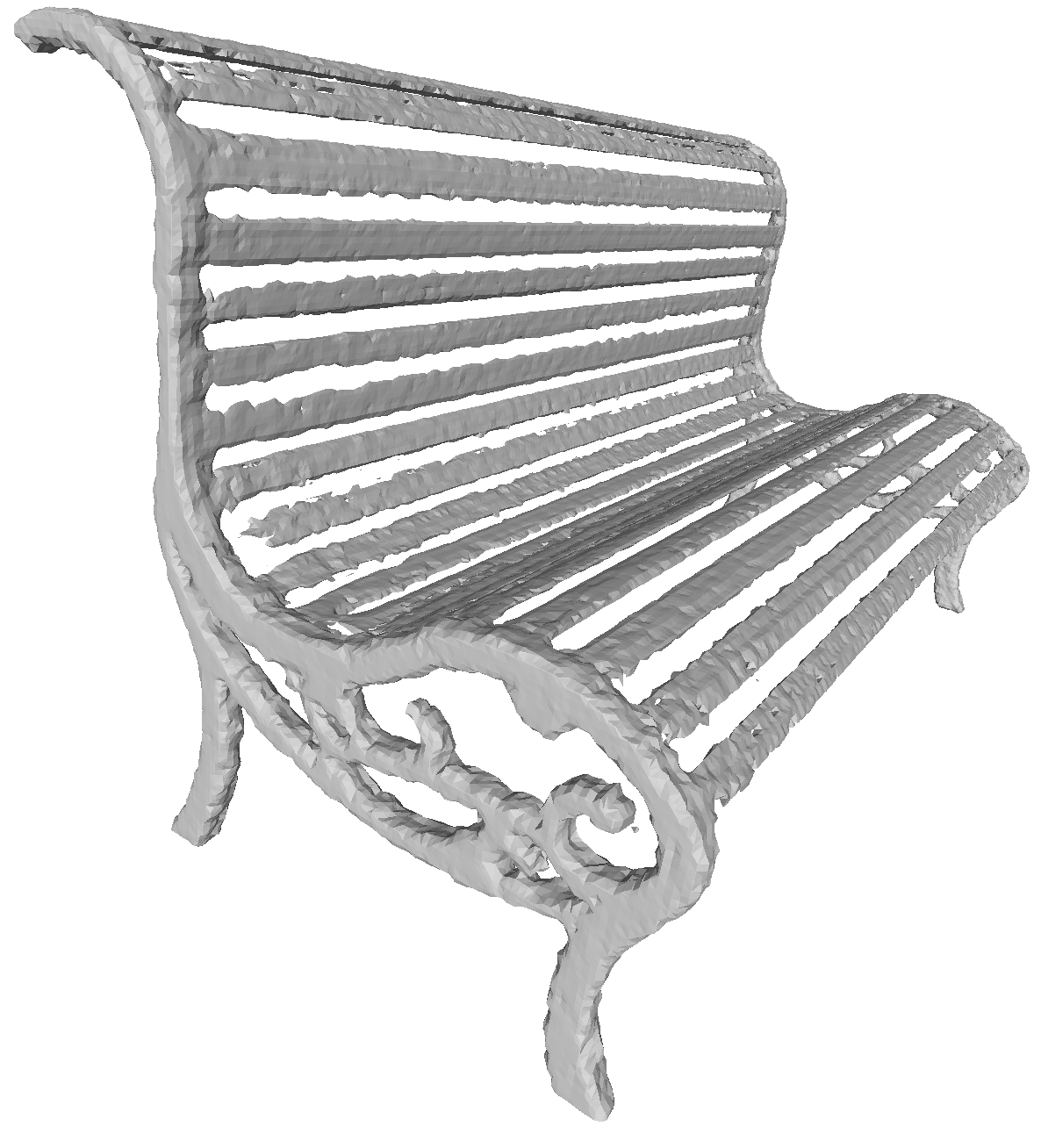} &
\includegraphics[width=\imgwidth]{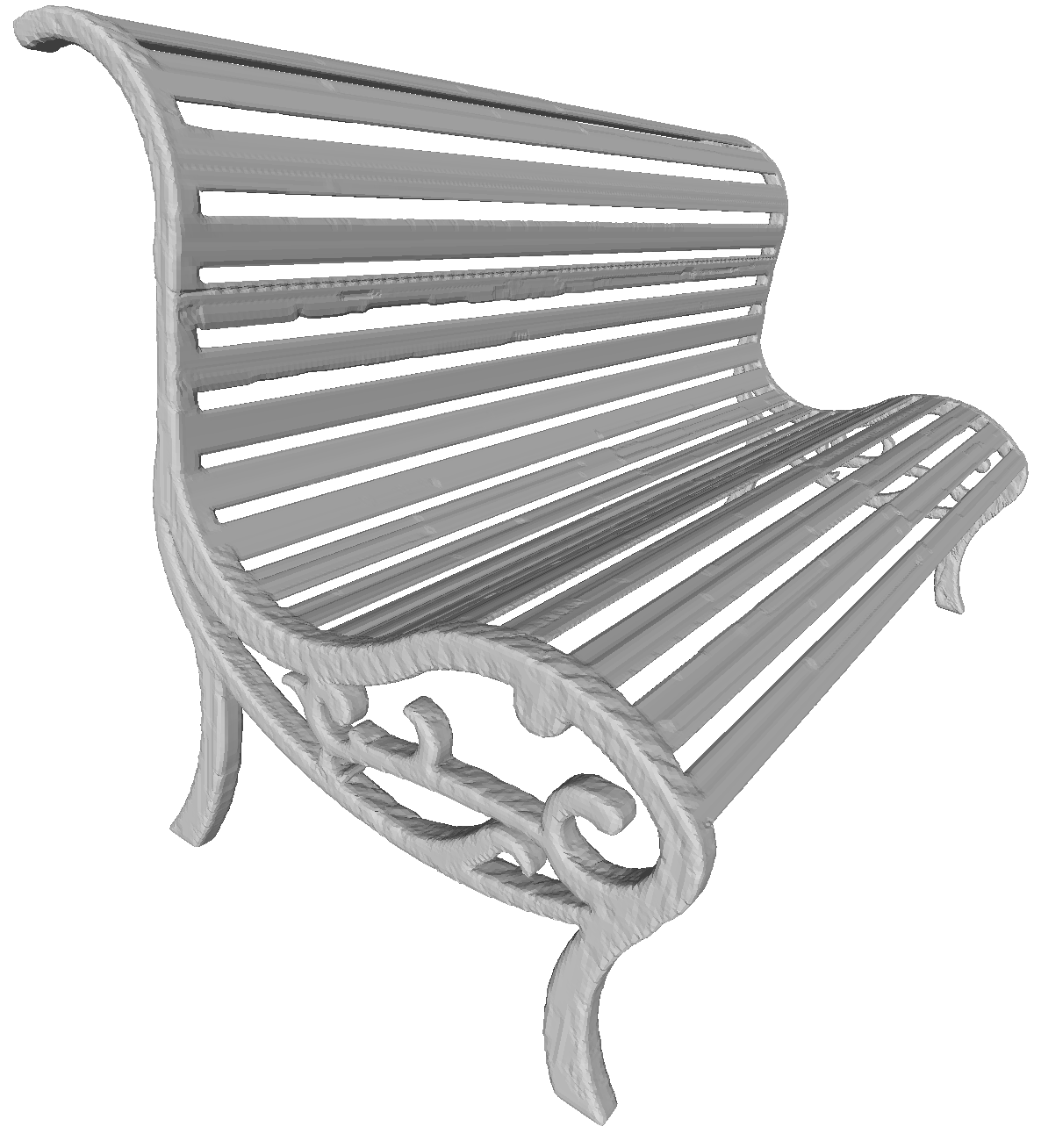} \\

\includegraphics[width=\imgwidth]{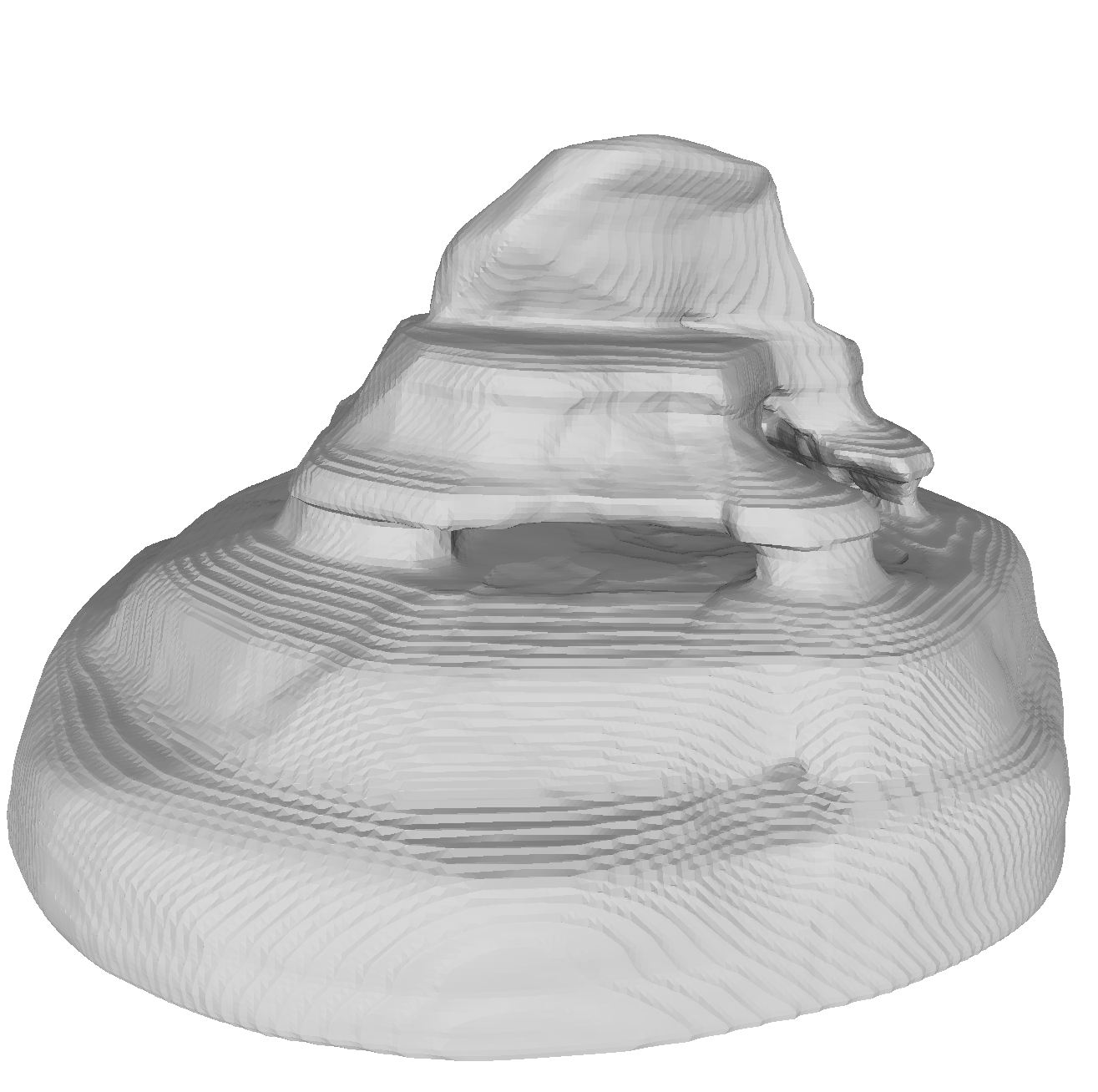} &
\includegraphics[width=\imgwidth]{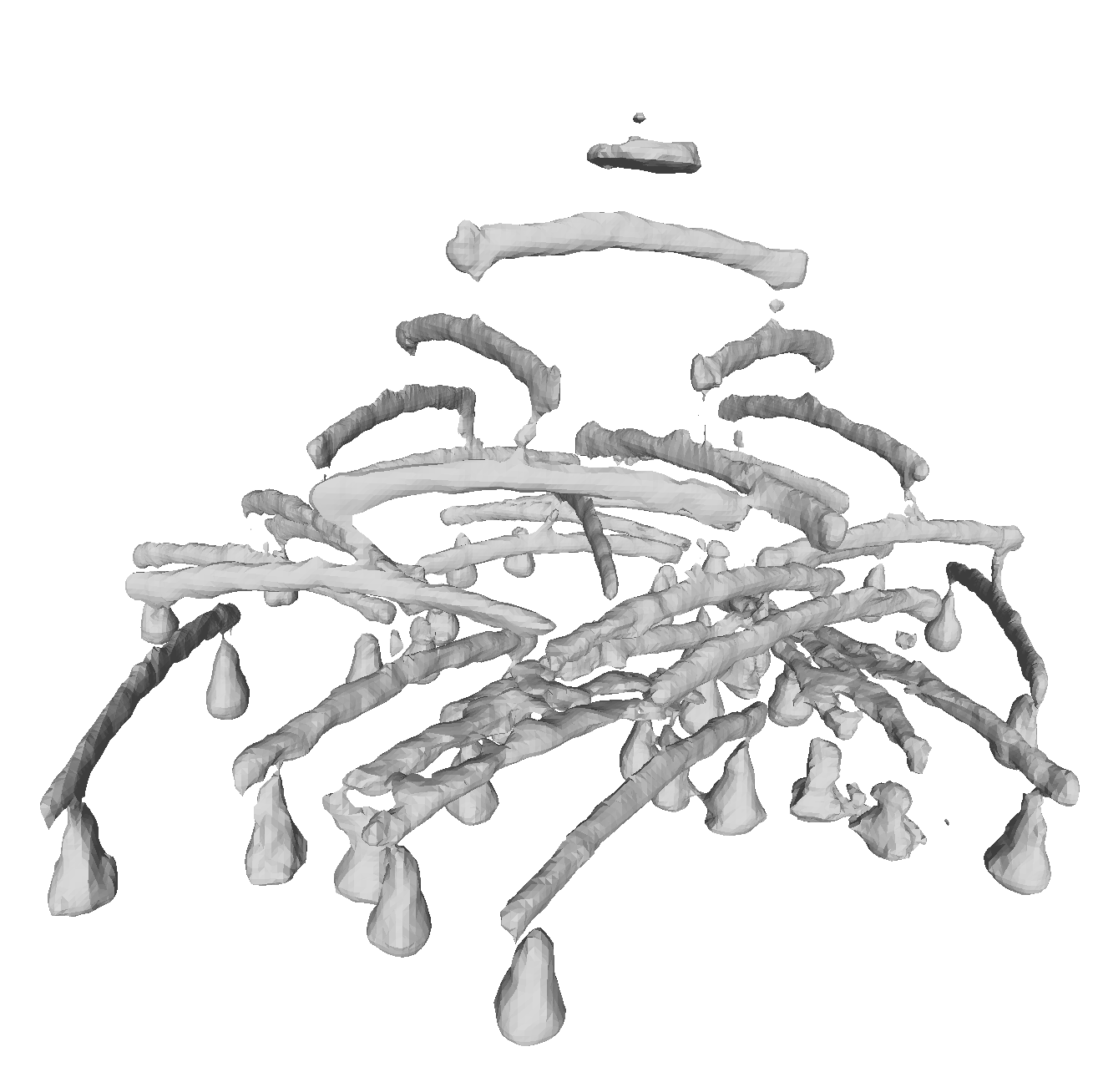} &
\includegraphics[width=\imgwidth]{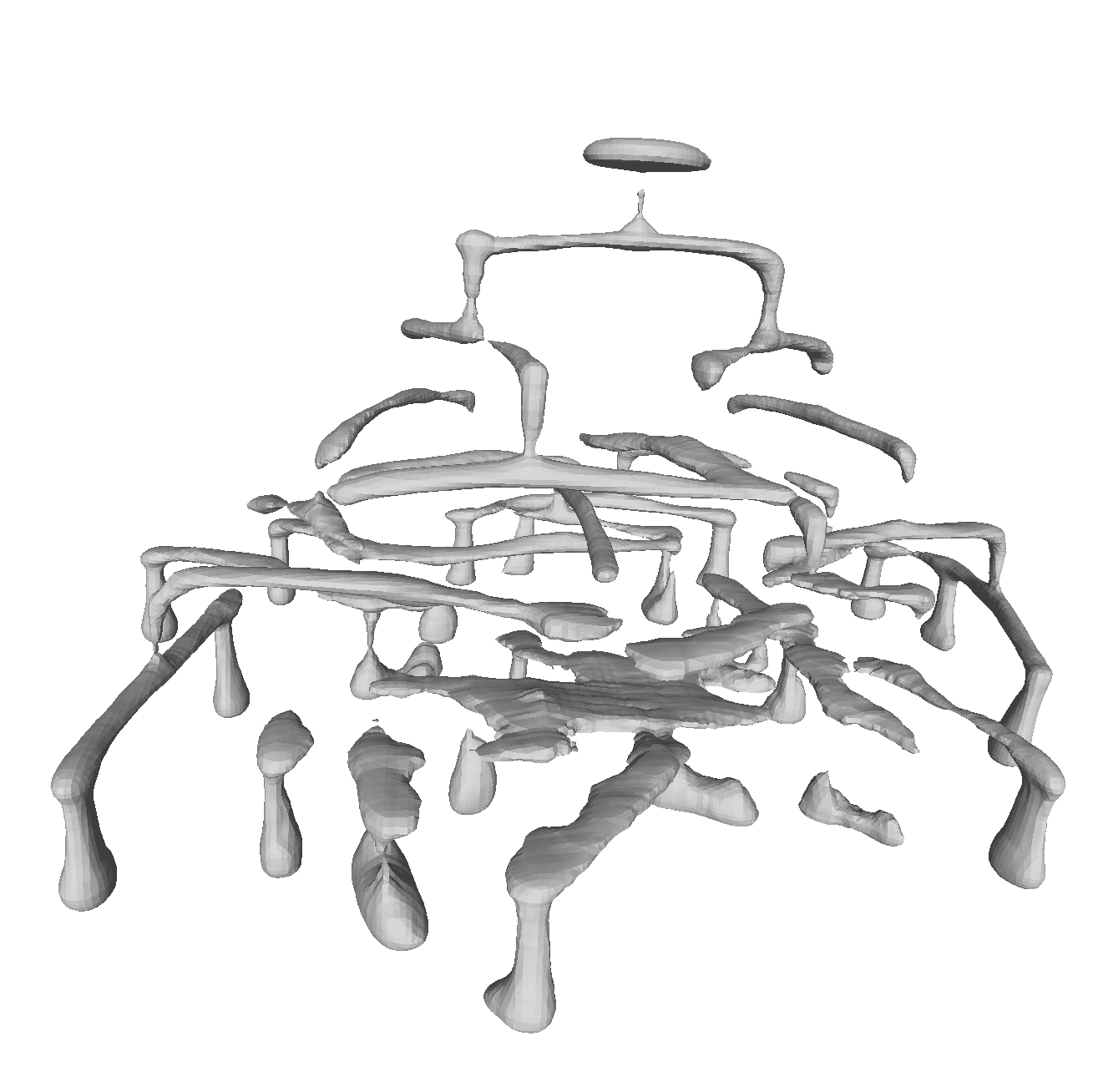} &
\includegraphics[width=\imgwidth]{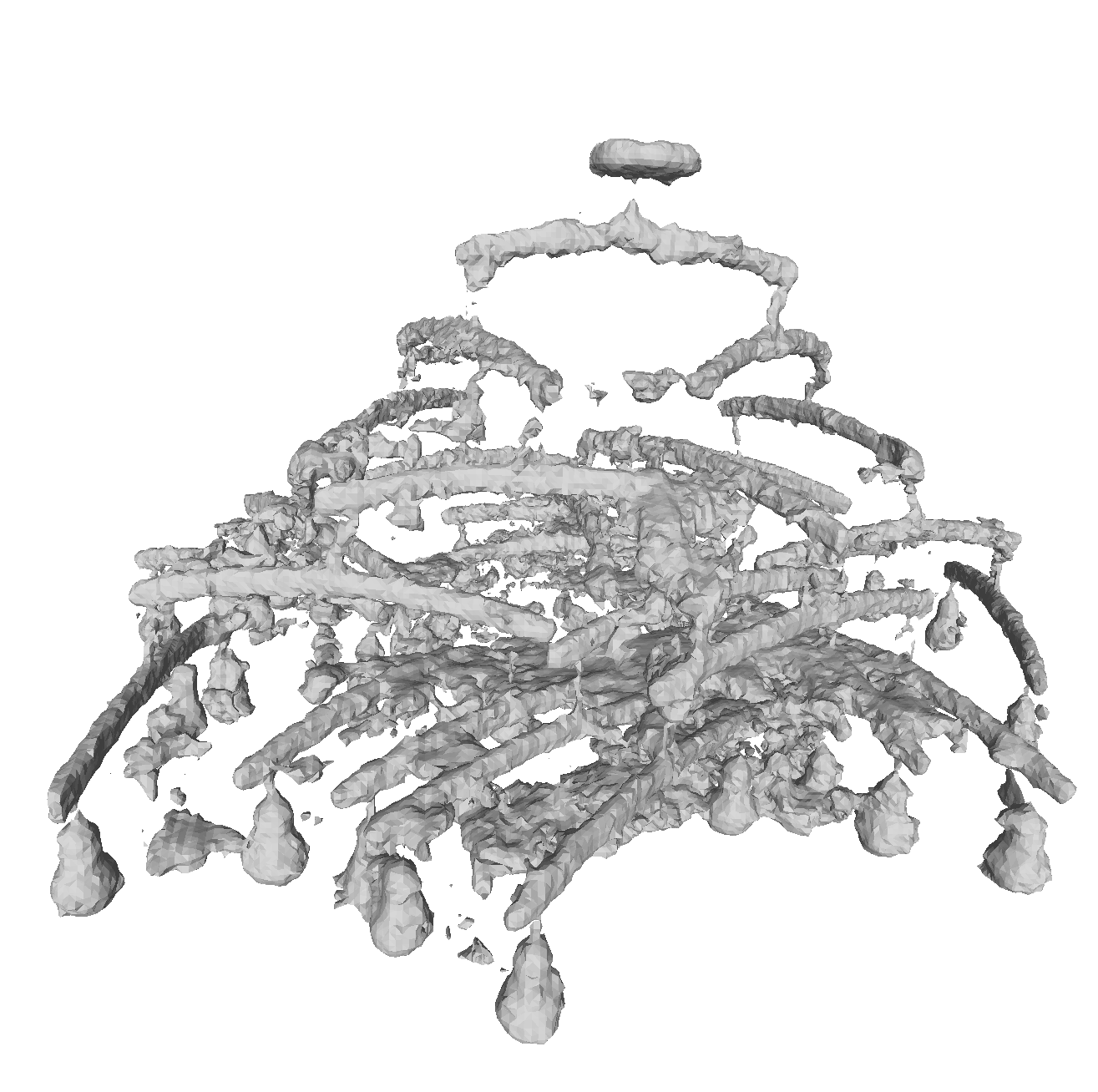} &
\includegraphics[width=\imgwidth]{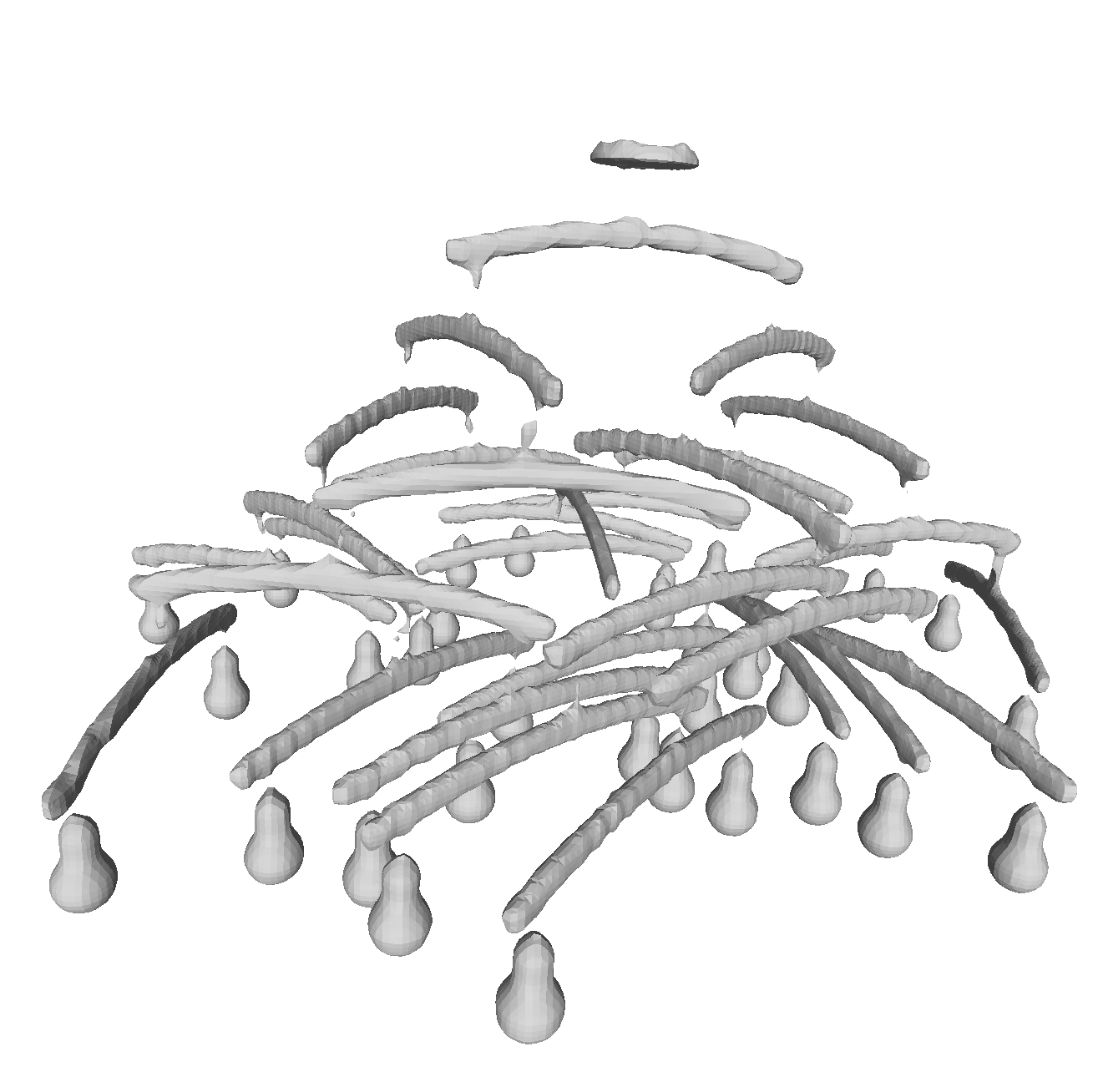} &
\includegraphics[width=\imgwidth]{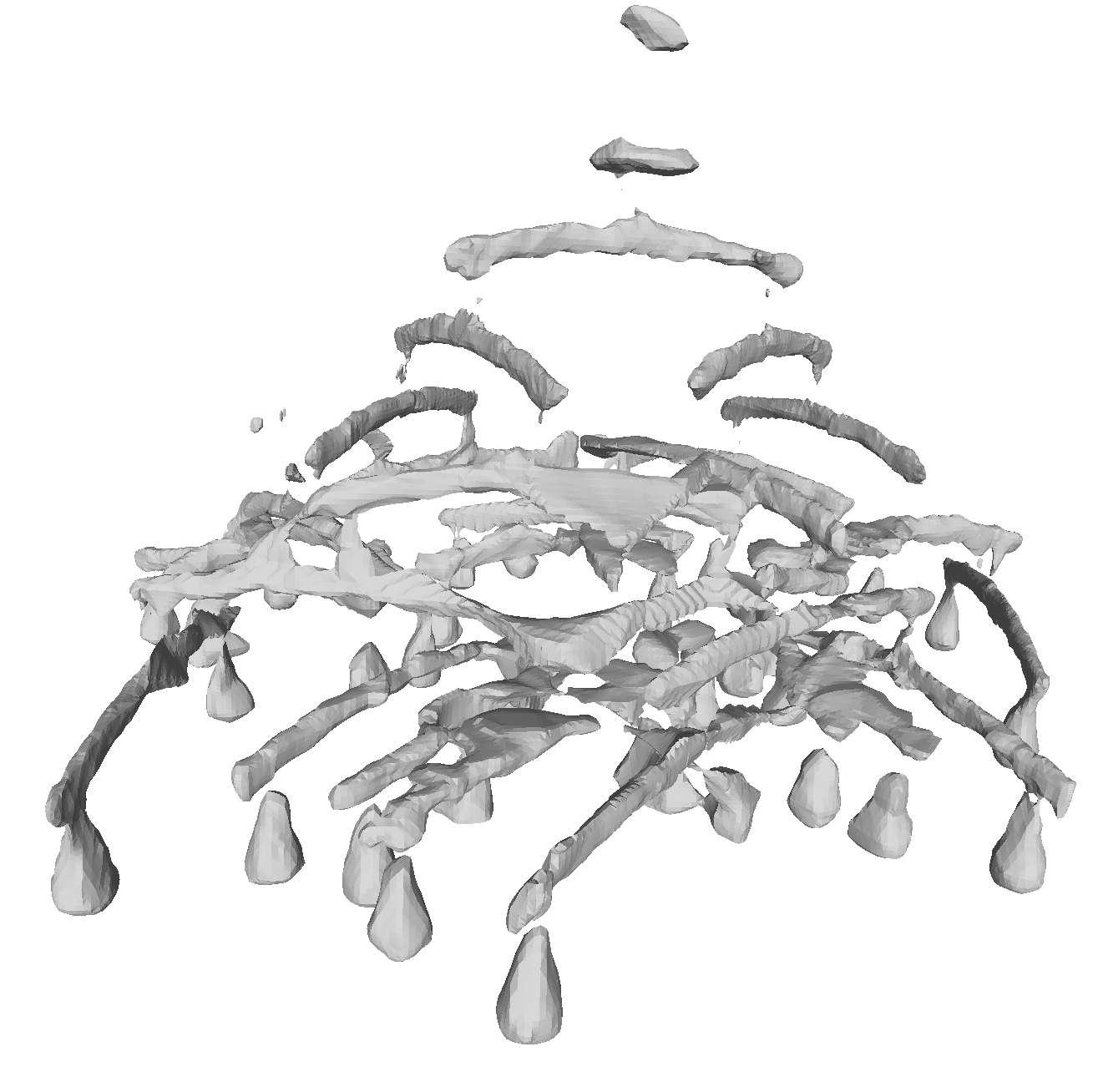} &
\includegraphics[width=\imgwidth]{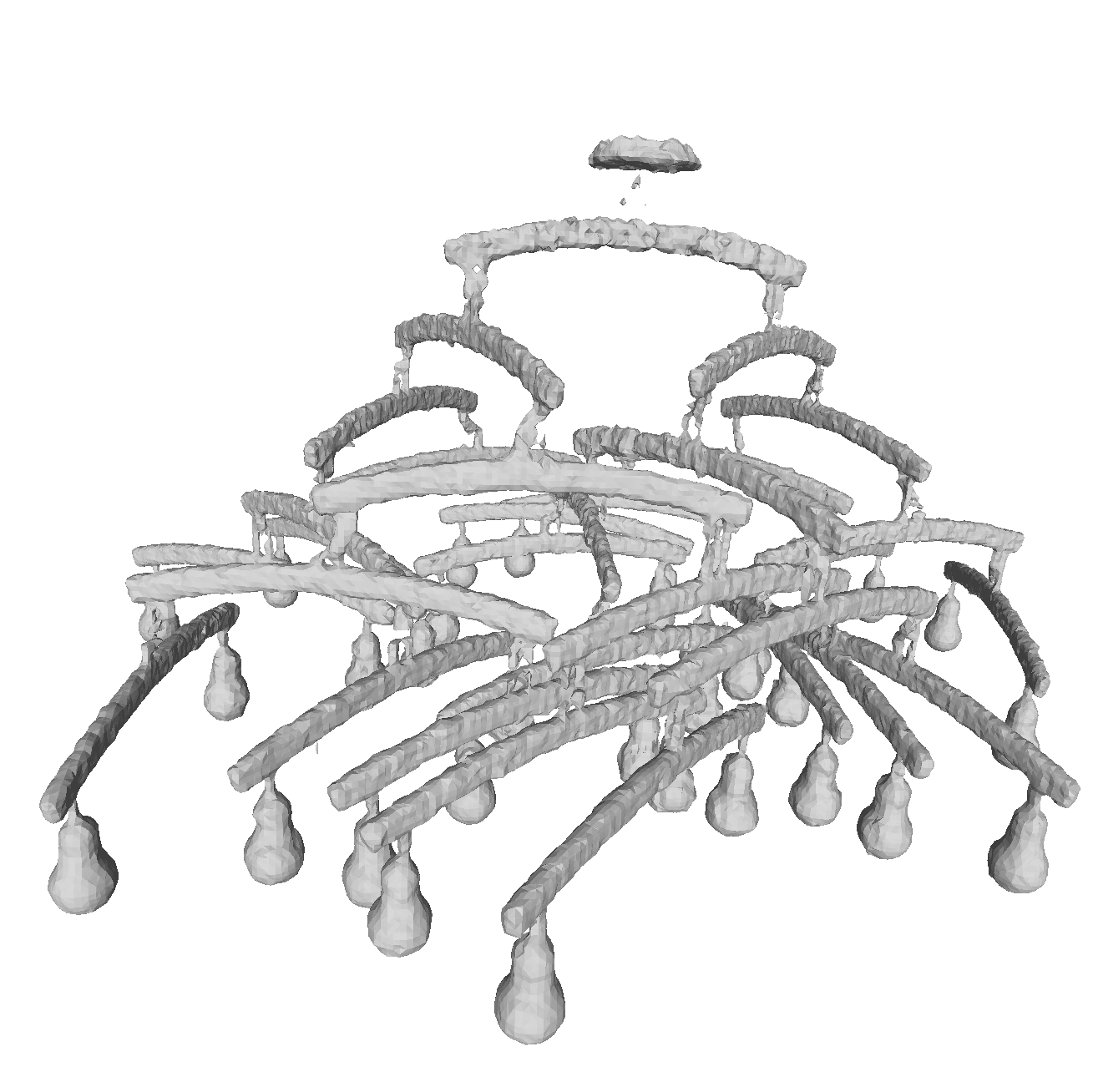} &
\includegraphics[width=\imgwidth]{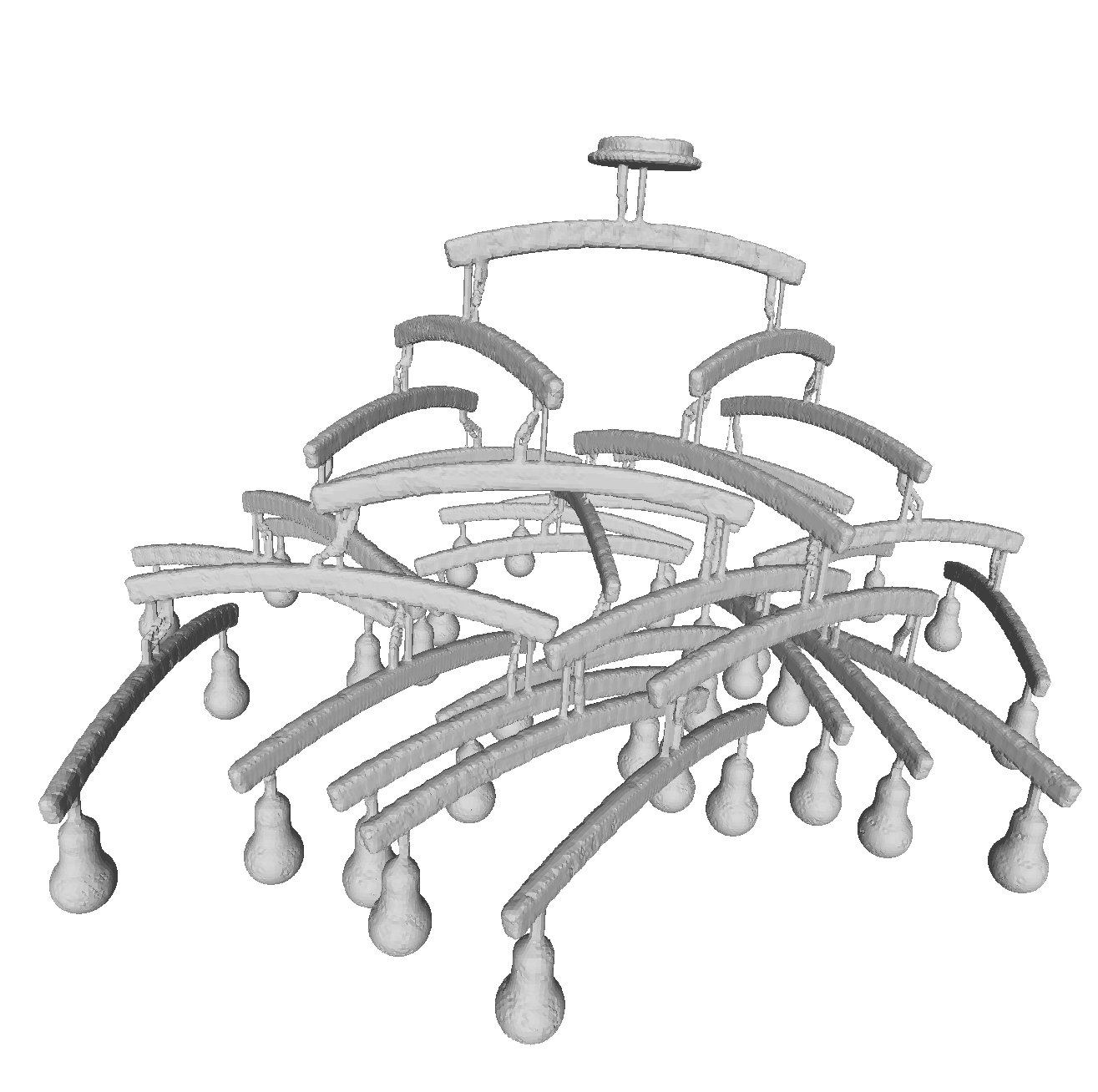} \\

\includegraphics[width=\imgwidth, height=3cm, keepaspectratio]{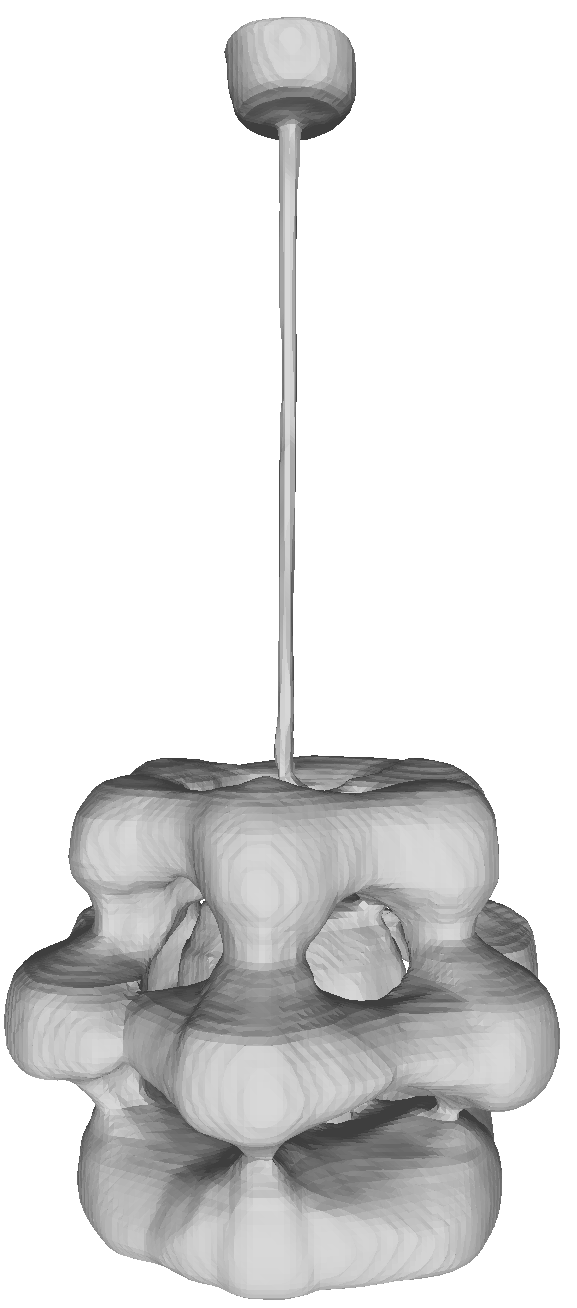} &
\includegraphics[width=\imgwidth, height=3cm, keepaspectratio]{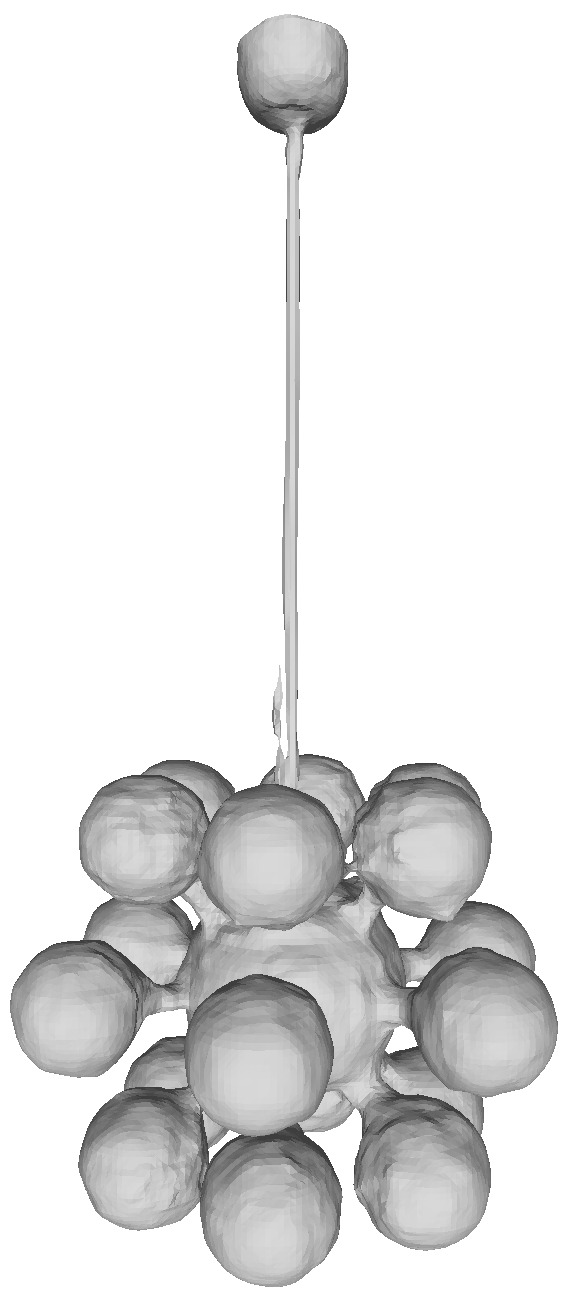} &
\includegraphics[width=\imgwidth, height=3cm, keepaspectratio]{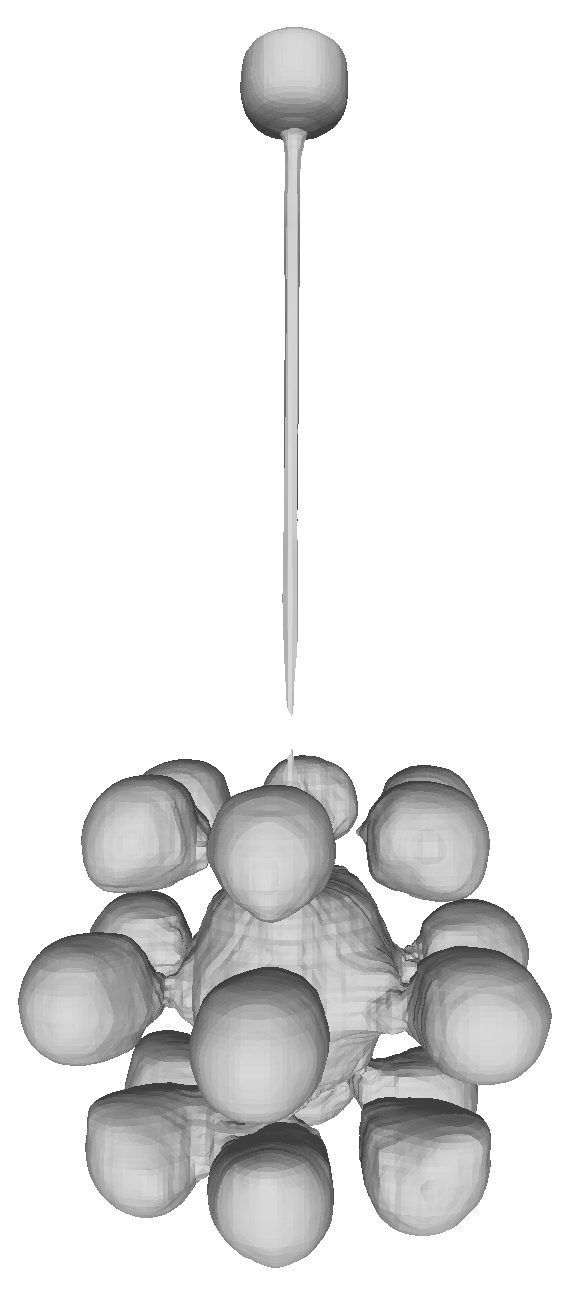} &
\includegraphics[width=\imgwidth, height=3cm, keepaspectratio]{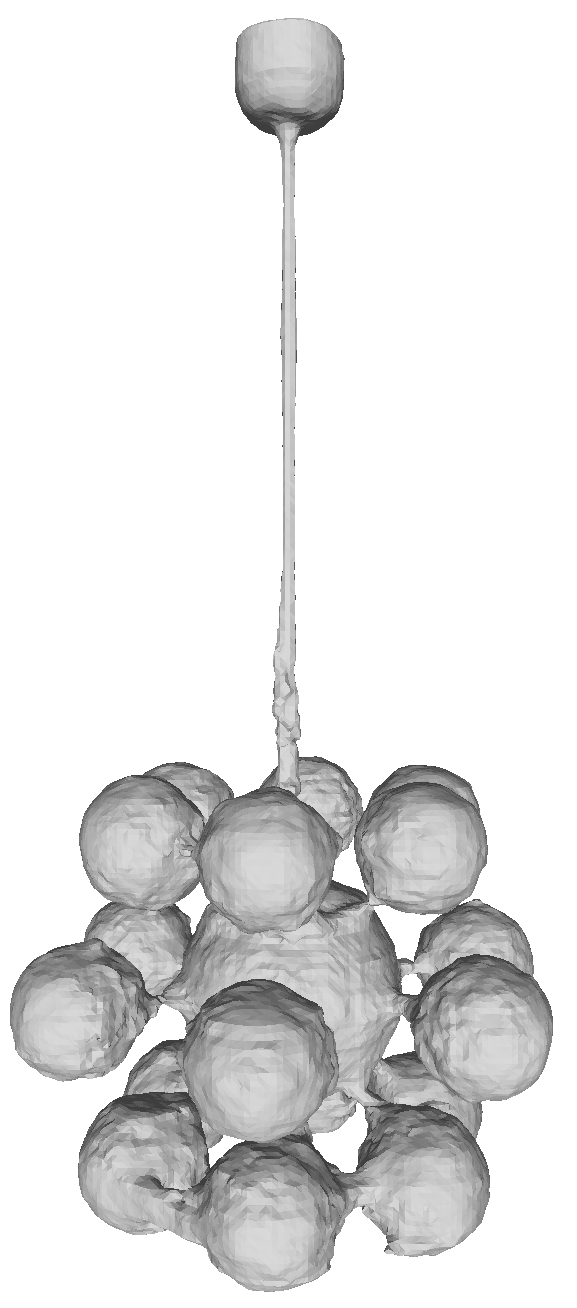} &
\includegraphics[width=\imgwidth, height=3cm, keepaspectratio]{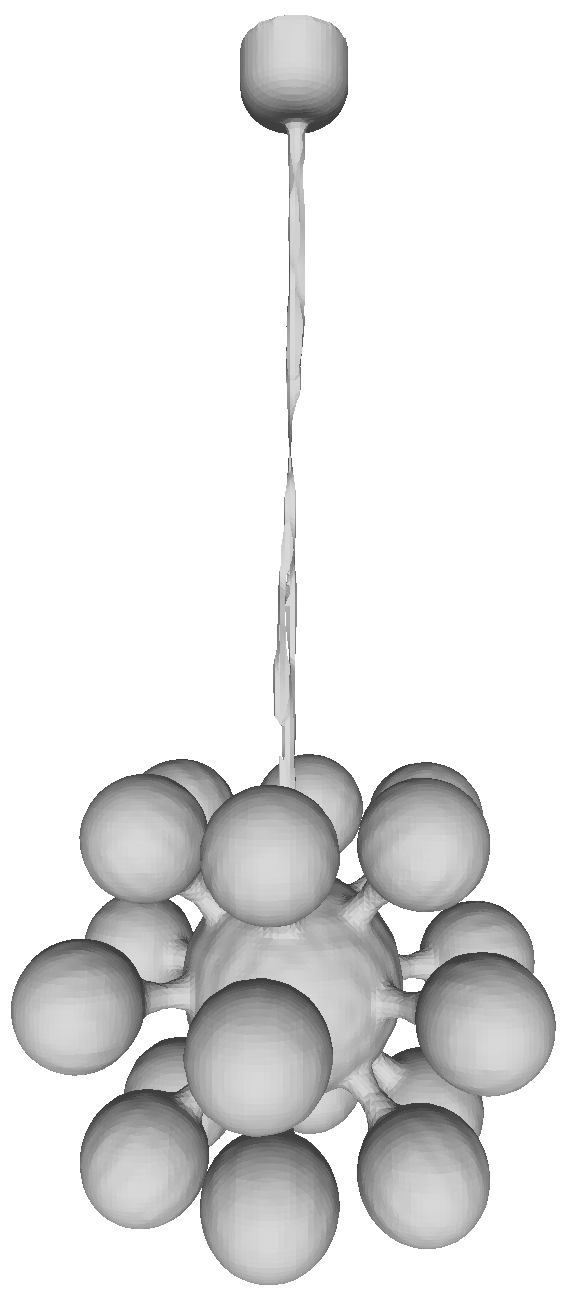} &
\includegraphics[width=\imgwidth, height=3cm, keepaspectratio]{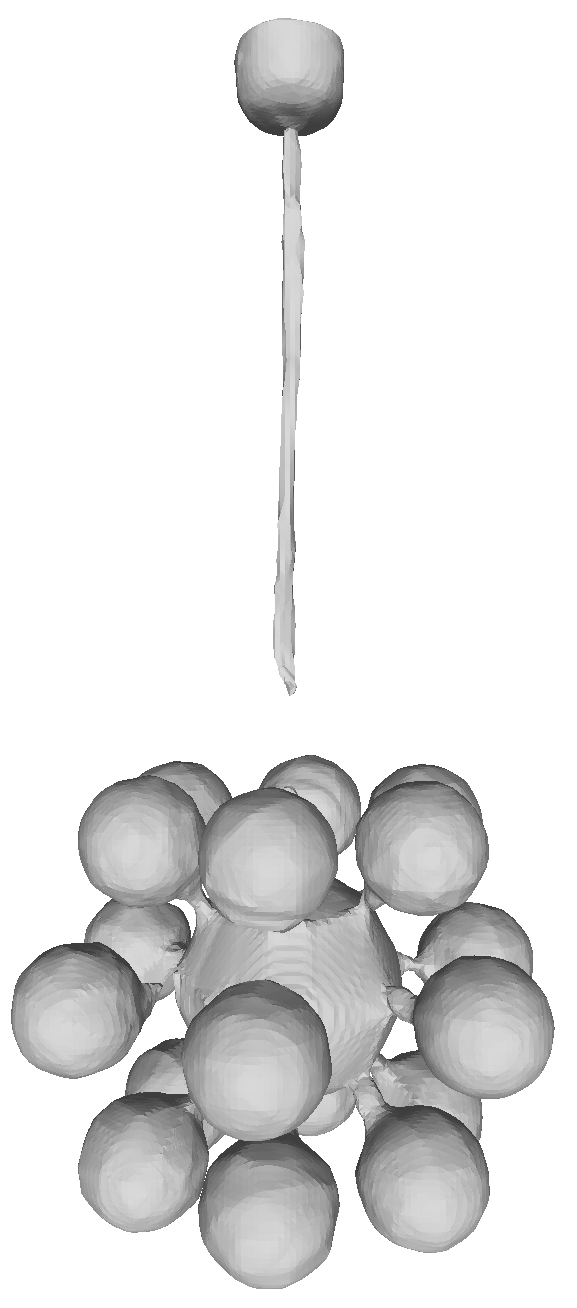} &
\includegraphics[width=\imgwidth, height=3cm, keepaspectratio]{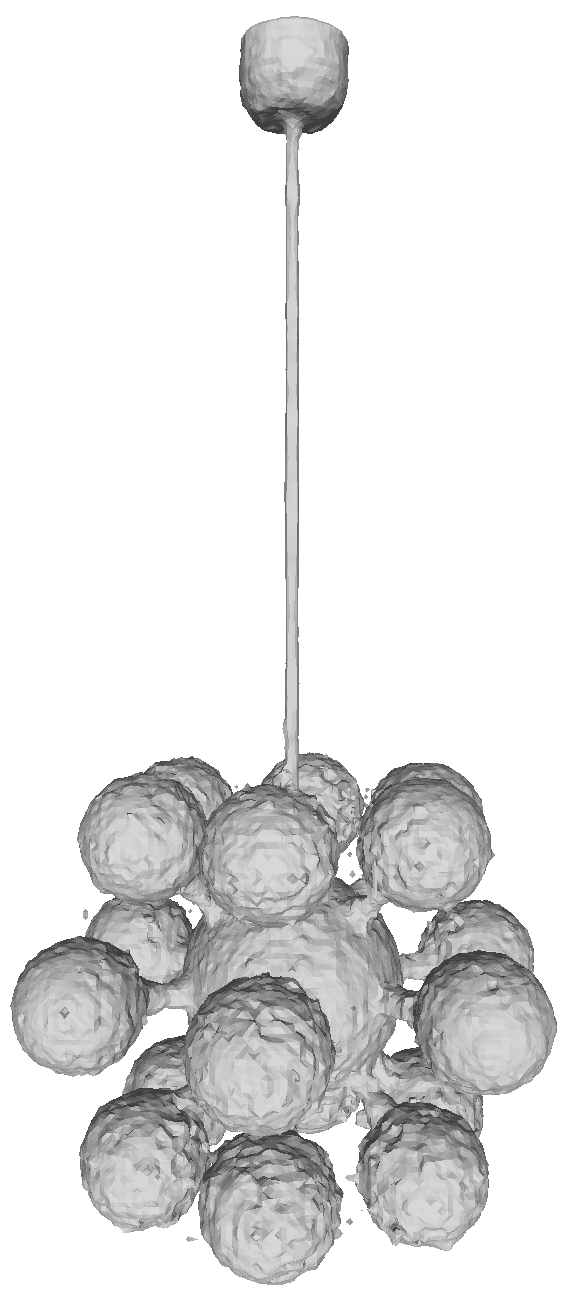} &
\includegraphics[width=\imgwidth, height=3cm, keepaspectratio]{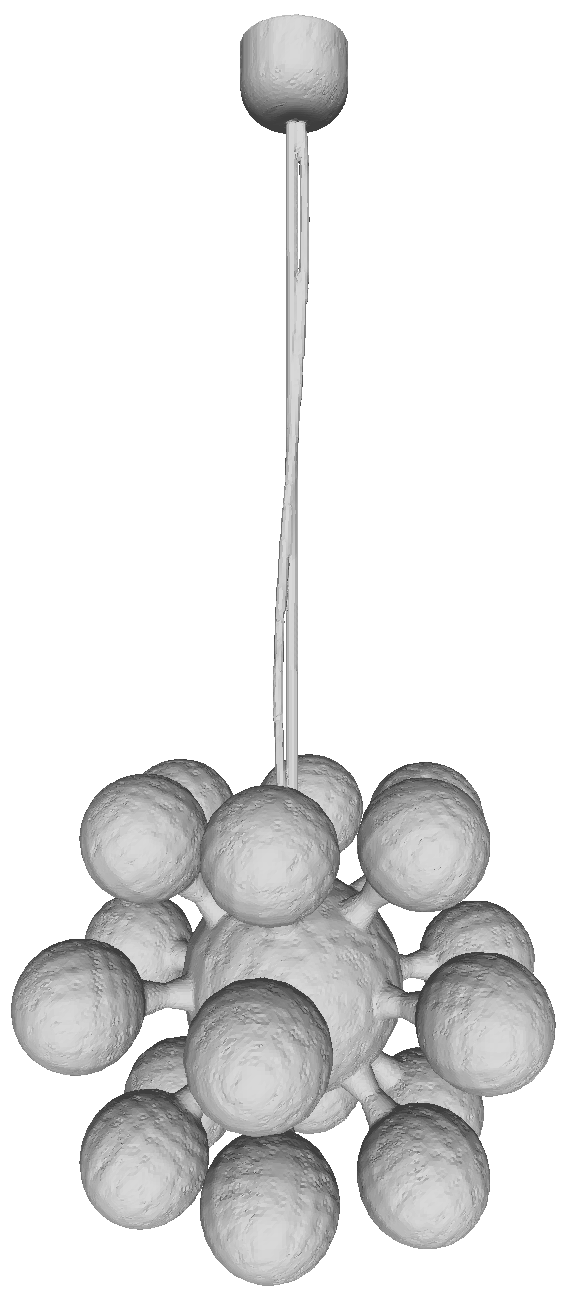} \\

\includegraphics[width=\imgwidth]{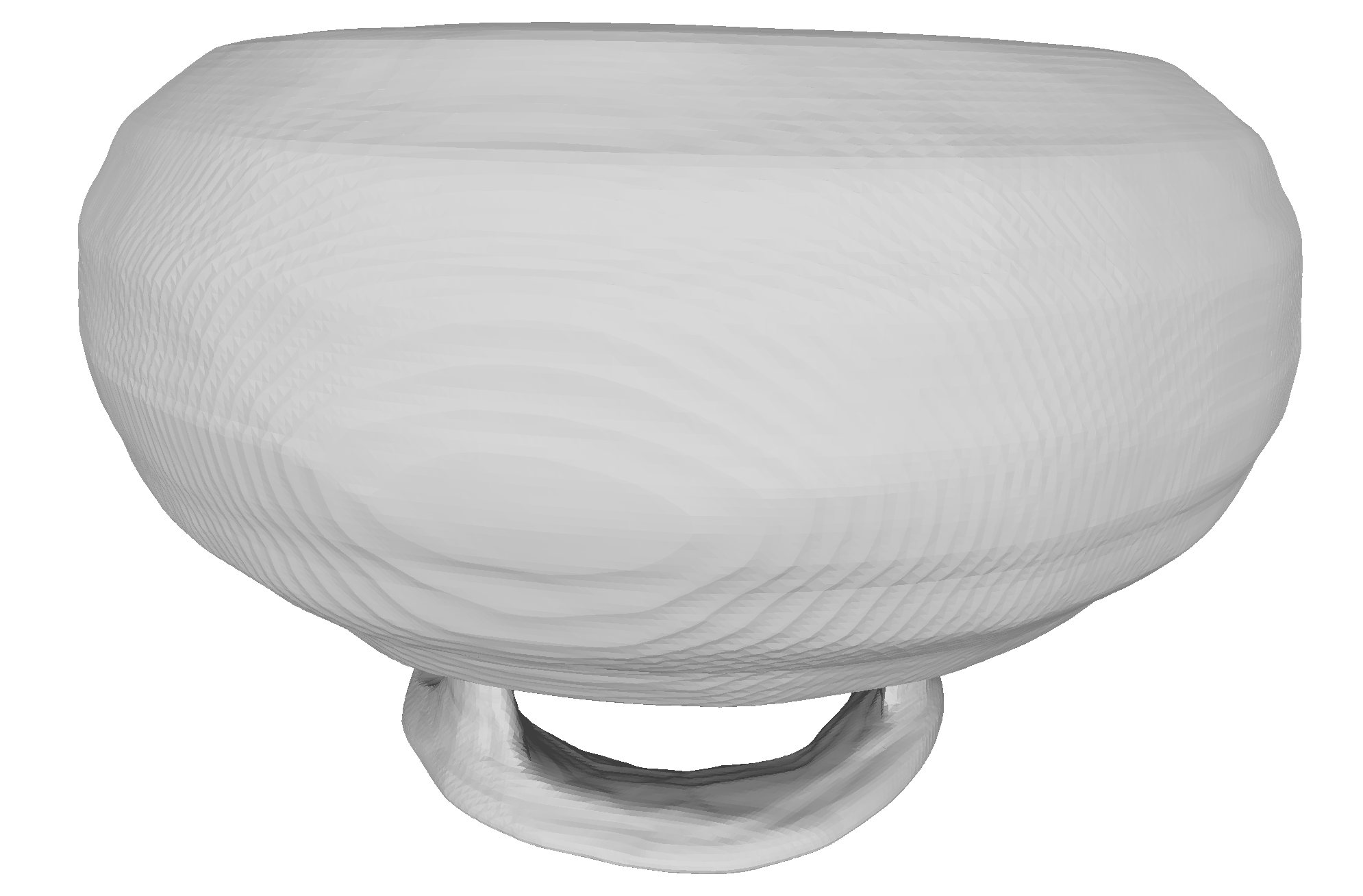} &
\includegraphics[width=\imgwidth]{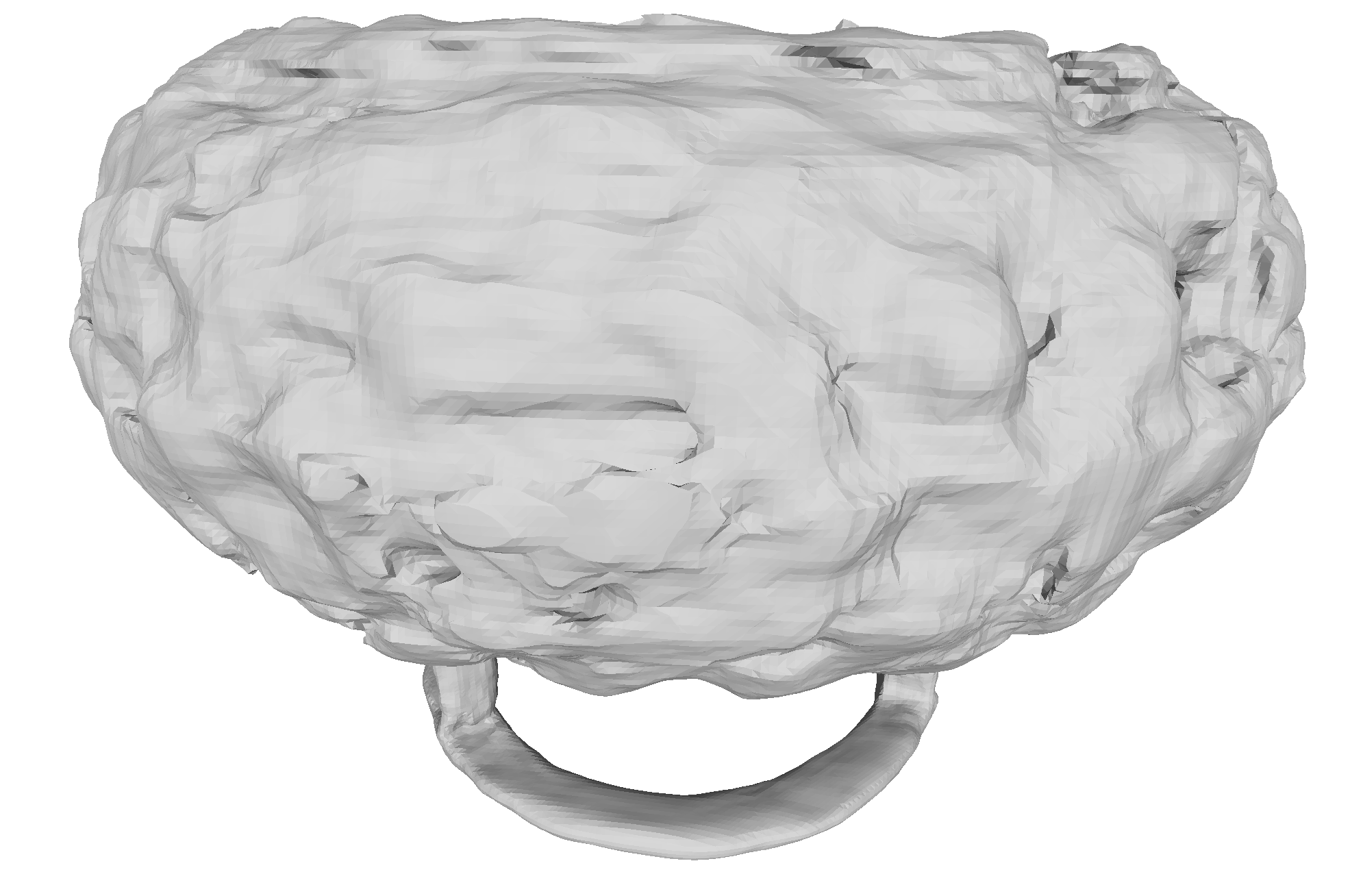} &
\includegraphics[width=\imgwidth]{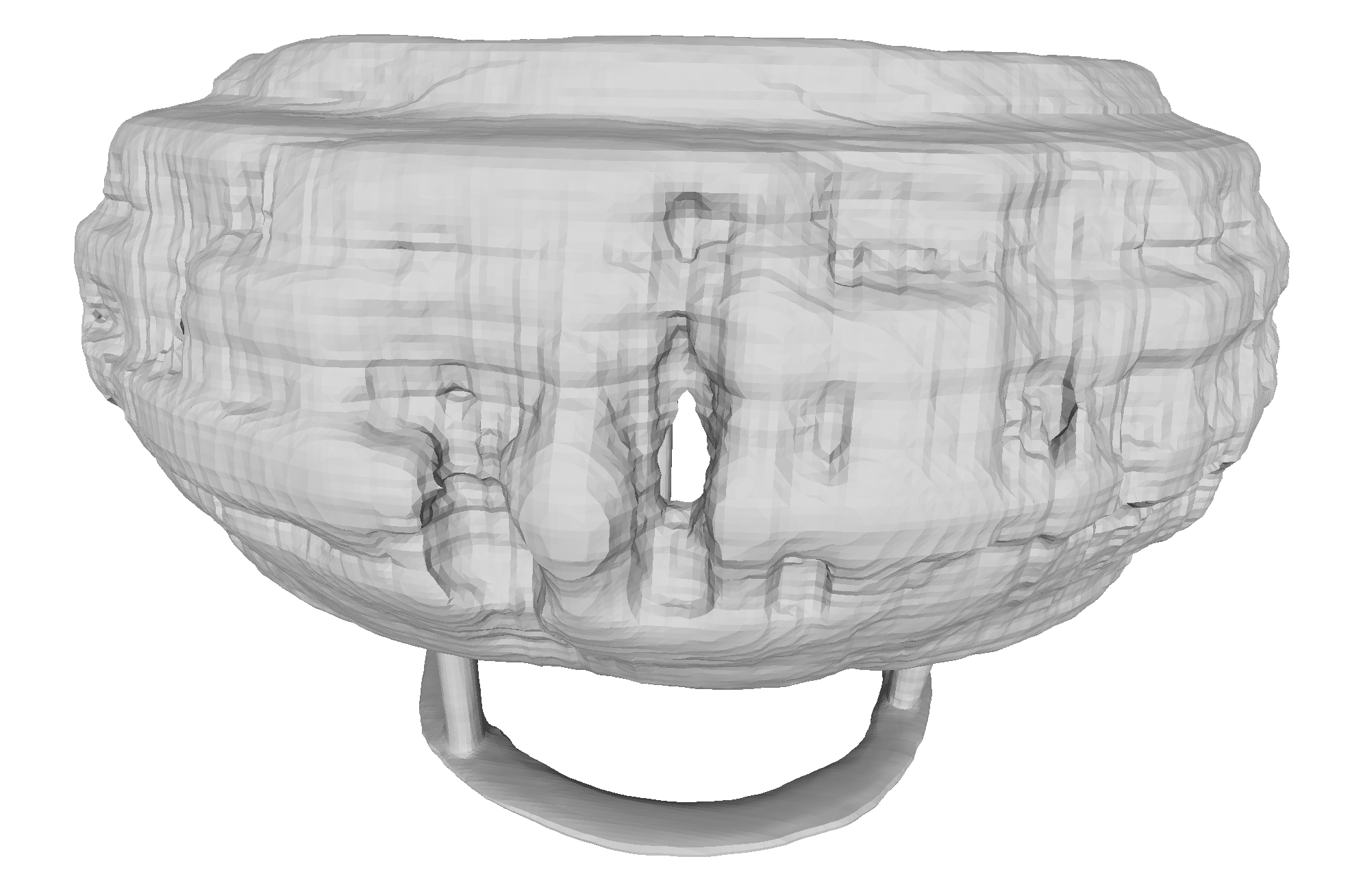} &
\includegraphics[width=\imgwidth]{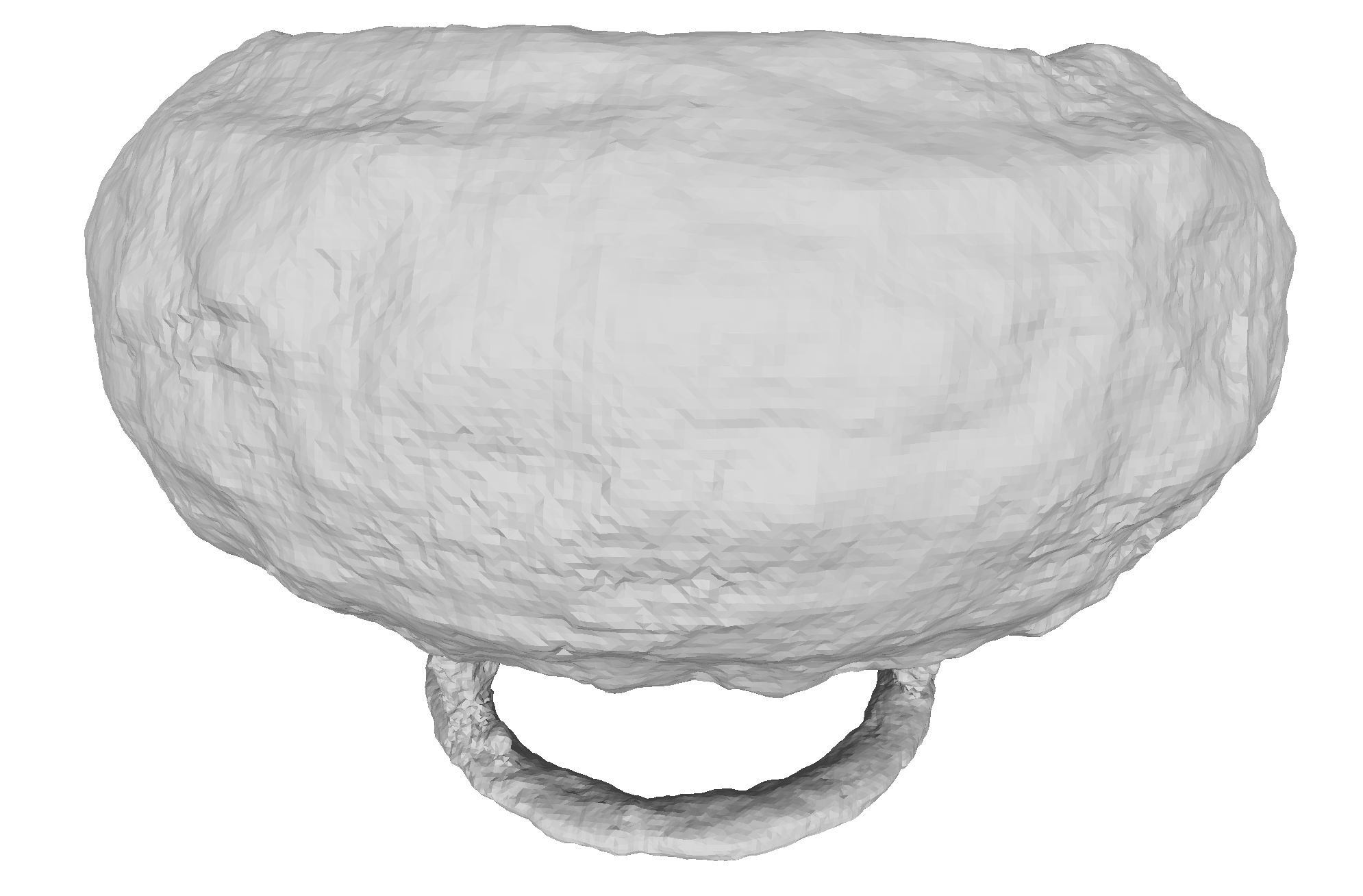} &
\includegraphics[width=\imgwidth]{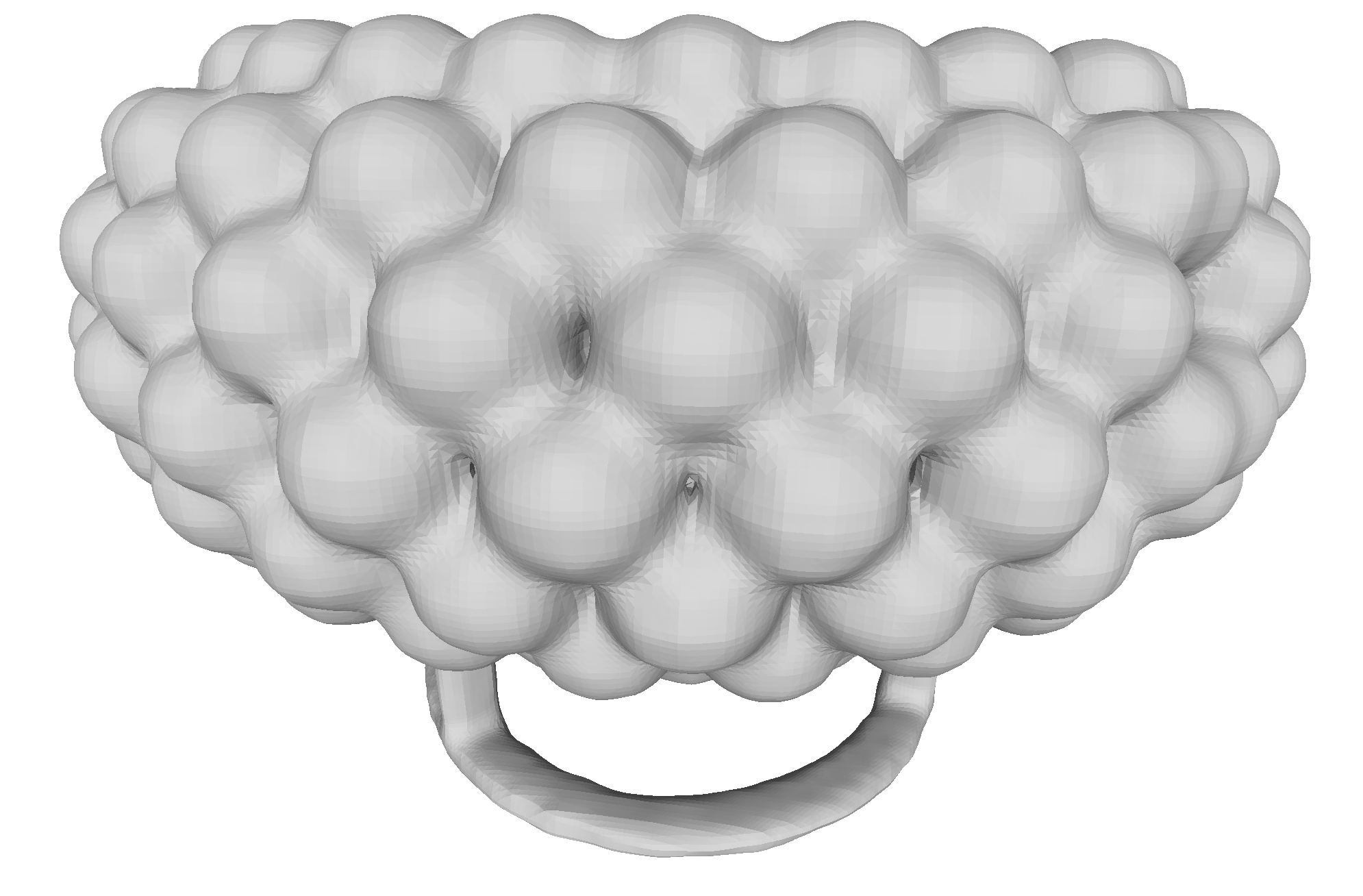} &
\includegraphics[width=\imgwidth]{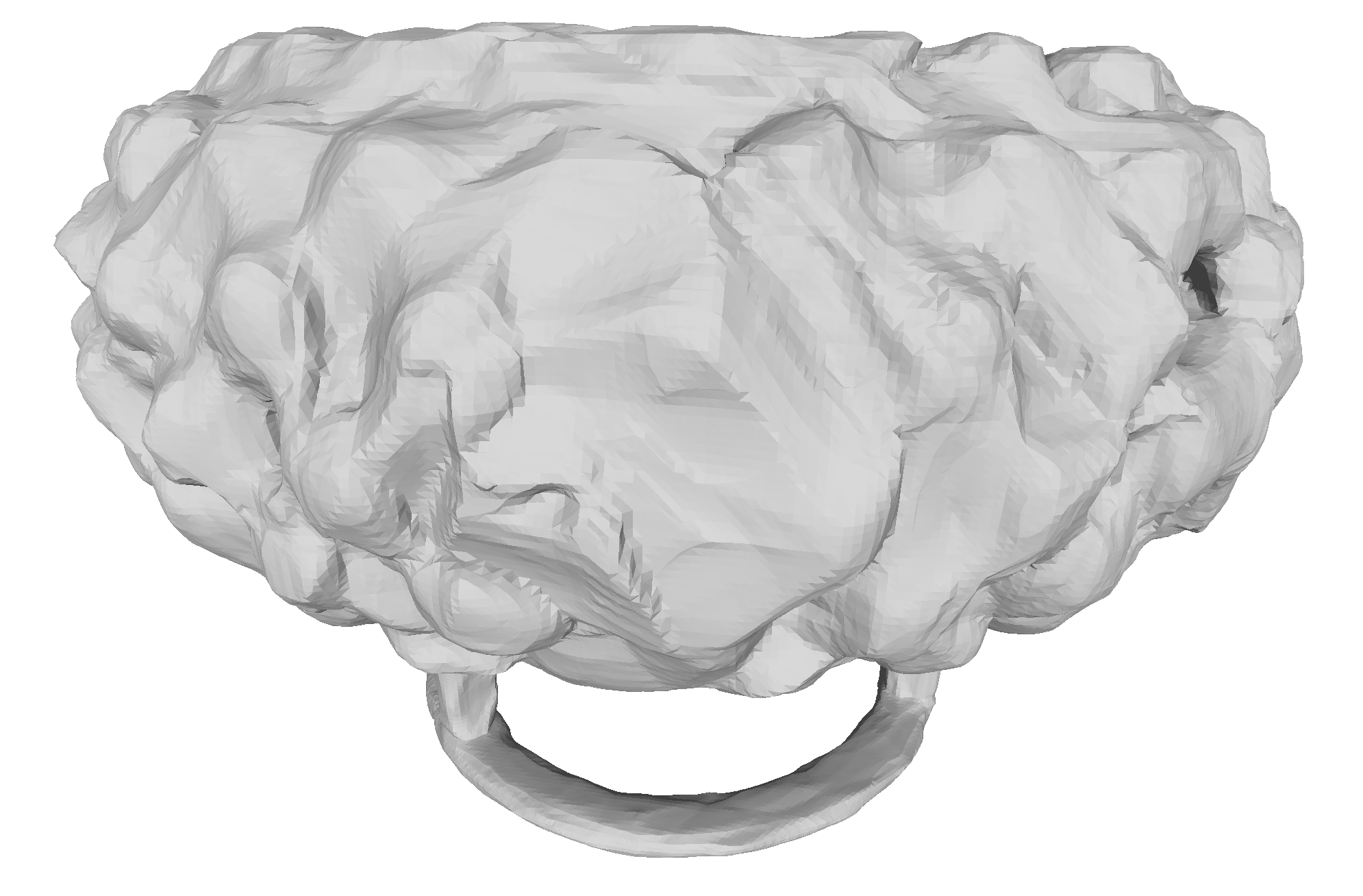} &
\includegraphics[width=\imgwidth]{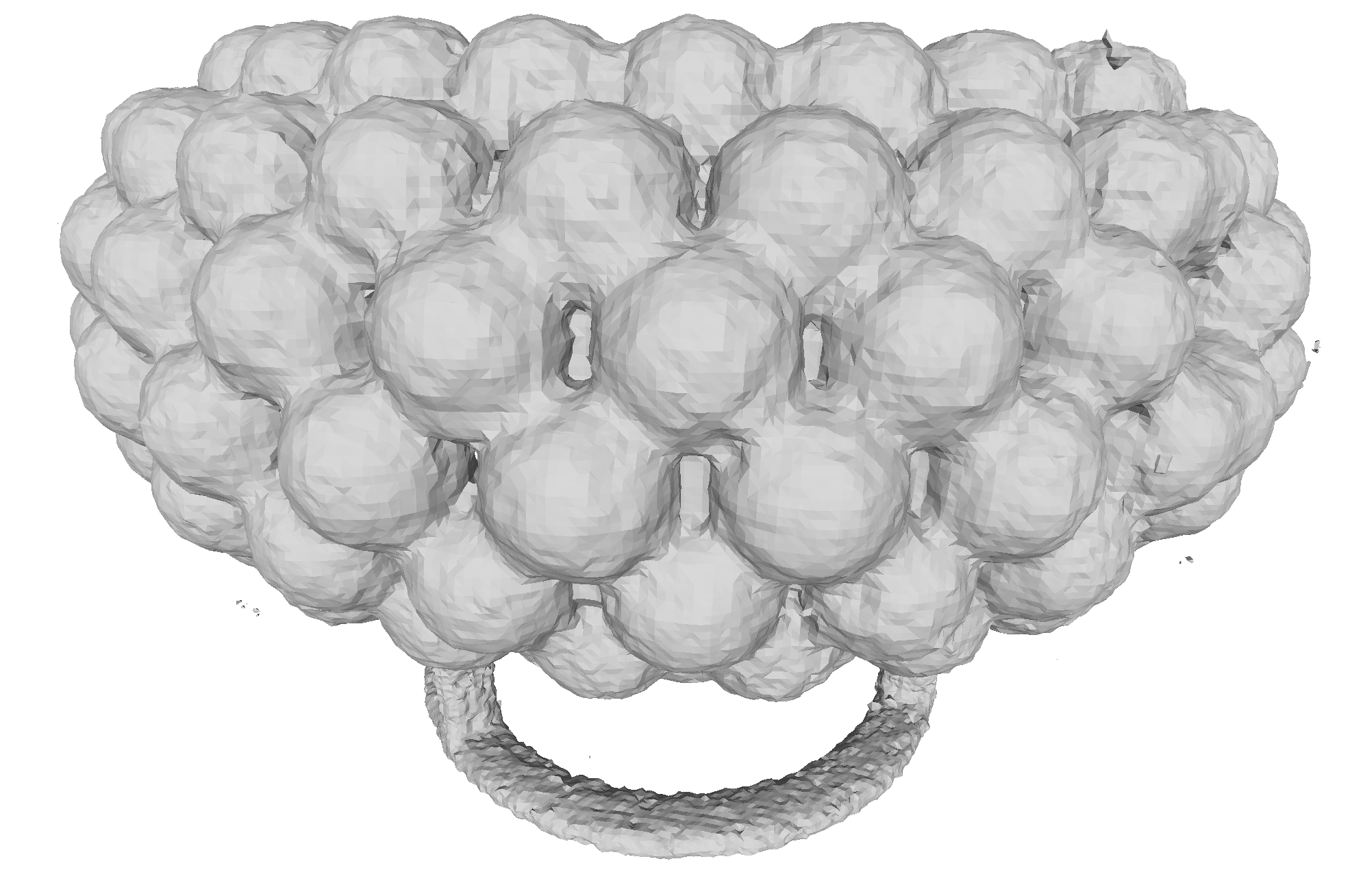} &
\includegraphics[width=\imgwidth]{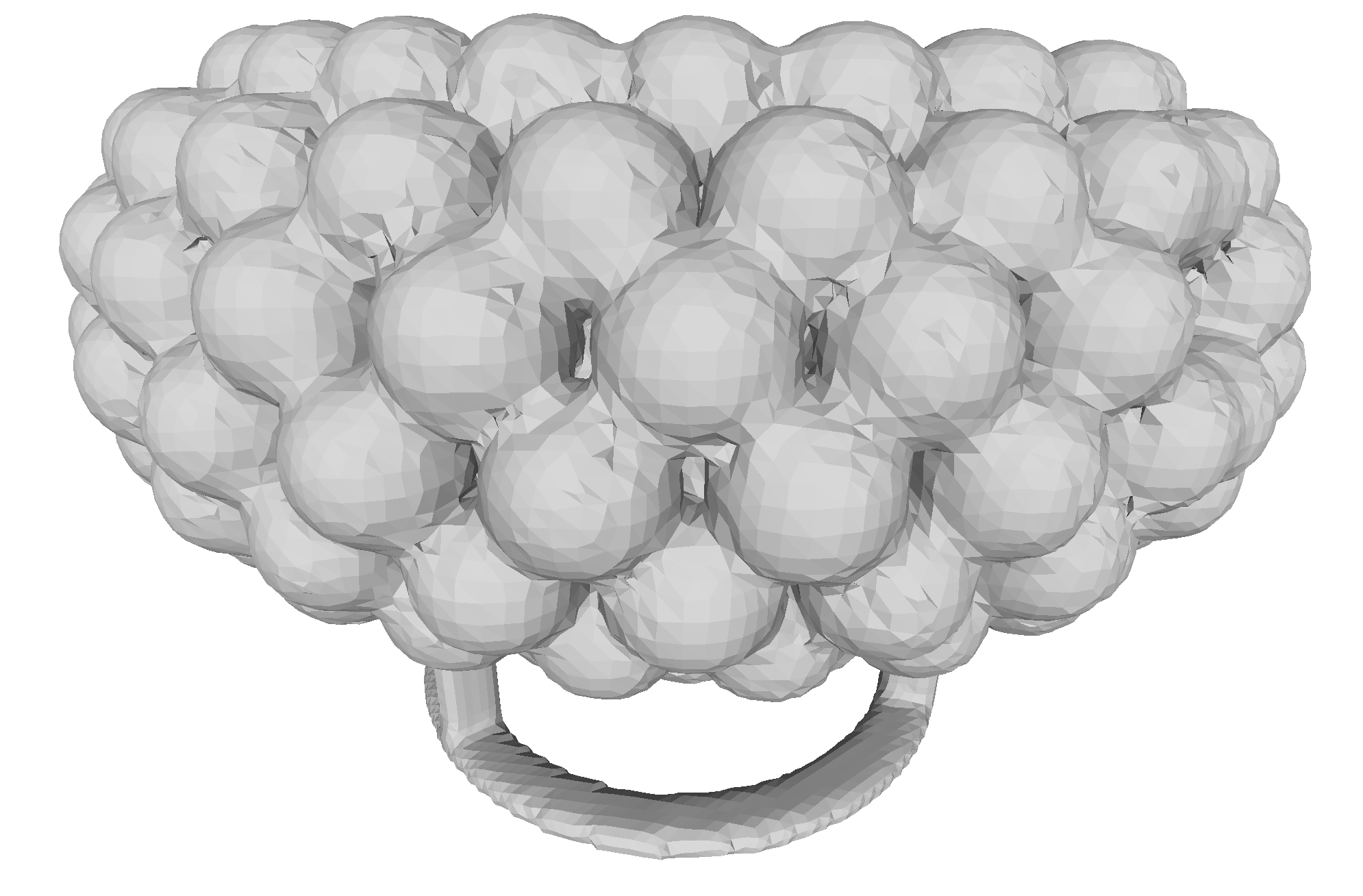} \\

  \end{tabular}
  \caption{Comparison in multiple object reconstruction on ShapeNet.}
  \label{fig:multi-shapes-reconstruction-results}
\end{figure*}

In Fig. \ref{fig:multi-shapes-reconstruction-results}, we present additional reconstruction results on the ShapeNet dataset, further demonstrating the strength of our method in preserving fine-grained details. We showcase reconstructions from multiple object categories used during training, including chairs with structural holes, lamps with significant geometric variation, and airplanes with complex shapes. As shown, our method effectively recovers detailed surface textures.

\section{Shape Completion}

We present additional shape completion examples on the DFAUST dataset.  
The results follow the same procedure as in the main paper and further demonstrate the consistency of our method.

\begin{figure*}
  \centering
\includegraphics[width=1.0\linewidth]{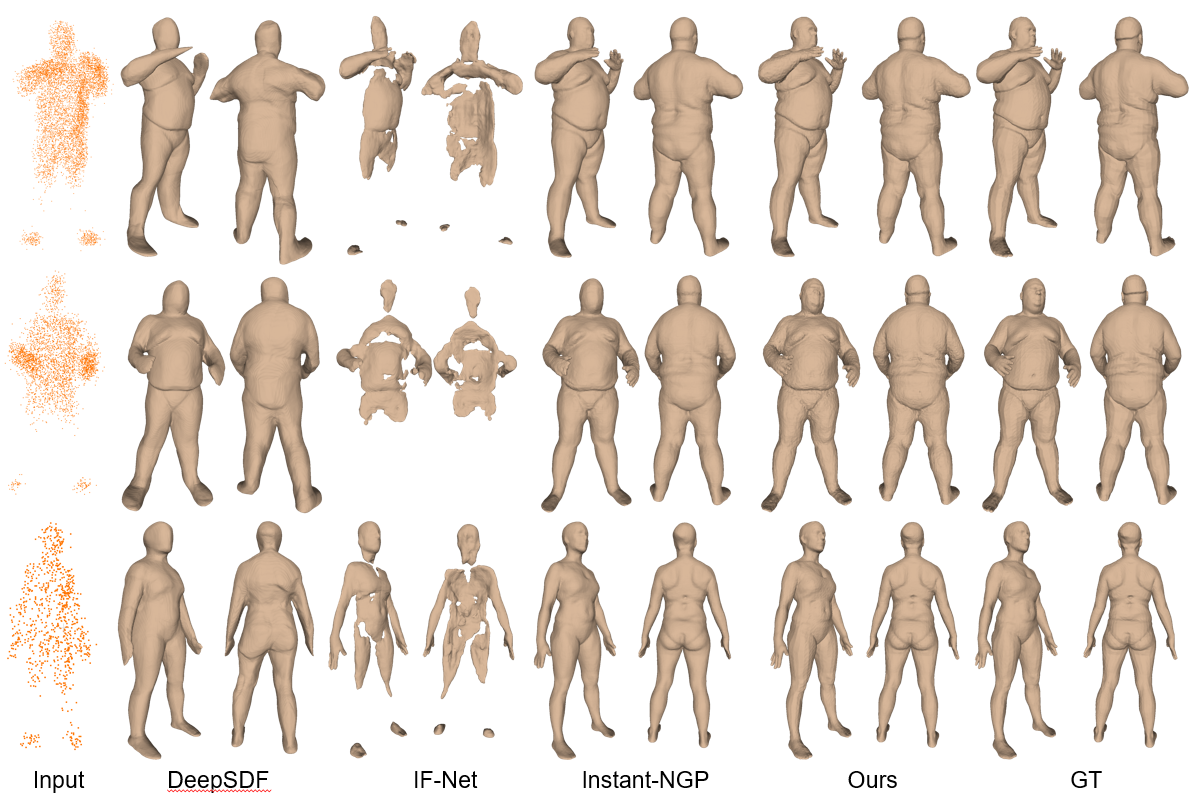} 
  \caption{Additional Shape Completion Results on DFAUST.}
  \label{fig:supp_completion_shapes}
\end{figure*}

\begin{table*}
\centering
\begin{tabular}{clllll}
\toprule
Methods & Models & CD & F-score & Precision & Recall \\ 
\midrule
\multirow{6}{*}{ACORN} & Armadillo & 1.40e-5 & 0.998 & 0.998 & 0.999 \\
 & Drango & 3.98e-6 & 1.000 & 1.000 & 1.000 \\
 & Asian Dragon & 1.19e-5 & 0.987 & 0.978 & \textbf{0.996} \\
 & Happy Buddha & 1.13e-5 & 0.972 & 0.950 & 0.994 \\
 & Lucy & 1.03e-5 & 0.986 & 0.973 & 1.000 \\
 & Thai Statue & \textbf{3.54e-4} & 0.948 & \textbf{0.902} & 1.000 \\ \midrule
\multirow{6}{*}{NGLOD} & Armadillo & 1.09e-5 & 0.999 & 0.998 & 0.999 \\
 & Drango & 4.18e-6 & 1.000 & 1.000 & 1.000 \\
 & Asian Dragon & 1.09e-5 & 0.984 & 0.979 & 0.990 \\
 & Happy Buddha & 9.86e-6 & 0.973 & 0.955 & 0.993 \\
 & Lucy & 5.03e-6 & 0.988 & 0.976 & 1.000 \\
 & Thai Statue & 3.65e-4 & 0.939 & 0.899 & 0.982 \\ \midrule
\multirow{6}{*}{Instant-NGP} & Armadillo & 1.58e-5 & 0.993 & 0.991 & 0.995 \\
 & Drango & 6.99e-6 & 0.999 & 1.000 & 0.999 \\
 & Asian Dragon & 1.36e-5 & 0.983 & 0.975 & 0.992 \\
 & Happy Buddha & 1.43e-5 & 0.966 & 0.942 & 0.990 \\
 & Lucy & 1.37e-5 & 0.974 & 0.969 & 0.979 \\
 & Thai Statue & 3.78e-4 & 0.941 & 0.898 & 0.988 \\ \midrule
\multirow{6}{*}{M-SDF} & Armadillo & 2.92e-3 & 0.883 & 0.793 & 0.995 \\
 & Drango & 5.45e-6 & 0.990 & 0.997 & 0.983 \\
 & Asian Dragon & 1.24e-3 & 0.947 & 0.906 & 0.992 \\
 & Happy Buddha & 1.93e-3 & 0.836 & 0.723 & 0.990 \\
 & Lucy & 8.82e-4 & 0.899 & 0.819 & 0.998 \\
 & Thai Statue & 8.23e-4 & 0.857 & 0.840 & 0.874 \\ \midrule
\multirow{6}{*}{HyperDiffusion} & Armadillo & 3.17e-5 & 0.849 & 0.843 & 0.855 \\
 & Drango & 1.11e-4 & 0.865 & 0.882 & 0.848 \\
 & Asian Dragon & 5.68e-5 & 0.836 & 0.850 & 0.822 \\
 & Happy Buddha & 3.20e-5 & 0.837 & 0.844 & 0.830 \\
 & Lucy & 2.53e-5 & 0.874 & 0.890 & 0.857 \\
 & Thai Statue & 4.64e-4 & 0.749 & 0.771 & 0.728 \\ \midrule
\multirow{6}{*}{Ours} & Armadillo & \textbf{9.14e-6} & \textbf{0.999} & \textbf{0.999} & \textbf{0.999} \\
 & Drango & \textbf{2.28e-6} & \textbf{1.000} & \textbf{1.000} & \textbf{1.000} \\
 & Asian Dragon & \textbf{9.80e-6} & \textbf{0.987} & \textbf{0.980} & 0.994 \\
 & Happy Buddha & \textbf{9.00e-6} & \textbf{0.974} & \textbf{0.956} & \textbf{0.994} \\
 & Lucy & \textbf{3.95e-6} & \textbf{0.988} & \textbf{0.976} & \textbf{1.000} \\
 & Thai Statue & 3.59e-4 & \textbf{0.948} & 0.901 & \textbf{1.000} \\ \bottomrule
\end{tabular}
\caption{Quantitative comparison on the Stanford models.} 
\label{tab:supp_stanford_quantitative_comparison}
\end{table*}

\end{document}